\documentclass{article}

\usepackage[main, final]{neurips_2026}

\usepackage[utf8]{inputenc} 
\usepackage[T1]{fontenc}    
\usepackage{hyperref}       
\usepackage{url}            
\usepackage{booktabs}       
\usepackage{amsfonts}       
\usepackage{nicefrac}       
\usepackage{microtype}      
\usepackage{xcolor}         

\usepackage{wrapfig}
\usepackage{enumitem}

\usepackage{amsmath}
\usepackage{amssymb}
\usepackage{mathtools}
\usepackage{amsthm}
\usepackage[capitalize,noabbrev]{cleveref}
\usepackage{subcaption}
\usepackage{multirow}
\usepackage[table]{xcolor}
\usepackage{amsthm}
\usepackage{titletoc}

\newtheorem{theorem}{Theorem}
\newtheorem{definition}{Definition}

\title{E0: Expressive Fine-Grained Discrete Action Prediction for Vision-Language-Action Models via Tweedie Discrete Diffusion}

\author{
\textbf{Zhihao Zhan}$^{1}$\thanks{First author. Email: \texttt{zhanzhh6@mail2.sysu.edu.cn}},
\textbf{Jiaying Zhou}$^{1}$,
\textbf{Likui Zhang}$^{1}$,
\textbf{Qinhan Lyu}$^{1}$,
\textbf{Hao Liu}$^{1}$,\\
\textbf{Weizheng Li}$^{1}$,
\textbf{Jusheng Zhang}$^{1}$,
\textbf{Ziliang Chen}$^{1}$,
\textbf{Tianshui Chen}$^{3,4}$,\\
\textbf{Ruifeng Zhai}$^{1}$,
\textbf{Keze Wang}$^{1,2,3}$,
\textbf{Liang Lin}$^{1,2,3}$,
\textbf{Guangrun Wang}$^{1,2,3}$\thanks{Corresponding author. Email: \texttt{wanggrun@gmail.com}}\\[1mm]
$^1$Sun Yat-sen University, $^2$Guangdong Key Laboratory of Big Data Analysis and Processing\\
$^3$X-Era AI Lab, $^4$Guangdong University of Technology\\
}

\begin{document}

\maketitle

\begin{abstract}
Vision–Language–Action (VLA) models offer a unified framework for robotic manipulation by integrating visual perception, language understanding, and control generation. However, existing VLA systems still struggle to generalize across diverse tasks, scenes, and camera viewpoints, and often produce coarse or unstable actions. We argue that these limitations are closely tied to the structural properties of actions in VLA settings, including the inherent multi-peaked nature of action distributions, the token-based symbolic reasoning of pretrained VLM/VLA backbones, and the effective finite resolution imposed by real-world robotic control. Motivated by these properties, we introduce E0, a tweedie discrete diffusion framework that formulates action generation as iterative denoising over quantized action tokens. By operating in a discrete action space with a principled diffusion process, E0 naturally aligns with token-based reasoning, supports fine-grained yet executable action control, and avoids the distributional mismatch of masking-based discrete diffusion. Experiments on LIBERO, VLABench, ManiSkill, and a real-world Franka arm demonstrate that E0 achieves state-of-the-art performance across diverse environments, outperforming strong baselines by 8.1\% on average. Project: \href{https://doo-mon.github.io/e0web/}{https://doo-mon.github.io/e0web/}.
\end{abstract}

\section{Introduction} \label{sec-intro}

Robotic manipulation in open-ended environments requires models that can perceive complex visual scenes, follow natural language instructions, and generate precise, reliable actions. Vision--Language--Action (VLA) models have recently emerged as a unified paradigm for this goal by integrating visual perception, language understanding, and control generation within a single framework. Leveraging large-scale multimodal pretraining, these models aim to generalize across tasks, scenes, and object categories. However, despite their promise, existing VLA systems still struggle to cope with diverse task instructions, environment configurations, and camera viewpoints, and often produce coarse or unstable actions that fail in fine-grained manipulation.

Recent VLA systems have explored a diverse range of formulations for action generation.
Representative approaches include autoregressive token prediction over discretized action spaces~\citep{brohan2022rt, zitkovich2023rt, kim2024openvla, pertsch2025fast}, masked diffusion(\citep{nie2025large, wu2025fast, ye2025dream}) over discrete action tokens~\citep{liang2025discrete, song2026fast}, as well as continuous diffusion models that iteratively denoise real-valued action trajectories~\citep{black2024pi_0, intelligence2025pi05, lin2025onetwovla, zhang2026atomicvla}.
While differing in architectural design and training objectives, these methods largely adopt action representations inherited from existing sequence modeling or diffusion paradigms, rather than being derived from a principled analysis of action features specific to VLA settings.

\begin{wrapfigure}{r}{0.5\textwidth}
  \centering
  \includegraphics[width=0.45\textwidth]{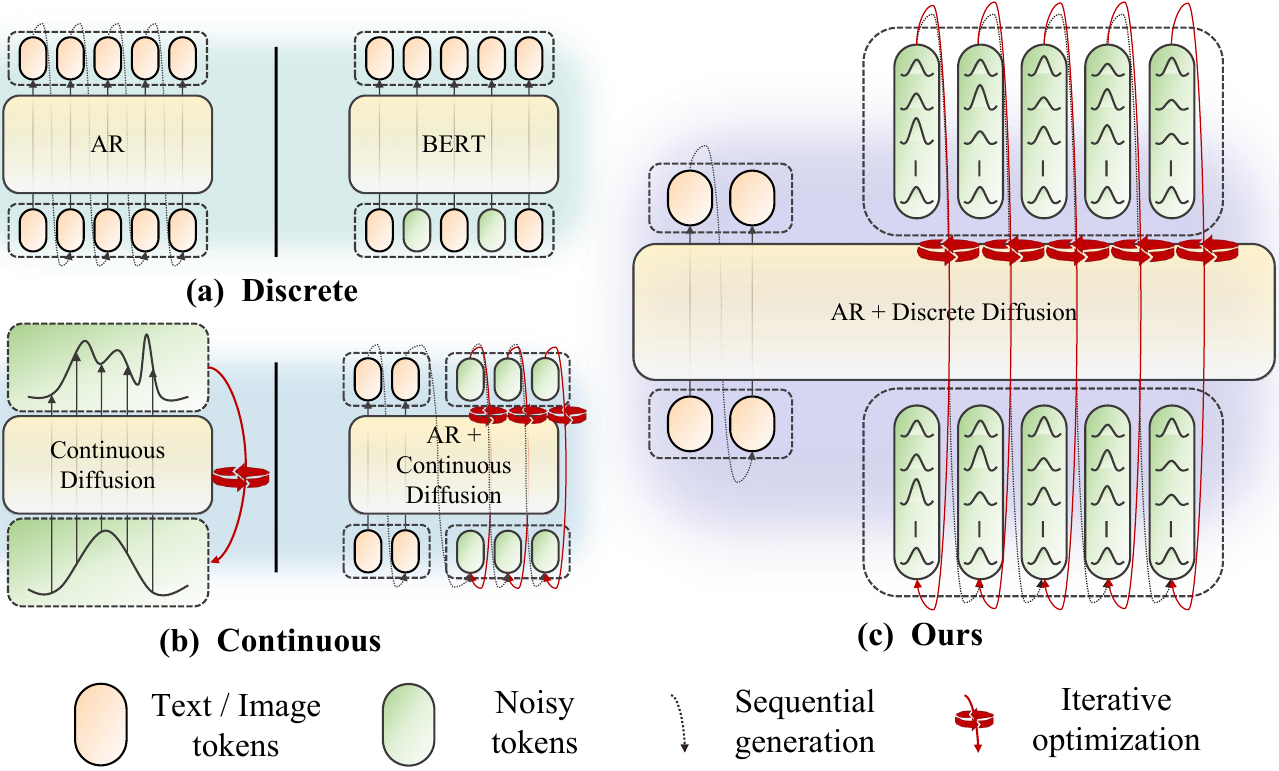}
  \caption{
  \textbf{Overview of action modeling paradigms.}
(a) \textit{Discrete modeling}: Autoregressive (AR) approaches~\citep{brohan2022rt,zitkovich2023rt, kim2024openvla} and recent mask-based discrete diffusion methods~\citep{liang2025discrete, song2026fast}, which operate over a small discrete action vocabulary.
(b) \textit{Continuous modeling}: Continuous diffusion--based policies~\citep{liu2024rdt, xu2025a0} and AR--diffusion hybrids~\citep{black2024pi_0, intelligence2025pi05, lin2025onetwovla} that regress continuous actions.
(c) \textit{Our approach}: E0 integrates AR-style conditioning with Tweedie discrete diffusion, preserving compatibility with pretrained vision--language backbones and allowing fine-grained action generation.
  }
  \label{fig-paradigm_comparison}
  \vspace{-0.8em}
\end{wrapfigure}

To better understand the challenges of action modeling in VLA systems, it is useful to examine the structural properties of actions in this setting.
First, robotic actions are inherently \textbf{multi-peaked} \citep{florence2021decisiveness,shafiullah2022behavior,florence2022implicit}: under the same task specification and visual context, multiple distinct action sequences may successfully accomplish the task, making action prediction fundamentally a distribution modeling problem rather than single-valued regression (see Appendix~\ref{sect:appendix-Multimodal} for additional analysis).
Second, action generation in VLA systems is tightly coupled with pretrained vision–language backbones, where high-level reasoning, instruction following, and semantic conditioning are mediated through \textbf{discrete symbolic tokens} \citep{liang2025discrete,li2025discrete,luo2026being}.
These backbones are optimized with token-level classification objectives, under which semantic information is primarily encoded through relative feature directions, shaping the representational and optimization characteristics of VLM/VLA models (see Appendix~\ref{sect:appendix-vlm} for additional analysis).
Third, although physical execution evolves in continuous time, practical robotic systems operate under finite control rates, encoder resolution, actuation latency, and environmental noise.
These factors collectively impose an \textbf{effective finite resolution} on action execution at the policy level, which may differ substantially from nominal hardware specifications due to real-world physical constraints (see  Appendix~\ref{sect:appendix-nature} for additional analysis). 
Taken together, these properties highlight fundamental constraints on action modeling in VLA systems, motivating formulations that can capture multi-peaked behaviors, align with token-based reasoning, and operate under finite-resolution execution limits.

For \textbf{multi-peaked} action distributions, deterministic action regression with squared-error objectives can average distinct valid behaviors, producing predictions that deviate from executable action modes. Our empirical analysis on LIBERO demonstrations confirms this geometric effect: the distance between the conditional mean and its nearest empirical action mode increases by 46.29\% from low- to high-multimodality states. These observations motivate action-generation formulations that preserve distinct behavioral modes, which we further investigate through controlled comparisons between E0 and a continuous-flow VLA.
For action generation mediated by \textbf{discrete symbolic tokens} in pretrained vision--language backbones, discrete action spaces enable representational and optimization-level alignment with token-based reasoning.
Finally, the \textbf{effective finite resolution} imposed by real-world robotic control makes discrete abstractions well matched to the granularity at which actions can be reliably generated and executed. 

Collectively, these observations suggest that discreteness in action modeling follows naturally from the multi-peaked structure of actions, the token-based reasoning of VLA models, and the finite resolution of real-world control. To this end, we propose $\text{E}_0$, built upon Tweedie Discrete Diffusion (TDD), a principled formulation featuring score calibration and projected reverse dynamics that provides mathematical bridges between continuous generative processes and discrete action representations. By applying Gaussian noise directly to discrete action spaces, TDD maintains forward-reverse consistency and avoids the distributional mismatch inherent in mask-based diffusion. Unlike AR-based methods, our approach supports arbitrarily fine discretization bins, enabling high-resolution action modeling beyond the constraints of language tokenizers.

\textbf{In summary, our contributions are twofold:}
\begin{enumerate}[itemsep=0.2em, topsep=0.2em, parsep=0pt, partopsep=0pt]
    \item We introduce Tweedie Discrete Diffusion (TDD) as a novel framework for action generation. TDD represents a principled formulation featuring score calibration and projected reverse dynamics that establishes rigorous mathematical bridges between continuous score-matching and discrete classification, supporting arbitrarily fine discretization while preserving pre-trained vision-language features.
    \item We demonstrate the effectiveness of $\text{E}_0$ through extensive experiments across multiple simulation benchmarks and real-world Franka manipulation tasks. Furthermore, we validate that the performance gains strictly arise from the TDD objective.  
\end{enumerate}

\section{Related Work} \label{sec-relate}

\noindent\textbf{Autoregressive VLA Models.}
Early efforts such as RT-1~\cite{brohan2022rt} and RT-2~\cite{zitkovich2023rt} pioneered VLA models, demonstrating the potential of scaling language-conditioned policies. Building on this foundation, OpenVLA~\cite{kim2024openvla} released the first open-source autoregressive VLA by integrating Llama-2~\cite{touvron2023llama} with strong vision encoders (DINOv2~\cite{oquab2023dinov2}, SigLIP~\cite{zhai2023sigmoid}), while SpatialVLA~\cite{qu2025spatialvla} introduced Ego3D position encoding to inject explicit 3D spatial cues. To enhance tokenized action generation, OpenVLA-OFT~\cite{kim2025fine} proposed an optimized fine-tuning recipe with parallel decoding and continuous action representations. $\pi_0$ FAST~\cite{pertsch2025fast} designed a frequency-aware tokenization scheme for efficient autoregressive training. In parallel, reasoning-augmented approaches such as CoT-VLA~\cite{zhao2025cot} and generative rollouts (GR-1~\cite{wu2023unleashing}, GR-2~\cite{cheang2024gr}) extend VLAs with textual or visual chain-of-thought to guide planning. Our work builds on these advances by preserving the strong contextual alignment afforded by autoregressive decoding, while addressing its inherent limitations in producing fine-grained continuous actions.

\noindent\textbf{Diffusion-based VLA Models.}
The application of diffusion to robotic imitation learning was first introduced by Diffusion Policy~\cite{chi2025diffusion}, demonstrating its promise for continuous action generation. Subsequent works have advanced this paradigm through increasingly sophisticated designs: CogACT~\cite{li2024cogact} developed diffusion action transformers with favorable scaling properties, while RDT~\cite{liu2024rdt} exploited DiT’s cross-modal fusion to generate complete action sequences directly from noisy action chunks, later extended by A0~\cite{xu2025a0} with affordance-based reasoning. Generalist frameworks such as $\pi_0$~\cite{black2024pi_0} and $\pi_{0.5}$~\cite{intelligence2025pi05} proposed powerful flow models for broad robotic control, achieving strong performance across diverse real-world tasks. Other recent directions include OneTwoVLA~\cite{lin2025onetwovla}, which enables dynamic switching between reasoning and execution; AtomicVLA~\cite{zhang2026atomicvla}, which leverages multiple action experts to precisely learn atomic actions; GR00t N1~\cite{bjorck2025gr00t} with a dual-system design, and GO-1~\cite{bu2025agibot}, a scalable generalist policy using latent actions. Finally, Hybrid VLA~\cite{liu2025hybridvla} combines autoregressive and diffusion decoding to leverage their complementary strengths. In contrast, our work preserves the expressive continuous modeling ability of diffusion while maintaining the emergent language grounding and multimodal reasoning capacities of AR VLAs.

\noindent\textbf{Discrete Diffusion.}
Recent advances have sparked growing interest in discrete diffusion models. A common formulation treats discrete diffusion as a masked modeling problem, in which noise is simulated through random token masking and the model is trained to reconstruct the original input~\cite{nie2025large, wu2025fast, ye2025dream}. Recently, an attempt has been made to extend discrete diffusion to VLA models~\cite{liang2025discrete, song2026fast}; however, their design follows a BERT-like~\cite{devlin2019bert} masking mechanism and requires additional architectural complexity to achieve competitive performance. In contrast, our discrete diffusion provides a reformulation that operates directly on float-encoded one-hot class representations, avoiding both continuous latent diffusion and heuristic masking strategies. During inference, the model predicts a clean categorical state at each step by applying an $\arg\max$ operation over output probabilities and converting the result back into a one-hot vector.  Such an autoregressive-style feedback mechanism enables gradual refinement of predictions. Building upon this foundation, our work makes more stable action generation and achieves competitive performance in complex control scenarios.

\section{Method} \label{sec-method}

\begin{figure*}[t]
  \centering
  \includegraphics[width=0.9\textwidth]{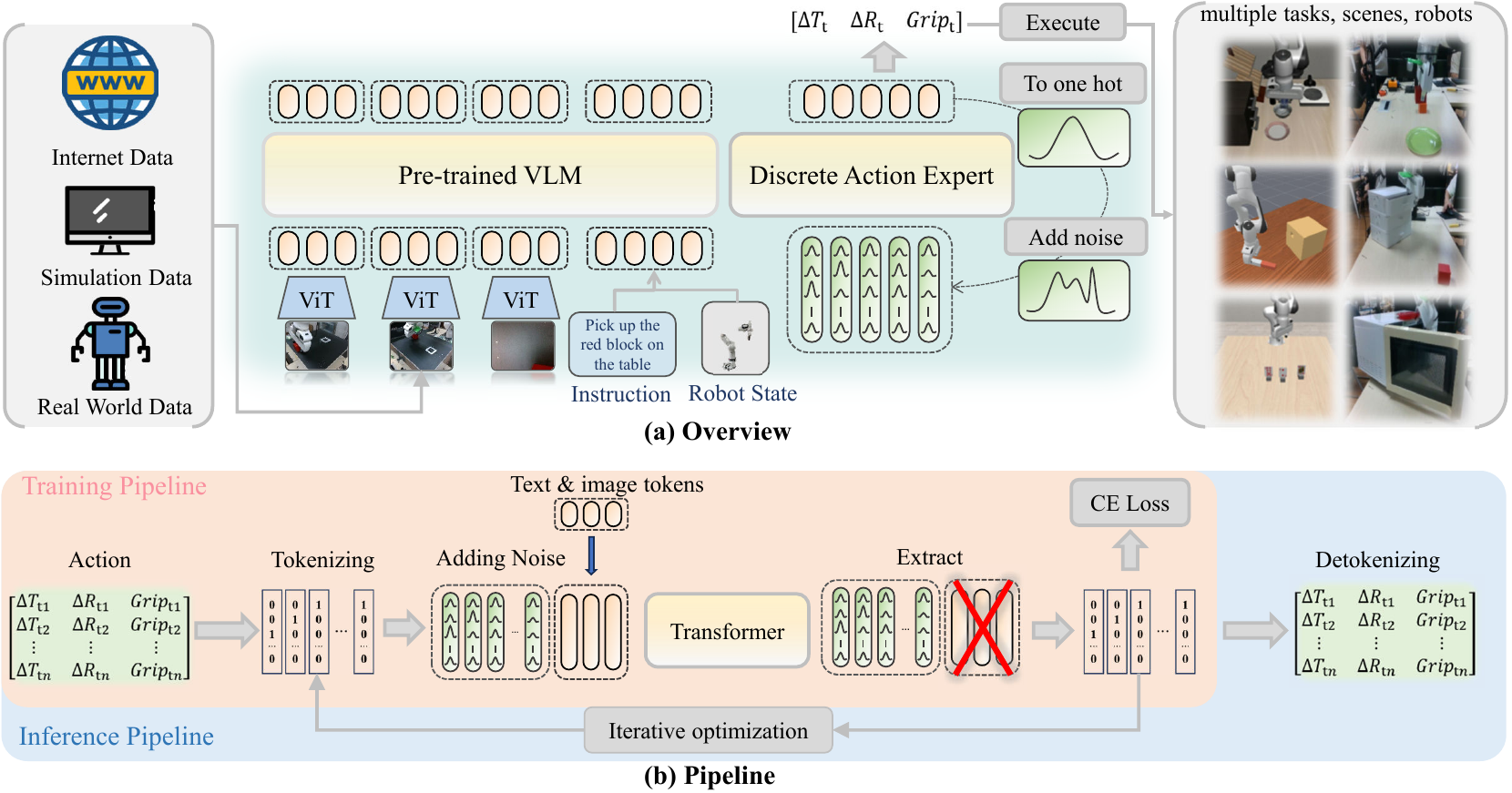}
  \vspace{-5pt}
  \caption{ 
    \textbf{Architectural illustration of E0.}
    (a) Overview of E0. 
    (b) Training and inference pipeline, showing how inputs are encoded, diffused, and decoded into executable action sequences.
    }
    \vspace{-15pt}
  \label{fig-e0}
\end{figure*}

\subsection{Preliminaries}
\noindent\textbf{Continuous-Time Flow Matching.}
Flow matching~\cite{lipman2022flow, tong2023improving} formulates generative modeling as learning a continuous-time velocity field that transports samples from a simple noise distribution to a target data distribution.
Given a clean sample $x_0$ and noise $\epsilon \sim \mathcal{N}(0, I)$, an intermediate point is constructed as $x_t = t \cdot \epsilon + (1 - t) \cdot x_0, \quad t \in [0,1]$. The model is trained to predict the velocity $u_t = \epsilon - x_0$, using a regression objective.

\noindent\textbf{Inference via ODE Integration.}
At inference time, samples are generated by solving the ordinary differential equation $\frac{dx_t}{dt} = v_\theta(x_t, t)$, from $t=1$ to $t=0$ using a numerical solver.
This process is deterministic and does not require an explicit noise variance schedule.

 \subsection{Tweedie Discrete Diffusion (TDD)}
To bridge the gap between continuous control spaces and the symbolic reasoning capabilities of pre-trained VLMs, we propose Tweedie Discrete Diffusion (TDD) as our core generative framework. TDD uniquely combines a continuous diffusion process with a discrete categorical optimization objective.  Definition: Let $x_0 \in \{e_1, ..., e_K\}$ represent discrete categorical data encoded as one-hot vectors. TDD is defined as a generative framework comprising a continuous Gaussian forward process:$$x_t = \sqrt{\alpha_t}x_0 + \sigma_t\epsilon$$where $\epsilon \sim \mathcal{N}(0, I)$. The reverse process is parameterized by a continuous-to-discrete softmax predictor $p_\theta(x_0|x_t)$. The model is optimized via the cross-entropy risk:
$$\mathcal{L} = \mathbb{E}[-\log p_\theta(x_0|x_t)]$$which, through Tweedie's identity, mathematically guarantees the optimization of the continuous score-matching objective $\nabla_{x_t} \log p_t(x_t)$ without the high-variance gradients associated with mask-based discrete diffusion. \textbf{See proof in Appendix~\ref{sec:fisher_equivalence}.}

\subsection{Architecture and Pipeline}

\noindent\textbf{VLM Backbone and Action Expert.} As shown in Figure~\ref{fig-e0}(a), we build on PaliGemma~\cite{beyer2024paligemma}, an open-source VLM, and additionally incorporate a 300M-parameter action expert as the representation backbone. Unlike traditional diffusion-based approaches~\cite{black2024pi_0, intelligence2025pi05} that map the outputs of a VLM into continuous action representations via an MLP, our method discretizes actions into bins with arbitrary granularity, while recording the corresponding statistics to enable accurate recovery of the original continuous actions. Notably, autoregressive VLAs~\cite{brohan2022rt, zitkovich2023rt, kim2024openvla} typically encode discrete actions by occupying tokens in the original vocabulary and fix the size of 256. Therefore, increasing the action resolution requires expanding the vocabulary, which incurs additional inference overhead and often necessitates retraining the model. In contrast, our method does not rely on the underlying token vocabulary: action discretization is implemented as a numerical mapping, enabling arbitrarily fine action granularity without modifying the model architecture.

Consistent with prior work~\cite{brohan2022rt, zitkovich2023rt, kim2024openvla, liang2025discrete}, we adopt a quantile-based discretization scheme (using the 1st--$M$th percentiles for each dimension to mitigate the influence of outliers). This effectively filters out extreme values and ensures smooth and stable performance during robotic inference. A single-timestep action can be represented as $D_{a} = 7$ (3 for translation, 3  for rotation, and 1 for the gripper). Alternatively, it can also be defined as $D_{a} = 8$ ( 7 for joint angles and 1 for the gripper, e.g., Franka Research 3). In our experiments, we set the number of bins to $2048$. For action chunking, tokens from $H$ (with $H=50$ in our experiments) future timesteps are arranged into a fixed layout, yielding a total of $L_{a} = H \times D_{a}$ action positions.

\noindent\textbf{Training Pipeline.}
As shown in Figure~\ref{fig-e0}(b), our objective is to model the conditional distribution $p(A_t \mid o_t)$, where $A_t = [a_t, a_{t+1}, \ldots, a_{t+H-1}]$ denotes an action chunk of horizon $H$ capturing temporal dependencies, and $o_t = [I_t^1, \ldots, I_t^n,\, l_t,\, q_t]$ represents the multimodal observation at time $t$. Here, $I_t^i$ is the $i$-th RGB image among $n$ camera views, $l_t$ encodes the language instruction, and $q_t$ denotes the robot’s proprioceptive state (e.g., joint angles). This formulation enables the model to jointly reason over heterogeneous modalities and predict coherent future actions conditioned on rich contextual cues.

Visual features $I_t^i$ and proprioceptive states $q_t$ are processed by their respective encoders and linearly projected into the same embedding space as the language tokens, yielding a unified cross-modal representation. Each action $a_t$ is discretized using the 1st--$M$th percentiles, producing a sequence of discrete tokens $\tilde{A}_t$ of length $L$, represented as one-hot vectors.

During training, a timestep $\tau \in [0,1]$ is sampled, and Gaussian noise $\varepsilon \!\sim\! \mathcal{N}(0,I)$ is added to the discretized actions:
\begin{equation}
    \tilde{A}_t^{\tau} =\tau\varepsilon+ (1-\tau )\tilde{A}_t.
\end{equation}

Since the discretized one-hot actions are inherently sharp, we apply an additional smoothing factor $\alpha=0.1$ on $\tilde{A}_t$ before adding noise, which improves training stability. Following~\cite{black2024pi_0}, we draw $\tau$ from a beta distribution instead of the uniform schedule~\cite{lipman2022flow, liu2022rectified}. This bias toward smaller $\tau$ (i.e., higher-noise regions) reflects the stronger multimodal constraints in embodied action prediction compared to image generation, encouraging the model to learn robust denoising behavior under high uncertainty.

Given the noisy action representation $\tilde{A}_t^\tau$ and observation $o_t$, the network outputs logits $v_\theta(\tilde{A}_t^\tau, o_t)$ defining a categorical distribution:
\begin{equation}
    p_\theta(A_t \mid \tilde{A}_t^\tau, o_t)
    = \mathrm{Softmax}\!\left( v_\theta(\tilde{A}_t^\tau, o_t) \right).
\end{equation}
Training minimizes the cross-entropy loss between predicted and ground-truth tokens:
\begin{small}
\begin{equation}
    \mathcal{L}_{\mathrm{CE}}(\theta)
    = -\,\mathbb{E}_{t}\!\left[
        \sum_{h=1}^{H}
        \log p_\theta\!\left(
            A_{t,h} = \tilde{A}_{t,h} \mid \tilde{A}_t^\tau, o_t
        \right)
    \right],
\end{equation}
\end{small}
which encourages accurate and temporally consistent sequence generation across the action horizon.

\noindent\textbf{Inference Pipeline.}
During inference, we initialize the action sequence from Gaussian noise and perform a multi-step iterative refinement process. At the first iteration, the observation $o_t$ is encoded once by the model to construct a key--value cache $\mathit{KV}(o_t)$ for cross-attention. This cache is reused across all subsequent refinement steps, such that only the action tokens are recomputed.

At each iteration $i \in \{1,\ldots, N\}$ (with $N=10$ in our experiments), the model takes as input the current noisy action representation $\tilde{A}_t^{\tau_i}$ together with the fixed observation cache $\mathit{KV}(o_t)$, and predicts a categorical distribution over discrete action tokens. The predicted tokens are converted into one-hot representations, which are used to construct an updated action vector for the next iteration. Noise annealing is implicitly controlled by the scalar timestep $\tau_i$, which monotonically decreases from $1$ to $0$ over refinement steps. Specifically, the next noisy action representation is obtained as:
\begin{equation}
    \tilde{A}_t^{\tau_{i+1}} 
    = \tau_{i+1}\, \varepsilon + (1-\tau_{i+1})\, \hat{A}_t^{(i)},
    \quad \varepsilon \sim \mathcal{N}(0,I),
\end{equation}
where $\varepsilon$ is fixed-variance isotropic Gaussian noise and $\hat{A}_t^{(i)}$ denotes the discrete action estimate at iteration $i$.

After $N$ refinement steps, the final discrete action tokens are deterministically detokenized into continuous actions, yielding the reconstructed action chunk $A_t = [a_{t,1}, \ldots, a_{t, H}]$. 

\textbf{The architecture details are in Appendix \ref{sec:infer_pipline_detailed}. The analysis of Preserving Physical Ordinality under Gaussian Noise Perturbations is in Appendix~\ref{sec:phy-ordinality}. The Benefits of Discrete Action Modeling are analyzed in Appendix \ref{sec:Benefits of Discrete Action Tokens}. The Benefits of Tweedie Discrete Diffusion are analyzed in Appendix \ref{sec:fisher_equivalence}.
}

\section{Experiment} \label{sec-exp}

\subsection{Training Setting}

The pretrained VLM is based on Gemma~\citep{team2024gemma}, which is configured as 
$\emph{width}=2048$, $\emph{mlp\_dim}=16384$, $\emph{depth}=18$, $\emph{num\_heads}=8$, and $\emph{head\_dim}=256$. The action expert adopts the same architecture but with a smaller configuration: $\emph{width}=1024$, $\emph{mlp\_dim}=4096$, $\emph{depth}=18$, $\emph{num\_heads}=8$, and $\emph{head\_dim}=256$. All models were fine-tuned on their corresponding datasets for 24 hours using a single NVIDIA RTX~PRO 6000 GPU. Each model was trained for a total of $30{,}000$ steps with a batch size of $32$. The learning rate followed a cosine decay schedule, implemented as \emph{CosineDecaySchedule}, with a warm-up phase of $1{,}000$ steps, a peak learning rate of $5\times10^{-5}$ and a final learning rate of $5\times10^{-6}$. The optimizer is AdamW~\citep{loshchilov2017decoupled}, configured with a gradient clipping norm of $1.0$. An exponential moving average with a decay rate of $0.999$ was applied to stabilize training. During inference, a single NVIDIA RTX~3090 or RTX PRO 6000 GPU was used for model evaluation and deployment on the server side.

While our model adopts the same backbone network as $\pi_0$ FAST~\citep{pertsch2025fast}, $\pi_0$~\citep{black2024pi_0}, and $\pi_{0.5}$~\citep{intelligence2025pi05}, the action modeling in these methods relies on autoregressive or diffusion-based action heads. Consequently, we consider them as baselines and retrain them using the same experimental setup to ensure a fair comparison and to control for variations in training settings and environments. For the remaining baselines (OpenVLA~\citep{kim2024openvla} and DiscreteDiffusionVLA~\citep{liang2025discrete}), we match the total number of training steps to that of our model to ensure a fair comparison.

\subsection{Simulation Experiment}
\begin{figure*}[t]
  \centering
  \includegraphics[width=0.8\textwidth]{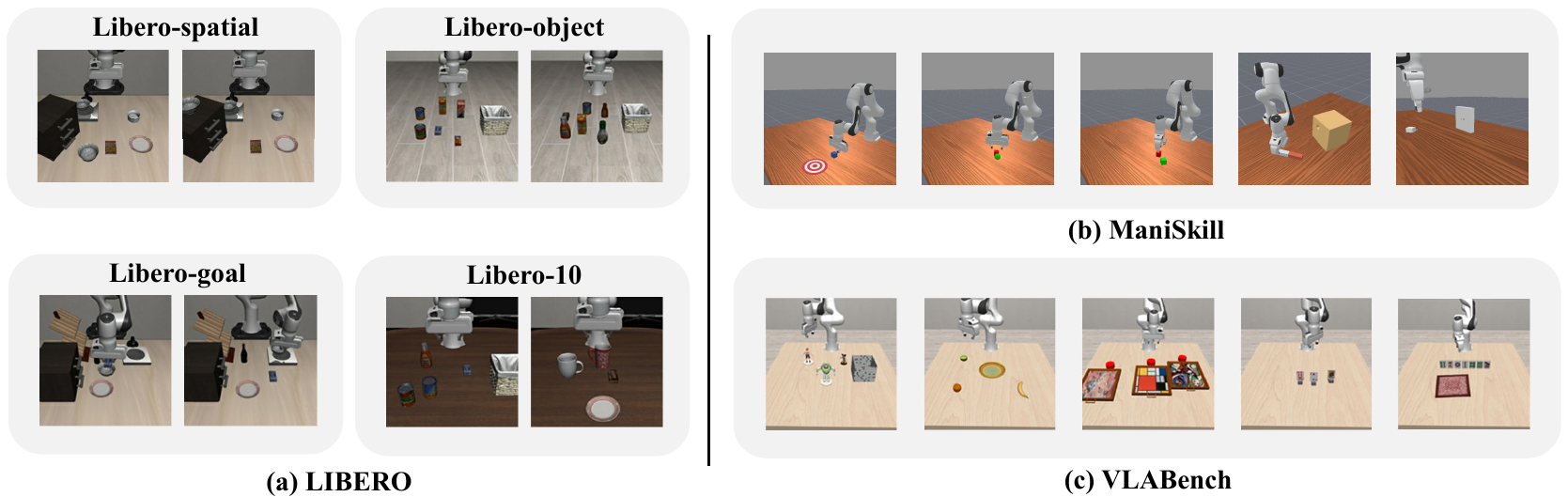}
  \caption{\textbf{Benchmarks for evaluation.} 
(a) \textbf{LIBERO}~\citep{liu2023libero}: tasks with varying objects, layouts, and goals, including long-horizon settings.  
(b) \textbf{ManiSkill}~\citep{tao2024maniskill3}: diverse fine-grained manipulation skills (push, pick, stack, insert, plug).  
(c) \textbf{VLABench}~\citep{zhang2024vlabench}: open-ended tasks requiring language grounding and commonsense reasoning (select toy/fruit/painting/poker/mahjong).}
  \label{fig-benchmark}
  \vspace{-11pt}
\end{figure*}

As shown in Figure~\ref{fig-benchmark}, we evaluate our proposed E0 on three representative simulation benchmarks. \textbf{LIBERO}~\citep{liu2023libero} provides a suite of manipulation tasks designed for benchmarking robot learning. It consists of four categories---\textit{Libero-Spatial}, \textit{Libero-Object}, \textit{Libero-Goal}, and \textit{Libero-10}. Following RDT~\cite {liu2024rdt}, we adopt a representative subset of five manipulation tasks with varying levels of difficulty---\textit{PegInsertionSide}, \textit{PickCube}, \textit{StackCube}, \textit{PlugCharger}, and \textit{PushCube}---for comparative evaluation in \textbf{ManiSkill}~\citep{tao2024maniskill3}. \textbf{VLABench}~\citep{zhang2024vlabench} is specifically designed to evaluate general-purpose robotic systems built upon large multimodal models. It emphasizes language understanding, commonsense reasoning, and multi-step task planning, covering five dimensions: grid texture perception, spatial reasoning, commonsense and world knowledge, semantic understanding, and physical laws. Additionally, we conduct supplementary experiments on 50 tasks in \textbf{RoboTwin 2.0}~\citep{chen2025robotwin}, including both single-arm and dual-arm manipulation tasks. \textbf{Further details are provided in the Appendix~\ref{sec:robotwin_detailed}.}

\begin{table*}[t]
\centering
\renewcommand{\arraystretch}{1.2}
\setlength{\tabcolsep}{5pt}
\caption{
\textbf{Performance comparison across multiple benchmarks (LIBERO, VLABench, ManiSkill).}
Each benchmark contains several task subsets, and the final column reports the \textbf{average success rate (\%)}.
A value of \textbf{0} indicates the model completely failed on this task, while “–” denotes cases where evaluation was \emph{not conducted} due to model or dataset constraints. \textbf{A “–”} in the final average column means the model was not tested on all benchmarks (direct averaging would be misleading). \textbf{Bold} and \underline{underline} numbers indicate the best and second-best results per column.
}

\resizebox{1.0\textwidth}{!}{
\begin{tabular}{c|cccc|ccccc|ccccc|c}
\toprule
\multirow{2}{*}{\textbf{Model}} &
\multicolumn{4}{c}{\textbf{LIBERO}~\citep{liu2023libero}} &
\multicolumn{5}{c}{\textbf{ManiSkill}~\citep{tao2024maniskill3}} &
\multicolumn{5}{c|}{\textbf{VLABench}~\citep{zhang2024vlabench}} &
\multirow{2}{*}{\makebox[1.0cm][c]{\textbf{Avg.}}} \\
\cmidrule(lr){2-5} \cmidrule(lr){6-10} \cmidrule(lr){11-15}
 & \makebox[1.0cm][c]{Spatial} & \makebox[1.0cm][c]{Object} & \makebox[1.0cm][c]{Goal} & \makebox[1.0cm][c]{Long} 
 & \makebox[1.0cm][c]{Insert} & \makebox[1.0cm][c]{Pick} & \makebox[1.0cm][c]{Stack} & \makebox[1.0cm][c]{Plug} & \makebox[1.0cm][c]{Push}
 & \makebox[1.0cm][c]{Toy} & \makebox[1.0cm][c]{Fruit} & \makebox[1.0cm][c]{Painting} & \makebox[1.0cm][c]{Poker} & \makebox[1.0cm][c]{Mahjong} &  \\
\midrule
\multicolumn{16}{l}{\textbf{Continuous Diffusion}} \\
\midrule
Diffusion Policy~\citep{chi2025diffusion}          & 78.3 & 92.5 & 68.3 & 50.5  & 0.0  & 40.0 & \textbf{80.0} & 0.0 & 88.0 & - & - & - & - & - & - \\
MDT~\citep{reuss2024multimodal}                    & 78.5 & 87.5 & 73.5 & 64.8 & - & - & - & - & - & - & - & - & - & - & - \\
RDT~\citep{liu2024rdt}                             & -    & -    & -    & -    & \underline{13.2} & \underline{77.2} & \underline{74.0} & \underline{1.2} & \textbf{100.0} & - & - & - & - & - & - \\
Dita~\citep{hou2025dita}                           & 84.2 & 96.3 & 85.4 & 63.8 & - & - & - & - & - & - & - & - & - & - & - \\
Octo~\citep{mees2024octo}                          & 78.9 & 85.7 & 84.6 & 51.1 & 0.0 & 0.0 & 0.0 & 0.0 & 0.0 & - & - & - & - & - & - \\
FLOWER~\citep{reuss2025flower}                     & 97.1 & 96.7 & 95.6 & 93.5 & - & - & - & - & - & - & - & - & - & - & - \\
$\pi_0$~\citep{black2024pi_0}                      & 96.8 & \underline{98.8} & 95.8 & 85.2 & 4.0 & 60.0 & 48.0 & 0.0 & \textbf{100.0} & \textbf{54.0} & \textbf{48.0} & 16.0 & 6.0 & 7.0 & 51.4 \\
$\pi_{0.5}$~\citep{intelligence2025pi05}           & \underline{97.4} & 98.4 & 97.2 & 93.6 & 8.0 & 68.0 & 72.0 & \textbf{4.0} & \textbf{100.0} & 24.0 & 18.0 & \textbf{36.0} & 20.0 & 6.5 & \underline{53.1}\\

\midrule
\multicolumn{16}{l}{\textbf{Discrete Methods}} \\
\midrule
TraceVLA~\citep{zheng2024tracevla}                 & 84.6 & 85.2 & 75.1 & 54.1 & - & - & - & - & - & - & - & - & - & - & - \\
SpatialVLA~\citep{qu2025spatialvla}                & 88.2 & 89.9 & 78.6 & 55.5 & - & - & - & - & - & - & - & - & - & - & - \\

UniVLA~\cite{wang2025unified}                      & 95.4 & \underline{98.8} & 93.6 & \underline{94.0} & - & - & - & - & - & - & - & - & - & - & - \\
Fast-dVLA~\citep{song2026fast}                     & 97.0 & \underline{98.8} & \underline{97.6} & 92.8 & - & - & - & - & - & - & - & - & - & - & - \\
OpenVLA~\citep{kim2024openvla}                     & 84.7 & 88.4 & 79.2 & 53.7 & 0.0 & 8.0 & 8.0 & 0.0 & 8.0 & 0.0 & 0.0 & 20.0 & 0.0 & 4.0 & 25.3 \\
DD-VLA~\citep{liang2025discrete}                   & 97.2 & 98.6 & 97.4 & 92.0 & 0.0 & 4.0 & 70.0 & 0.0 & \textbf{100.0} & 24.0 & 36.0 & 0.0 & 12.0 & 0.0 & 45.1 \\
$\pi_0$ FAST \citep{pertsch2025fast}               & 96.4 & 96.8 & 88.6 & 60.2 & 0.0 & \textbf{80.0} & 52.0 & 0.0 & \underline{92.0} & \underline{46.0} & \underline{42.0} & \underline{26.0} & \underline{30.0} & \textbf{20.8} & 52.2 \\
\rowcolor{gray!20}
E0 (ours) & \textbf{98.8} & \textbf{99.2} & \textbf{98.0} & \textbf{94.6} & \textbf{24.0} & 76.0 & 72.0 & \textbf{4.0} & \textbf{100.0} & \textbf{54.0} & 34.0 & 12.0 & \textbf{72.0} & \underline{18.8} & \textbf{61.2} \\
\bottomrule
\end{tabular}
}
\vspace{-11pt}
\label{tab:multi_bench_tasks}
\end{table*}

\noindent\textbf{Analysis on LIBERO.} We evaluate $\text{E}_0$ against existing methods, including continuous diffusion models (e.g., $ \pi_0 $, $\pi_{0.5}$) and other methods. To comprehensively validate the superiority of TDD over alternative discrete generation paradigms, we specifically compare against OpenVLA~\citep{kim2024openvla}, $\pi_0$ FAST, TraceVLA~\citep{zheng2024tracevla}, and the recently proposed Discrete Diffusion VLA (DD-VLA) \citep{liang2025discrete}, Fast-dVLA \citep{song2026fast}, SpatialVLA~\citep{qu2025spatialvla}. To ensure fairness, we trained $ \pi_0 $, $\pi_0$ FAST, and $\pi_{0.5}$ using the same configuration and validated them under identical settings. This benchmark includes a wide range of models, many of which employ sophisticated training strategies and architectural enhancements to maximize performance. Despite the highly competitive nature of the benchmark, our model achieved the highest average performance (97.7\%), demonstrating the effectiveness of the proposed approach even when trained with mixed data. Mask-based discrete diffusion models like DD-VLA demand massive compute to reconstruct partially corrupted token sequences via heuristic masking. When scaling DD-VLA's training configuration to align with our compute budget (120,000 steps per GPU), its performance drops significantly, highlighting the high variance and instability of masked objectives. Under comparable computational complexity, $\text{E}_0$ achieves an average success rate of 97.7\% on LIBERO, outperforming the scaled DD-VLA (60.3\%) and surpassing the fully trained Fast-dVLA (96.6\%) and official DD-VLA (96.3\%) metrics. By bypassing heuristic masking in favor of a mathematically direct Gaussian forward process over discrete spaces, TDD provides a smoother optimization landscape and superior sample efficiency. Experimental results show that our model exhibits more accurate grasping and is better at executing task instructions in some of the more complex scenarios. \textbf{Further details are provided in the Appendix~\ref{sec:libero_detailed}.}

\noindent\textbf{Analysis on ManiSkill.}
From the experimental results, we observe that in complex manipulation tasks requiring precise spatial alignment—such as peg insertion and plug-in-socket operations—$\pi_0$ and $\pi_0$ FAST perform worse than RDT, which fully exploits the fine-grained motion synthesis capability of diffusion models, enabling accurate and smooth action prediction. By integrating the strengths of both diffusion-based refinement and autoregressive conditioning, E0 effectively combines semantic understanding with precise motor control, thereby attaining the best performance across alignment-intensive tasks.
\textbf{Further details are provided in the Appendix~\ref{sec:maniskill_detailed}.}

\begin{wrapfigure}{r}{0.6\textwidth}
\vspace{-18pt}
    \centering
    \includegraphics[width=\linewidth]{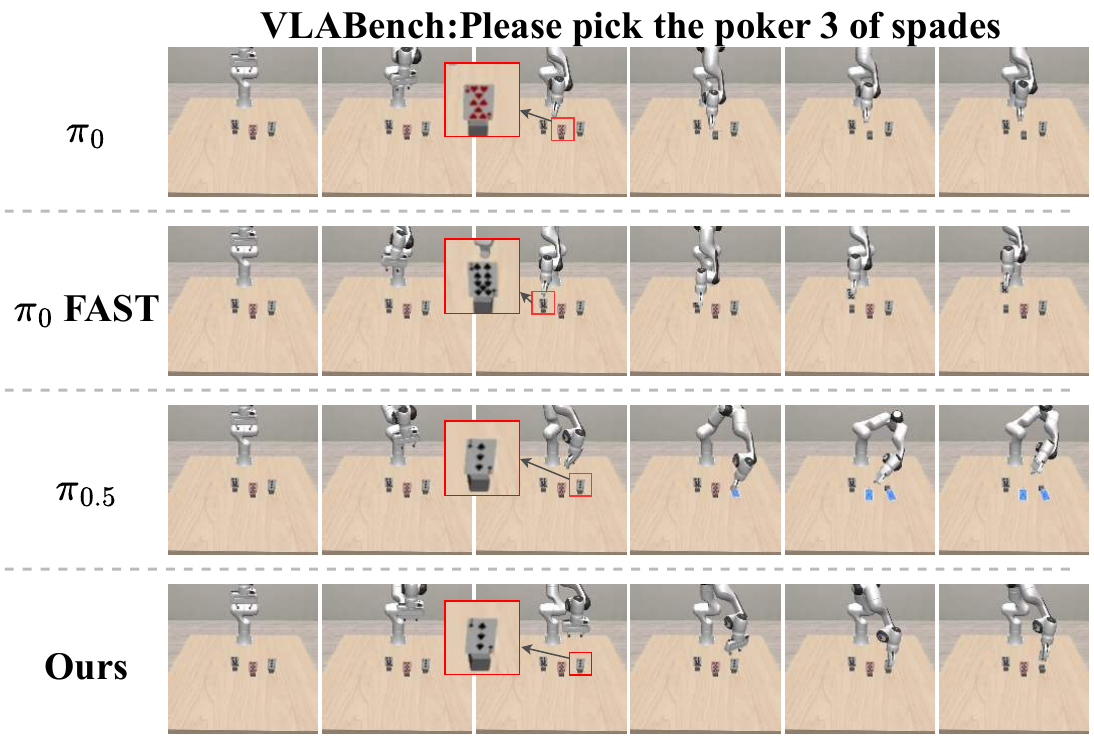}
    
    \caption{
    \textbf{Comparison on the VLABench benchmark.}
    In the task \emph{“pick up the spade~3”}, our E0 correctly identifies and precisely grasps the target card, showing superior multimodal reasoning and control ability.
    }

    \label{fig-model_compare_on_vlabench}
    \vspace{-10pt}
\end{wrapfigure}

\noindent\textbf{Analysis on VLABench.}
We observe that continuous action models exhibit noticeable performance degradation on tasks that demand explicit reasoning or fine-grained visual discrimination. This behavior is anticipated, as $\pi_0$ relies solely on action supervision during training, without incorporating explicit constraints for contextual reasoning. Consequently, its autoregressive backbone often struggles to maintain semantic coherence during inference. In contrast, $\pi_0$ FAST shows marked improvements through its autoregressive generation framework, which enhances task-level reasoning. However, it still suffers from relatively imprecise action generation. Our proposed method, E0, achieves a balanced trade-off between these paradigms. By integrating the contextual grounding capabilities of vision–language reasoning with the fine-grained control offered by discrete diffusion, it enables robust multimodal understanding and generates precise, semantically consistent action trajectories. As shown in Figure~\ref{fig-model_compare_on_vlabench}, $\pi_0$ and $\pi_0$ FAST misinterpret the textual instruction and select incorrect suits, while $\pi_{0.5}$ identifies the correct card but fails to grasp it precisely. In contrast, E0 both recognizes the correct suit and executes an accurate grasp, demonstrating strong semantic understanding and precise control. \textbf{Further details are provided in the Appendix~\ref{sec:vlabench_detailed}.}

\noindent\textbf{Computational Complexity.}
In LIBERO benchmarks, E0 achieves approximately 11.8Hz inference frequency, with only 0.0114 / 0.0215s additional latency compared to $\pi_0$ / $\pi_{0.5}$, respectively, demonstrating negligible real-time overhead.
\textbf{Further details are provided in the Appendix \ref{sec:complexity}, verifying the efficiency of our method.}

\begin{wrapfigure}{r}{0.6\textwidth}
\vspace{-22pt}
    \centering

    \begin{subfigure}[b]{\linewidth}
        \centering
            \includegraphics[width=\linewidth]{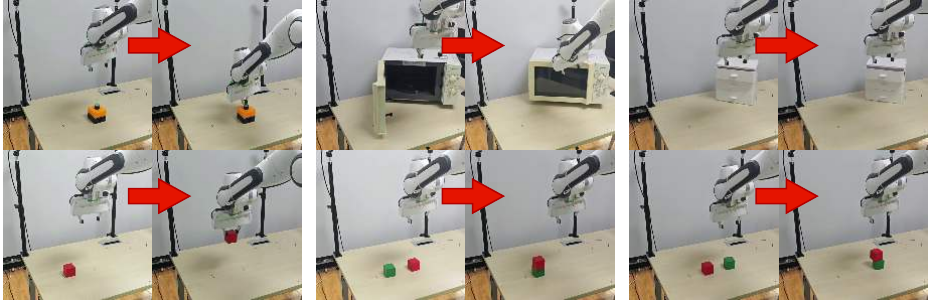}
        \caption{\textbf{Short-horizon tasks.}}
    \end{subfigure}

    \vspace{-2pt}

    \begin{subfigure}[b]{\linewidth}
        \centering
        \includegraphics[width=\linewidth]{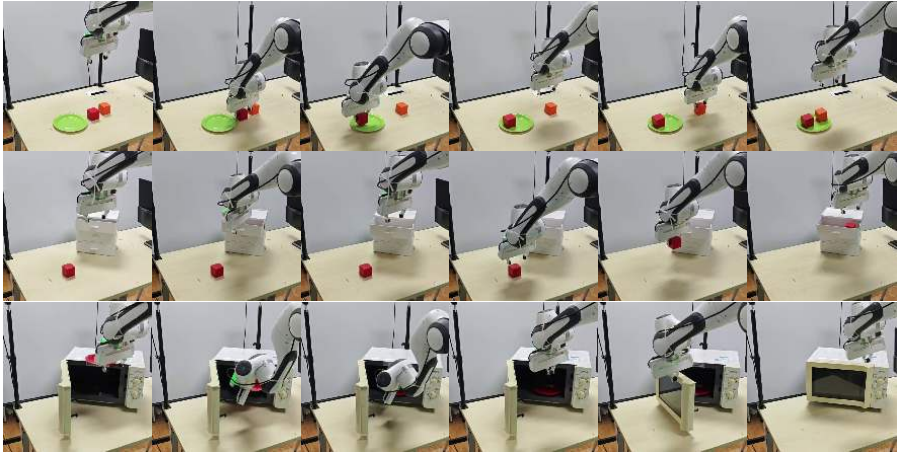}
        \caption{\textbf{Long-horizon tasks.}}
    \end{subfigure}
    
    \caption{\footnotesize
    \textbf{Performance on real-world robotic experiments.}
    (a) Short-horizon tasks (\textit{press button}, \textit{close door}, \textit{pull drawer}, \textit{pick block}, \textit{stack block}, \textit{stack block unseen}).
    (b) Long-horizon tasks (\textit{pick block twice}, \textit{pull drawer and put in block}, and \textit{put in plate and close door}).
    }

    \label{fig:real_robot_task}
    \vspace{-12pt}
\end{wrapfigure}

\subsection{Real World Experiment}

For real-world validation, we employ a Franka Research 3 robotic arm and evaluate our method across eight categories of manipulation tasks (see Figure~\ref{fig:real_robot_task}). We utilize Gello~\citep{wu2024gello} to control the Franka arm and collect demonstration data. For the short-horizon tasks, we collect 50 trajectories per task, while for the long-horizon tasks, we collect 80 trajectories each. We train our model on a mixture of these trajectories and evaluate its performance by executing 20 trials per task (For E0 and $\pi_{0.5}$, we conduct an additional evaluation round and calculate the average SR). As shown in the Table~\ref{tab:realworld_results}, our model achieves the highest average success rate across all tasks. These results demonstrate E0's capability to maintain semantic coherence and precise control across extended action sequences. Moreover, we observe impressive generalization to previously unseen configurations. For instance, in the \textit{stack block} task, even when the positions of the red and green blocks are swapped—differing from the training data—the model accurately identifies their colors and completes the stacking correctly (see Figure~\ref{fig:real_robot_task}). Similarly, in the \textit{pick block twice} task, while the green plate and the red and orange blocks were fixed during data collection, the model successfully handles varying placements during validation, demonstrating robust generalization and adaptability.
\textbf{Further details are provided in the Appendix~\ref{sec:hardware_detailed}, Appendix~\ref{sec:real_detailed}, Appendix~\ref{sec:real_more_generalization_detailed} and Appendix~\ref{sec:real_instruction_detailed}.}

\begin{table*}[!htbp]
\centering
\vspace{-3pt}
\caption{
\textbf{Task success rates in real-world robot experiments.}
We evaluate policy performance across both short-horizon and long-horizon manipulation tasks. For the $\pi_{0.5}$ and E0, we conduct an additional evaluation run and report the average results across the two runs.
\textbf{Bold} numbers indicate the best performance in each column.
}
\label{tab:realworld_results}
\resizebox{0.9\textwidth}{!}{
\begin{tabular}{c|ccccc|ccc|c}
\toprule
\multirow{3}{*}{\textbf{Model}} &
\multicolumn{5}{c|}{\textbf{Short-horizon Tasks}} &
\multicolumn{3}{c|}{\textbf{Long-horizon Tasks}} &
\multirow{2}{*}{\textbf{Average}} \\ 
\cmidrule(lr){2-6} \cmidrule(lr){7-9}
& Pick & Press & Stack & Pull & Close & Pick Twice & Open \& Put & Put \& Close &  \\ 
& SR (\%) $\uparrow$ & SR (\%) $\uparrow$ & SR (\%) $\uparrow$ & SR (\%) $\uparrow$ & SR (\%) $\uparrow$ 
& SR (\%) $\uparrow$ & SR (\%) $\uparrow$ & SR (\%) $\uparrow$ & SR (\%) $\uparrow$ \\
\midrule
$\pi_0$ FAST~\cite{pertsch2025fast} & 20.0 & 20.0 & 5.0 & 10.0 & 15.0 & 10.0 & 0.0 & 0.0 & 10.0 \\
$\pi_0$~\cite{black2024pi_0} & 60.0 & 60.0 & 40.0 & 30.0 & \textbf{45.0} & 45.0 & \textbf{55.0} & 10.0 & 43.1 \\
$\pi_{0.5}$~\cite{intelligence2025pi05} & 65.0 & 47.5 & \textbf{55.0} & \textbf{42.5} & 40.0 & 47.5 & 50.0 & 15.0 & 45.3 \\
\midrule
\rowcolor{gray!20}
E0 (ours) & \textbf{75.0} & \textbf{70.0} & 52.5 & 32.5 & 35.0 & \textbf{60.0} & 40.0 & \textbf{20.0} & \textbf{48.1} \\
\bottomrule
\end{tabular}}
\vspace{-8pt}
\end{table*}

\begin{figure*}[h]
  \centering
  \includegraphics[width=0.95\textwidth]{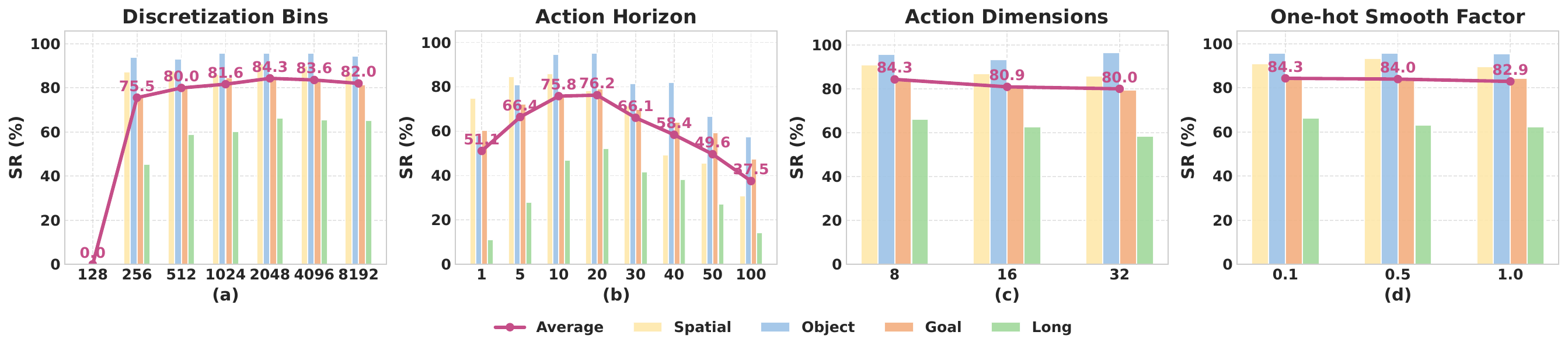}
  \vspace{-5pt}
  \caption{
  \textbf{Comprehensive ablation analysis of key hyperparameters in the LIBERO environments.}
We investigate four crucial design factors influencing our model: (a) \textit{Discretization bins}—increasing bin granularity enhances precision up to 2048 bins, beyond which gains saturate; (b) \textit{Action horizon}—a moderate prediction length balances reactivity and temporal consistency; (c) \textit{Action dimensions}—configured action-slot width, where the smallest sufficient width achieves the highest average success rate without redundant padded positions; and (d) \textit{One-hot smooth factor}—moderate decay values smooth discrete logits, stabilizing diffusion and improving overall success rate.}
\vspace{-11pt}
  \label{fig-ablation}
\end{figure*}

\subsection{Ablation Study}

We perform an extensive ablation study to examine the impact of key hyperparameters in the LIBERO environments. As shown in Figure~\ref{fig-ablation}, finer action discretization improves control precision up to a moderate resolution, beyond which the enlarged vocabulary introduces optimization noise with limited gain. Moderate one-hot decay before noise injection further stabilizes denoising and improves success rates by encouraging exploration of nearby actions (see Figure~\ref{fig-ablation}). Besides, to disentangle the effect of discretization from the proposed Tweedie Discrete Diffusion framework, we trained $\pi_0$ and $\pi_{0.5}$ with continuous regression on one-hot targets and observed low success rates across four tasks, suggesting that the gains are not solely due to discretization. 

Figure~\ref{fig-ablation}(c) investigates the effect of configured action-slot width rather than the physical action dimensionality. On LIBERO, the 8-slot configuration achieves the highest average success rate of 84.3\%, compared with 80.9\% and 80.0\% for 16 and 32 slots, respectively. All configurations execute the same physical action channels, while larger widths introduce redundant padded positions. These results suggest that using the smallest sufficient action-slot width is preferable for a fixed embodiment. \textbf{Further details are provided in the Appendix~\ref{sec:ablation_detailed}.}

Action chunking, which predicts short future sequences, effectively balances responsiveness and temporal consistency. Longer horizons initially enhance smoothness but eventually reduce adaptability due to accumulated open-loop errors, with 10–20 steps yielding the best trade-off. Finally, quantile-based normalization outperforms standard mean–std scaling by handling heavy-tailed action distributions, leading to more stable and robust performance across environments (see Table~\ref{tab-ablation-stat}).

\subsection{Action Multimodality and Prediction Precision.}
We investigate the influence of action multimodality using 1,693 LIBERO demonstration episodes. Across 10,000 query states, the normalized distance between the conditional mean and its nearest empirical action mode increases by 46.29\% from low- to high-multimodality groups. We further compare E0 with a training-matched $\pi_{0.5}$ baseline on 3,000 identical observations. E0 consistently achieves lower nearest-mode errors across all three multimodality groups, reducing the overall error by 6.43\%. It also reduces the off-mode rate by 27.41\% and target-action ADE by 31.44\%, while improving demonstration-consistent mode selection by 9.13 percentage points. These findings demonstrate improved mode consistency and action precision under matched evaluation conditions. \textbf{Detailed analyses are provided in Appendix~\ref{app:multimodality} and Appendix~\ref{app:multimodality2}.}

\subsection{Empirical Training Diagnostics.}
We track cross-entropy (CE), a categorical Fisher--Rao proxy, and a time-scaled score-error proxy throughout training. Between the first and last logged steps, these quantities decrease by 96.95\%, 93.53\%, and 98.11\%, respectively. Early- and late-window averages also show reductions in CE and the Fisher--Rao proxy. These measurements document their empirical training behavior and complement the categorical-to-score analysis. \textbf{Detailed results are provided in Appendix~\ref{app:fisher_empirical}.}

\begin{wrapfigure}{r}{0.6\textwidth}
\vspace{-22pt}
    \centering
    \includegraphics[width=\linewidth]{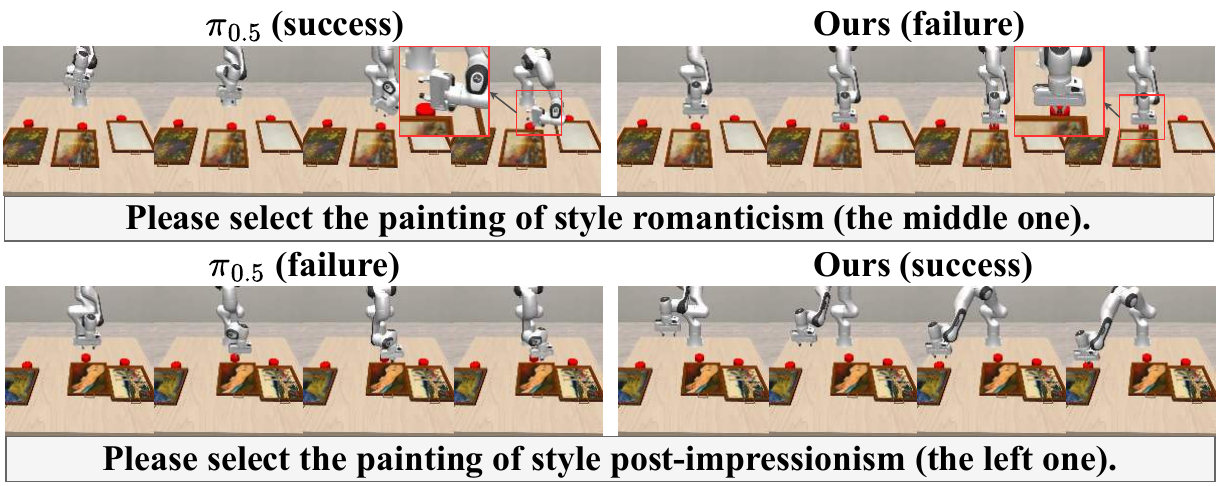}
    
    \caption{\footnotesize
    \textbf{Comparison on the VLABench \textit{Select Painting} task.}
    E0 shows more consistent visual--language grounding and smoother control than $\pi_{0.5}$.
    }
    
    \label{fig-vlabench_painting_compare_pi05}
    \vspace{-10pt}
\end{wrapfigure}

\subsection{Case Analysis}
Within \textbf{VLABench}, the \textit{Select Painting} is where our E0 performs weakest (see Table~\ref{tab:multi_bench_tasks}). To investigate the cause, we analyzed all simulated trajectories and compared them with $\pi_{0.5}$, the strongest baseline on this task. We found that most models fail to capture the task semantics: instead of selecting the painting matching the textual description, they tend to press the \textbf{middle button} by default, ignoring left or right options. To examine the behavior underlying the baseline's aggregate success rate, we inspected all 50 evaluation rollouts of $\pi_{0.5}$. The correct target was located on the left, in the center, and on the right in 9, 31, and 10 trials, respectively. Center-directed behavior was observed in all 50 rollouts. All 18 successful trials occurred when the correct target was in the center, while none of the 19 trials with targets on either side succeeded. These observations are consistent with a strong positional bias in the evaluated rollouts.

As shown in Figure~\ref{fig-vlabench_painting_compare_pi05}, E0 exhibits more meaningful interaction by attempting to align visual and textual cues, occasionally moving toward the correct button, though sometimes failing due to limited painting-related data and partial \textit{catastrophic forgetting} during fine-tuning. Notably, E0 produces smoother, more stable motions than $\pi_{0.5}$, which often acts erratically despite achieving higher success ``by chance.'' Overall, E0 maintains strong control quality, and future work will focus on improving its visual–language grounding to enhance task understanding. \textbf{Further details are provided in the Appendix~\ref{sec:vlabench_detailed}.}

\section{Conclusion} \label{sec-con}

In this work, we revisit action modeling in VLA systems from the perspective of three structural properties: multi-peaked action distributions, the token-based representations of pretrained VLMs, and the finite resolution of real-world robotic control.
Motivated by these properties, we introduce E0, a Tweedie Discrete Diffusion framework for fine-grained action generation through categorical denoising and iterative refinement.
Our theoretical analysis establishes a connection between categorical prediction and continuous score estimation, while empirical studies reveal that increasing action multimodality is associated with greater deviations between conditional-mean predictions and valid action modes.
Controlled comparisons further demonstrate that E0 achieves lower off-mode prediction rates, improved mode consistency, and more accurate action predictions than continuous-flow baselines.
Extensive evaluations across simulated benchmarks and real-world manipulation tasks demonstrate the effectiveness of E0, highlighting discrete diffusion as a promising approach to precise and generalizable robotic action generation.

\clearpage

\begin{ack}

\paragraph{Funding Statement.} This work was supported in part by National High-Level Young Talent Program (Grant 2025HY00260104), in part by the Fundamental Research Funds for Higher Education Institutions allocated to Sun Yat-sen University (Grant 25hytd007), in part by the Guangdong Provincial High-Level Young Talent Program (Grant 2025HYSPT0707), in part by the Tuoyuan (Grant HT-99982025-0564), in part by the Faculty Start-up Research Fund (Grant 67000-12255002), in part by the Huawei Strategic Research Institute Talent Fund, and in part by the Key Development Project of the Artificial Intelligence Institute of Sun Yat-sen University (Grant 2025RGZN009).

\paragraph{Competing Interests Disclosure.} None of the authors have financial relationships with entities that could potentially be perceived to influence the submitted work during the last 36 months prior to this submission.

\end{ack}

\bibliographystyle{plainnat}
\bibliography{neurips_2026} 

\clearpage
\appendix

\titlecontents{appsection}
  [2.2em]
  {\addvspace{8pt}\normalfont}
  {\contentslabel{2.2em}}
  {}
  {}

\titlecontents{appsubsection}
  [4.8em]
  {\addvspace{3pt}\normalfont}
  {\contentslabel{2.6em}}
  {}
  {}

\startcontents[appendix]
\section*{\centering Appendix contents}
\printcontents[appendix]{app}{1}[2]{}

\clearpage
\clearpage

\section{Architecture and Pipeline} \label{sec:infer_pipline_detailed}

During inference, we initialize the action sequence from Gaussian noise and perform a multi-step iterative refinement procedure. At the first iteration, the observation $o_t$ is encoded once by the model to construct a key--value cache $\mathit{KV}(o_t)$ for cross-attention. This cache is reused across all subsequent refinement steps, such that only the action tokens are recomputed.

We discretize a continuous refinement variable $\tau \in [1,0]$ into $N$  (with $N=10$ in our experiments) steps using a uniform schedule, i.e., $\tau_i = 1 - \frac{i-1}{N}$ for $i=1,\ldots, N$. At each iteration $i$, the current noisy action representation $\tilde{A}_t^{\tau_i}$ is passed into the model together with the fixed cache $\mathit{KV}(o_t)$. The model produces logits $v_\theta(\tilde{A}_t^{\tau_i}, \mathit{KV}(o_t))$, which define a categorical distribution over discrete action tokens:
\begin{equation}
    p_\theta(A_t \mid \tilde{A}_t^{\tau_i}, o_t)
    = \mathrm{Softmax}\!\left( v_\theta(\tilde{A}_t^{\tau_i}, \mathit{KV}(o_t)) \right).
\end{equation}

The predicted categorical distribution is decoded into discrete token indices via $\arg\max$ decoding,
and each token is converted into a one-hot vector in a $B$-dimensional discrete action space:
\begin{equation}
    \mathrm{onehot}(k) \in \{0,1\}^B,
    \quad \mathrm{onehot}(k)_j =
    \begin{cases}
        1, & j = k, \\
        0, & j \neq k.
    \end{cases}
\end{equation}
The resulting one-hot vectors form the intermediate action representation $\hat{A}_t^{(i)}$.

To obtain the input for the next refinement step, we apply an implicit noise annealing scheme controlled by $\tau_{i+1}$:
\begin{equation}
    \tilde{A}_t^{\tau_{i+1}}
    = \tau_{i+1}\, \varepsilon + (1 - \tau_{i+1})\, \hat{A}_t^{(i)},
    \quad \varepsilon \sim \mathcal{N}(0,I),
\end{equation}
where $\varepsilon$ is fixed-variance isotropic Gaussian noise.
As $\tau$ decreases monotonically, the signal-to-noise ratio increases, yielding progressively more deterministic action representations without introducing an explicit variance schedule.

After $N$ refinement steps, the final predicted token indices $\{k_h\}_{h=1}^H$ are deterministically mapped back into continuous actions by inverting the discretization:
\begin{equation}
    a_{t,h} = \mathrm{Detokenize}(k_h), \quad h = 1,\ldots,H.
\end{equation}

In practice, following~\cite{black2024pi_0,intelligence2025pi05}, we precompute per-dimension action statistics and normalize all action values into the $[0,1]$ interval prior to discretization. This preprocessing enables efficient quantile-based discretization, where actions can be easily encoded into discrete tokens and inverse-mapped back to the original continuous action space through the corresponding rescaling operation. Collecting across the horizon yields the reconstructed action chunk:
\begin{equation}
    A_t = [a_{t,1}, a_{t,2}, \ldots, a_{t,H}].
\end{equation}

\section{Complexity and Performance} \label{sec:complexity}

E0 does not introduce additional iterative overhead compared to our primary baselines ($\pi_0$~\cite{black2024pi_0} and $\pi_{0.5}$~\cite{intelligence2025pi05} ), and the latency remains fully suitable for real-time control.
We can clarify the inference complexity through the following points:

\textbf{Apples-to-Apples Comparison:} Baselines like $\pi_0$~\cite{black2024pi_0} and $\pi_{0.5}$~\cite{intelligence2025pi05}  are not single-pass AR models for action generation. They employ a hybrid architecture consisting of an autoregressive VLM backbone paired with a continuous diffusion-based action head. Crucially, they also require iterative denoising during inference (typically 10 steps). Our method simply replaces this continuous diffusion head with a discrete diffusion variant, keeping the number of denoising steps identical.

\textbf{Empirical Justification: }As established in our real-world and simulated latency benchmarks, replacing the continuous diffusion head with our discrete variant yields highly comparable inference overhead. The model is hosted on a server and communicates with the real-world robot via a wireless network. Because network transmission rates introduce high variance, measuring real-world latency yields unstable and inaccurate reflections of the model's true forward-pass speed.
To provide the exact, isolated inference frequency, we conducted extensive, controlled simulation experiments. We measured the exact time from the client transmitting data to the model outputting the final action chunk (excluding robotic execution time).
E0 operates at approximately 11.8Hz, adding only a fractional delay. Average inference time is only 0.0114 / 0.0215s longer than $\pi_0$~\cite{black2024pi_0} and $\pi_{0.5}$~\cite{intelligence2025pi05}. This confirms that the denoising iterations do not bottleneck real-time closed-loop control. (See Table~\ref{tab:efficiency})

\begin{table*}[ht]
\centering
\caption{
\textbf{Efficiency and inference statistics on LIBERO tasks.} 
We report the number of inferences ($n$), total inference time ($t$, in seconds), average control frequency (Hz), and overall success rate (SR).
}
\label{tab:efficiency}
\resizebox{0.9\textwidth}{!}{
\begin{tabular}{c|cc|cc|cc|cc|c|c}
\toprule
\multirow{2}{*}{\textbf{Model}} 
& \multicolumn{2}{c|}{\textbf{Spatial}} 
& \multicolumn{2}{c|}{\textbf{Object}} 
& \multicolumn{2}{c|}{\textbf{Goal}} 
& \multicolumn{2}{c|}{\textbf{Long}} 
& \multirow{2}{*}{\textbf{Hz} }
& \multirow{2}{*}{\textbf{SR (\%)} $\uparrow$} \\
& $n$ & $t$ & $n$ & $t$ & $n$ & $t$ & $n$ & $t$ & & \\
\midrule
$\pi_0$~\cite{black2024pi_0} 
& 3024 & 226.426 & 3827 & 278.521 & 3302 & 239.643 & 7620 & 556.740 & 13.659 & 94.15 \\

$\pi_{0.5}$~\cite{intelligence2025pi05} 
& 5769 & 374.863 & 7232 & 452.431 & 6018 & 378.104 & 14403 & 902.264 & 15.858 & 96.65 \\

\rowcolor{gray!20}
E0 (ours) 
& 5457 & 474.433 & 7064 & 597.437 & 5876 & 497.646 & 13641 & 1141.340 & 11.819 & \textbf{97.65} \\

\bottomrule
\end{tabular}}
\end{table*}

Regarding peak GPU memory usage, we utilize the JAX/XLA framework for both training and inference. Because JAX pre-allocates GPU memory during program initialization to optimize XLA execution, memory usage remains strictly constant during runtime. In our experimental setup, loading the model for closed-loop inference requires a stable footprint of 24 GB VRAM, which is the same setup for both E0 and $\pi_0$ / $\pi_{0.5}$.

\section{Preserving Physical Ordinality under Gaussian Noise Perturbations} \label{sec:phy-ordinality}

While it is true that raw one-hot vectors are equidistant, the network successfully recovers physical ordinality through representation learning. We can directly defend this design choice using proven theoretical guarantees from recent literature:

\textbf{(1) Theoretical Guarantees from ``Regression as Classification"}
The precise phenomenon of how orthogonal categorical bins can capture continuous geometry better than direct regression—was recently mathematically proven in \cite{stewart2023regression}.
The authors provide theoretical proofs demonstrating that reformulating continuous regression as a discrete classification task fundamentally changes the implicit bias of the neural network. Even though the cross-entropy loss over orthogonal one-hot vectors has no explicit notion of distance,
\cite{stewart2023regression} mathematically proves that this formulation drives the network to learn highly localized, distinct feature representations in its latent space. Because physically adjacent actions share highly overlapping contextual data, these localized features naturally organize to perfectly recover the underlying continuous geometry. Conversely, they prove that continuous regression (MSE) suffers from theoretical optimization barriers that force the network into learning overly smoothed, global features, strictly limiting fine-grained precision.

\textbf{(2) Theoretical Bounds in Action Spaces}
Furthermore, in the context of action modeling, \cite{tang2020discretizing} provides a theoretical analysis showing that discretizing continuous spaces into independent categorical bins mathematically circumvents the restrictive parametric priors of continuous distributions (like Gaussians). They prove that this orthogonal formulation allows the model to express arbitrarily complex, multimodal action distributions while maintaining favorable gradient variance—something inherently impossible when regressing to a single continuous scalar.

\textbf{(3) High-Precision via Strict Penalties}
Building on this theory, treating the bins as strictly orthogonal is actually a deliberate feature that drives robotic precision. In continuous regression, predicting an action that is "close" incurs only a small penalty, leading to the theoretically proven over-smoothing. Formulating the space as independent categories imposes a strict, high penalty for any incorrect bin, forcing the network to capture exact, high-frequency details. This is precisely why discrete binning paradigms consistently achieve state-of-the-art fine-grained control in modern VLA models. Beyond the VLA domains, treating discretized continuous values as independent, orthogonal categories is a well-established architectural paradigm with strong empirical backing. A foundational precedent is PixelRNN~\cite{van2016pixel} and PixelCNN~\cite{van2016conditional}.

\section{Detailed Ablation Analysis}  \label{sec:ablation_detailed}

\noindent\textbf{Disentangling discretization and diffusion.} A common question is whether the observed performance improvements stem from the proposed Tweedie Discrete Diffusion framework, or merely from the underlying action discretization scheme.

To answer the question directly, we modified $\pi_0$ and $\pi_{0.5}$ to train via continuous regression (MSE loss) to predict ground-truth one-hot vectors, applying a softmax only at the final inference step. The results validate our framework:

\begin{itemize}
    \item \textbf{Quantitative failure} The success rate plummeted to nearly 0\% across all four evaluated tasks on LIBERO~\cite{liu2023libero}.

    \item \textbf{Qualitative observations} In simulation videos, the policies exhibited wildly exaggerated motion amplitudes and completely diverged from the correct trajectories. This confirms that regressing to one-hot vectors is fundamentally unsuitable for fine-grained motion modeling.

    \item \textbf{Theoretical intuition} This catastrophic drop in performance is intuitive when viewed through the lens of classification. Predicting discrete tokenized bins is inherently a classification problem, not a continuous regression one. Just as one would not use MSE loss to train a standard image classifier, using continuous regression to predict categorical one-hot vectors fails to properly capture the underlying probability distribution at each denoising step.    
\end{itemize}

This experiment explicitly confirms that our performance gains do not merely come from discretizing the space, but are fundamentally driven by optimizing the intermediate denoisings via cross-entropy loss under the Tweedie Discrete Diffusion framework.

\noindent\textbf{Effect of Discretization Bins.} A key advantage of our model is its ability to flexibly discretize continuous control signals into an arbitrary number of bins, enabling a fine-grained trade-off between numerical precision and representation compactness. To evaluate this, we systematically vary the discretization resolution from 128 to 8192 bins (see Table~\ref{tab-ablation-bins}). At low resolutions (e.g., 128 bins), the model fails to learn meaningful trajectories due to the inability of coarse quantization to accurately encode joint values. This results in erratic and unstable robot motions. As the bin count increases, control accuracy improves, with performance saturating around 2048 bins. At this point, the discretization error becomes negligible compared to diffusion noise, offering sufficient precision for smooth and stable control. However, further increasing the resolution to 4096 or 8192 bins yields only marginal gains or even slight degradation in performance. This is attributed to the larger token vocabulary and increased noise, which complicate optimization. Unlike conventional autoregressive VLA models limited to 256 discrete bins, our method supports significantly higher-resolution action representations, enabling superior control fidelity and precision.

\begin{table*}[htbp]
\centering
\caption{
\textbf{Ablation study on discretization bins in the LIBERO environments.} 
We evaluate different numbers of discretization bins used to discretize continuous action values. Unlike autoregressive VLAs that typically fix the bin size to 256, our model supports arbitrary discretization granularity. Performance improves with higher resolution and peaks at 2048 bins, indicating that this level of precision is sufficient for accurate action representation. \textbf{Bold} numbers mark the best performance per column.
}
\label{tab-ablation-bins}
\resizebox{0.8\textwidth}{!}{
\begin{tabular}{c|cccc|c}
\toprule
\multirow{2}{*}{\textbf{bins}} & \textbf{Libero-Spatial} & \textbf{Libero-Object} & \textbf{Libero-Goal} & \textbf{Libero-Long} & \textbf{Average} \\
& SR (\%) $\uparrow$ & SR (\%) $\uparrow$ & SR (\%) $\uparrow$ & SR (\%) $\uparrow$ & SR (\%) $\uparrow$ \\
\midrule
128              & 0.0           & 0.0           & 0.0           & 0.0           &  0.00 \\
256              & 87.2          & 93.8          & 75.8          & 45.4          & 75.55 \\
512              & 87.8          & 93.0          & 80.2          & 58.8          & 79.95 \\
1024             & 86.2          & 95.6          & \textbf{84.6} & 60.2          & 81.65 \\
2048             & \textbf{90.8} & \textbf{95.8} & 84.4          & \textbf{66.2} & \textbf{84.30} \\
4096             & 89.8          & \textbf{95.8} & 83.2          & 65.4          & 83.55 \\
8192             & 87.0          & 94.4          & 81.4          & 65.2          & 82.00 \\
\bottomrule
\end{tabular}}

\end{table*}

\noindent\textbf{Effect of Configured Action-Slot Width.} We investigate the effect of the model's configured action-slot width $D_{\mathrm{slot}}\in\{8,16,32\}$ on LIBERO. The physical action dimensionality remains unchanged across all configurations. The 8-slot setting is the smallest supported interface that accommodates the original action channels. For wider configurations, the same actions are zero-padded to the specified width, and only the original physical channels are used for robot execution.

As shown in Table~\ref{tab-ablation-dims}, the 8-slot configuration achieves the highest average success rate of 84.3\%, while increasing the width to 16 and 32 reduces the average success rate to 80.9\% and 80.0\%, respectively. Although larger interfaces support heterogeneous embodiments, they introduce additional action positions without adding physical control information. These redundant positions may complicate denoising and increase computational overhead. The results indicate that matching the configured width to the smallest sufficient action interface is preferable for single-embodiment training.

\begin{table*}[htbp]
\centering
\caption{
\textbf{Effect of configured action-slot width on LIBERO.} The physical action dimensions remain fixed, while the model's internal action-slot width $D_{\mathrm{slot}}$ varies. Larger configurations introduce zero-padded positions without changing the executed control channels. \textbf{Bold} numbers mark the best performance per column.
}
\label{tab-ablation-dims}
\resizebox{0.8\textwidth}{!}{
\begin{tabular}{c|cccc|c}
\toprule
\multirow{2}{*}{\textbf{dims}} & \textbf{Libero-Spatial} & \textbf{Libero-Object} & \textbf{Libero-Goal} & \textbf{Libero-Long} & \textbf{Average} \\
& SR (\%) $\uparrow$ & SR (\%) $\uparrow$ & SR (\%) $\uparrow$ & SR (\%) $\uparrow$ & SR (\%) $\uparrow$ \\
\midrule
8               & \textbf{90.8} & 95.8          & \textbf{84.4} & \textbf{66.2} & \textbf{84.30} \\
16              & 86.8          & 93.4          & 80.8          & 62.6          & 80.90 \\
32              & 85.8          & \textbf{96.4} & 79.4          & 58.4          & 80.00 \\
\bottomrule
\end{tabular}}
\end{table*}

\noindent\textbf{Effect of One-hot Smooth Factor.} Prior to injecting noise into the one-hot action representation, we apply a one-hot smooth factor to scale the vector. This smoothing operation attenuates the sharp peaks of the one-hot distribution, thereby reducing the dominance of high-confidence logits and enabling smoother transitions within the action space during diffusion. As shown in Table~\ref{tab-ablation-decay}, 0.1 achieves the highest average success rate. Stronger attenuation encourages the model to better explore neighboring discrete actions and prevents overfitting to rigid one-hot boundaries. Conversely, higher smooth factor values retain excessively sharp logits, which hinder denoising stability and degrade generalization performance. These findings indicate that moderately suppressing one-hot amplitude prior to noise injection enhances the robustness of policy learning under diffusion-based action modeling.

\begin{table*}[htbp]
\centering
\caption{
\textbf{Ablation study on one-hot decay in the LIBERO environments.}
We examine the influence of the decay coefficient applied to one-hot action vectors prior to noise addition. This decay attenuates the sharp peaks of discrete logits, reducing the dominance of high-confidence tokens during diffusion. Smaller decay values lead to improved stability and higher overall success rates. \textbf{Bold} numbers mark the best performance per column.
}
\label{tab-ablation-decay}
\resizebox{0.8\textwidth}{!}{
\begin{tabular}{c|cccc|c}
\toprule
\multirow{2}{*}{\textbf{decay}} & \textbf{Libero-Spatial} & \textbf{Libero-Object} & \textbf{Libero-Goal} & \textbf{Libero-Long} & \textbf{Average} \\
& SR (\%) $\uparrow$ & SR (\%) $\uparrow$ & SR (\%) $\uparrow$ & SR (\%) $\uparrow$ & SR (\%) $\uparrow$ \\
\midrule
0.1               & 90.8          & \textbf{95.8} & \textbf{84.4} & \textbf{66.2} & \textbf{84.30} \\
0.5               & \textbf{93.2} & \textbf{95.8} & 83.6          & 63.2          & 83.95 \\
1.0               & 89.6          & 95.4          & 84.2          & 62.4          & 82.90 \\
\bottomrule
\end{tabular}}

\end{table*}

\noindent\textbf{Effect of Action Chunk.} We adopt an action chunking strategy, wherein the policy outputs a short sequence of future actions conditioned on the current observation. This approach has shown particular effectiveness in real-world imitation learning and VLA models (e.g.,~\cite{chi2025diffusion, liu2024rdt, black2024pi_0}). By predicting multiple future actions simultaneously, the policy captures implicit temporal context, thereby mitigating non-Markovian ambiguities such as pauses, delayed responses, or lingering motions—without requiring additional historical inputs that may introduce causal confusion.

Overlapping action chunks enable temporal averaging across consecutive predictions, yielding smoother and more stable control signals. Additionally, this method allows the model to operate at a lower inference frequency while generating high-frequency motor commands—a key advantage for high-precision manipulation under constrained computational budgets. Furthermore, predicting a sequence of look-ahead actions effectively compensates for inference latency during real-world deployment, promoting consistent system dynamics and responsive robot control.

As shown in Table~\ref{tab-ablation-action-steps}, increasing the action horizon initially improves performance by enhancing temporal coherence and execution smoothness. However, overly long prediction horizons degrade performance across all evaluated environments, as they reduce reactivity to environmental changes and increase compounding errors from open-loop execution. In contrast, shorter horizons (10–20 steps) provide an optimal balance between short-term responsiveness and long-term planning stability, resulting in the highest overall success rates. These findings validate our choice of a short-horizon action chunking strategy for real-world robotic deployment, where responsiveness and motion stability are equally critical.

\begin{table*}[htbp]
\centering
\caption{
\textbf{Ablation study on the number of action horizons in the LIBERO environments.}
We analyze how the length of the predicted action chunk affects the success rate across four tasks. A moderate horizon yields the highest average success rate, while both very short and very long horizons lead to degraded performance, indicating a trade-off between reactivity and temporal consistency. \textbf{Bold} numbers mark the best performance per column.
}
\label{tab-ablation-action-steps}
\resizebox{0.8\textwidth}{!}{
\begin{tabular}{c|cccc|c}
\toprule
\multirow{2}{*}{\textbf{Action Horizon}} & \textbf{Libero-Spatial} & \textbf{Libero-Object} & \textbf{Libero-Goal} & \textbf{Libero-Long} & \textbf{Average} \\
& SR (\%) $\uparrow$ & SR (\%) $\uparrow$ & SR (\%) $\uparrow$ & SR (\%) $\uparrow$ & SR (\%) $\uparrow$ \\
\midrule
1                 & 74.8          & 58.0          & 60.4          & 11.2          & 51.10 \\
5                 & 84.6          & 80.8          & 72.2          & 28.0          & 66.40 \\
10                & \textbf{86.0} & 94.6          & 75.6          & 47.0          & 75.80  \\
20                & 78.4          & \textbf{95.2} & \textbf{79.2} & \textbf{52.2} & \textbf{76.25} \\
30                & 71.4          & 81.4          & 69.8          & 41.6          & 66.05 \\
40                & 49.4          & 82.0          & 64.0          & 38.2          & 58.40 \\
50                & 45.6          & 66.6          & 59.2          & 27.2          & 49.65 \\
100               & 31.0          & 57.6          & 47.4          & 14.2          & 37.55 \\

\bottomrule
\end{tabular}}
\end{table*}

\noindent\textbf{Effect of Data Normalization.} We compare two normalization techniques for processing continuous action vectors within our model. Both methods apply element-wise normalization based on statistical parameters (e.g., mean and standard deviation) computed independently for each action dimension across the entire dataset.

\begin{equation}
\label{eq:normalize}
\tilde{x} = \frac{x - \mu}{\sigma + 10^{-6}},
\end{equation}
where $\mu \in \mathbb{R}^d$ and $\sigma \in \mathbb{R}^d$ denote the mean and standard deviation of each action dimension, respectively. 
This is the standard mean–std normalization that assumes a Gaussian-like distribution.

\begin{equation}
\label{eq:quantile_norm}
\tilde{x} = \frac{(x - q_{0.01})}{(q_{0.99} - q_{0.01}) + 10^{-6}} \times 2 - 1,
\end{equation}
where $q_{0.01}$ and $q_{0.99}$ represent the 1st and 99th percentiles of the empirical action distribution for each dimension. This quantile-based normalization scales values into $[-1, 1]$ and reduces sensitivity to outliers or heavy-tailed distributions.

Both normalization parameters $(\mu, \sigma, q_{0.01}, q_{0.99})$ are computed independently along each action dimension to accurately reflect the heterogeneous range and variance of robot control signals. As shown in Table~\ref{tab-ablation-stat}, quantile normalization offers consistently improved stability and robustness across various LIBERO environments, making it a more suitable choice for diffusion-based policy learning in settings with diverse and non-uniform control signal distributions.

\begin{table*}[htbp]
\centering
\caption{
\textbf{Ablation study on normalization statistics in LIBERO environments .} 
Comparison between mean–std normalization and quantile-based normalization. \textbf{Bold} numbers mark the best performance per column.
}
\label{tab-ablation-stat}
\resizebox{0.8\textwidth}{!}{
\begin{tabular}{c|cccc|c}
\toprule
\multirow{2}{*}{\textbf{Normalization Type}} & \textbf{Libero-Spatial} & \textbf{Libero-Object} & \textbf{Libero-Goal} & \textbf{Libero-Long} & \textbf{Average} \\
& SR (\%) $\uparrow$ & SR (\%) $\uparrow$ & SR (\%) $\uparrow$ & SR (\%) $\uparrow$ & SR (\%) $\uparrow$ \\
\midrule
Mean–Std          & 13.2          & 0.0           & 9.2           & 7.8           & 7.6   \\
Quantile          & \textbf{90.8} & \textbf{95.8} & \textbf{84.4} & \textbf{66.2} & \textbf{84.3} \\
\bottomrule
\end{tabular}}
\end{table*}

\begin{figure*}[h]
  \centering
  \includegraphics[width=0.8\textwidth]{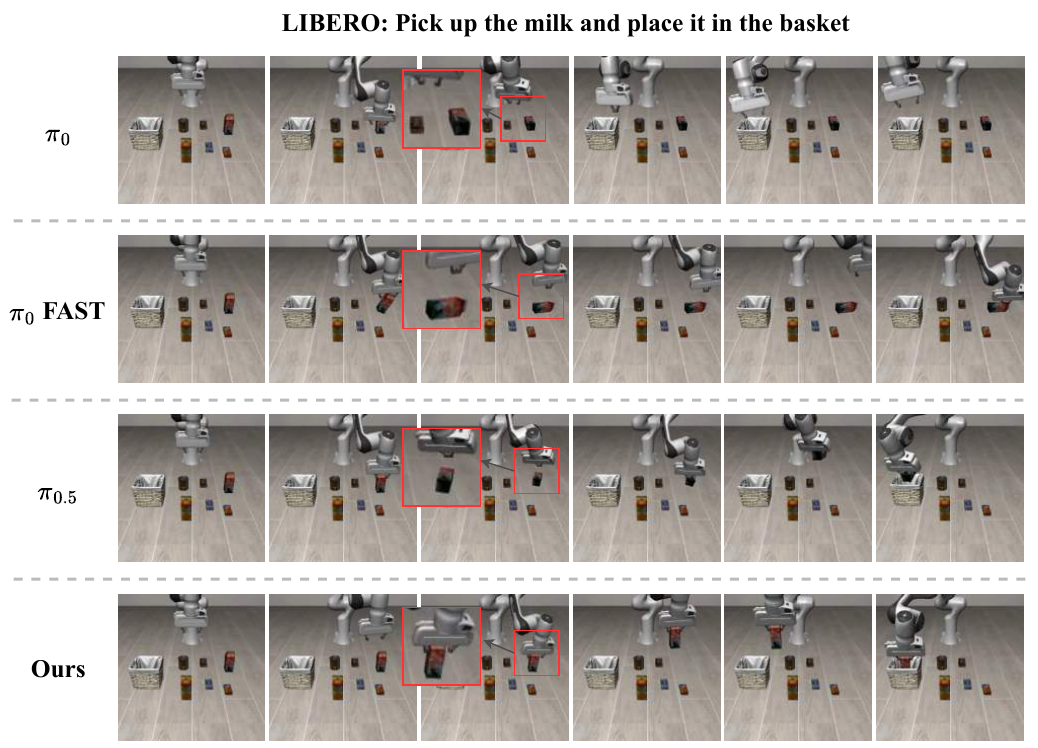}
  \caption{
  \textbf{Comparison on the LIBERO benchmark.}
  In the task \textit{pick up the milk and place it in the basket}, our E0 successfully grasped the object with the correct posture and completed the task without knocking it over.
  }
      \label{fig-model_compare_on_libero}
\end{figure*}

\section{More Details on LIBERO} \label{sec:libero_detailed}

To evaluate the generalization behavior of our model under diverse distribution shifts, we adopt \textbf{LIBERO}~\cite{liu2023libero} as a comprehensive benchmark for robot manipulation. LIBERO provides 130 language-conditioned tasks grouped into four procedurally generated suites (\textit{spatial}, \textit{object}, \textit{goal}, and \textit{libero-100}), each targeting a different dimension of knowledge transfer, including spatial reasoning, object concepts, behavioral goals, and mixed declarative–procedural knowledge. These tasks are created through a scalable generation pipeline that samples human-activity-inspired templates, constructs scene layouts, and specifies PDDL-based goal predicates, resulting in highly diverse manipulation scenarios with controllable variations in object types, spatial layouts, and goal semantics.

Following the common practice in prior work  (use \textit{libero-10} instead of \textit{libero-100})  and consistent with the configurations adopted by most existing models~\cite{black2024pi_0, kim2024openvla,pertsch2025fast}, \textbf{we train a single unified policy across all LIBERO tasks, instead of training task-specific models}. All training is performed using a single RTX PRO 6000 GPU over 30,000 steps. Each task provides 50 high-quality teleoperated demonstrations, enabling sample-efficient behavioral cloning and ensuring that performance differences primarily reflect the model’s ability to transfer and retain knowledge across tasks.

As illustrated in Figure~\ref{fig-model_compare_on_libero}, the \textit{pick up the milk and place it in the basket} task reveals clear behavioral differences among the baseline models and our proposed E0. Both $\pi_0$ and $\pi_0$-FAST frequently approached the target object with suboptimal gripper orientation, often making lateral contact with the milk carton, which destabilized the object and occasionally knocked it over during the grasping phase. Although $\pi_{0.5}$ demonstrated slightly improved task semantics, it still exhibited inconsistent end-effector alignment, leading to partial or unstable grasps.

In contrast, our E0 consistently produced a precise, well-aligned grasp, approaching the milk carton with the correct wrist rotation and a controlled closing trajectory. This allowed the policy to lift the object cleanly without disturbance and reliably place it into the basket. Such stable behavior suggests that E0 learns more accurate contact priors, smoother approach trajectories, and better-conditioned fine-motor actions compared to previous policies. The improvement also highlights E0’s enhanced robustness in visually cluttered scenes and its ability to preserve fine-grained manipulation correctness, even when trained jointly across many heterogeneous LIBERO tasks.

Although the LIBERO benchmark has already been pushed to very high success rates by many recent models (see Table~\ref{tab-libero_results}), it remains one of the most widely adopted and reliable platforms for stress-testing manipulation policies, particularly for evaluating robustness, motion stability, and semantic correctness under diverse visual and task variations. Despite the overall high performance saturation, our E0 still exhibits clear qualitative and quantitative advantages, demonstrating smoother behaviors, better grasp stability, and more precise action execution than strong baselines.

\begin{table*}[htbp]
\centering
\caption{
\textbf{LIBERO task performance (success rate, SR) results.}
Evaluation across four subsets—\textit{Spatial}, \textit{Object}, \textit{Goal}, and \textit{Long-horizon}. Diffusion-based and autoregressive VLAs are compared, showing that large-scale generalist models significantly improve robustness and our proposed E0 achieves the highest average SR, demonstrating strong cross-task generalization and stability. Furthermore, we evaluated E0 using three independent runs under identical settings. \textbf{Bold} numbers denote the best results per column.
}
\label{tab-libero_results}
\resizebox{0.8\textwidth}{!}{
\begin{tabular}{c|cccc|c}
\toprule
\multirow{2}{*}{\textbf{Model}} & \textbf{Libero-Spatial} & \textbf{Libero-Object} & \textbf{Libero-Goal} & \textbf{Libero-Long} & \textbf{Average} \\
& SR (\%) $\uparrow$ & SR (\%) $\uparrow$ & SR (\%) $\uparrow$ & SR (\%) $\uparrow$ & SR (\%) $\uparrow$ \\
\midrule
Diffusion Policy~\cite{chi2025diffusion} & 78.3 & 92.5 & 68.3 & 50.5 & 72.4 \\
MDT~\cite{reuss2024multimodal}                        & 78.5 & 87.5 & 73.5 & 64.8 & 76.1 \\
\midrule
OpenVLA~\cite{kim2024openvla}                         & 84.7 & 88.4 & 79.2 & 53.7 & 76.5 \\
Octo \cite{mees2024octo}                              & 78.9 & 85.7 & 84.6 & 51.1 & 75.1 \\
Dita~\cite{hou2025dita}                               & 84.2 & 96.3 & 85.4 & 63.8 & 82.4 \\
TraceVLA~\cite{zheng2024tracevla}                     & 84.6 & 85.2 & 75.1 & 54.1 & 74.8 \\
SpatialVLA~\cite{qu2025spatialvla}                    & 88.2 & 89.9 & 78.6 & 55.5 & 78.1 \\
UniVLA~\cite{wang2025unified}                         & 95.4 & 98.8 & 93.6 & 94.0 & 95.5 \\
FLOWER~\cite{reuss2025flower}                         & 97.1 & 96.7 & 95.6 & 93.5 & 95.7 \\
DiscreteDiffusionVLA~\cite{liang2025discrete}         & 97.2 & 98.6 & 97.4 & 92.0 & 96.3 \\
Fast-dVLA~\cite{song2026fast}                         & 97.0 & 98.8 & 97.6 & 92.8 & 96.6 \\
$\pi_0$ FAST \cite{pertsch2025fast}                   & 96.4 & 96.8 & 88.6 & 60.2 & 85.5 \\
$\pi_0$~\cite{black2024pi_0}                          & 96.8 & 98.8 & 95.8 & 85.2 & 94.2 \\
$\pi_{0.5}$~\cite{intelligence2025pi05}               & 97.4 & 98.4 & 97.2 & 93.6 & 96.7 \\
\midrule
\rowcolor{gray!20}
\raisebox{3pt}{E0 (ours)} & \shortstack{ 98.60 \\[-2pt] {\small\textit{$\pm$ 0.72}}}  & \shortstack{ 99.60 \\[-2pt] {\small\textit{$\pm$ 0.20}}} & \shortstack{ 97.87 \\[-2pt] {\small\textit{$\pm$ 0.23}}} & \shortstack{ 94.20 \\[-2pt] {\small\textit{$\pm$ 1.40}}} & \shortstack{ 97.57 \\[-2pt] {\small\textit{$\pm$ 0.56}}} \\
\bottomrule
\end{tabular}}

\end{table*}

\section{More Details on ManiSkill} \label{sec:maniskill_detailed}

To further assess the generalization capability and fine-grained manipulation performance of our model across diverse operational scenarios, we additionally conduct experiments on \textbf{ManiSkill}~\cite{tao2024maniskill3}. ManiSkill is an advanced GPU–parallelized simulation and benchmarking framework for embodied AI, offering one of the most comprehensive suites of manipulation tasks to date. It covers 12 distinct categories, ranging from tabletop manipulation and dexterous hand control to room-scale mobile manipulation, bimanual coordination, and high-fidelity digital twin environments. Its highly efficient GPU-based simulation and rendering pipeline enables large-scale visual policy training on a single GPU with minimal memory overhead, making it a suitable platform for evaluating both generalization and visual robustness.

Following the experimental design used in RDT~\cite{liu2024rdt}, \textbf{we leverage the official demonstration dataset (5 tasks with a total of 5,000 trajectories) and retrain all baseline models under identical data conditions to ensure fair comparison}. All training is performed on a single RTX PRO 6000 GPU for 30,000 steps, using the same evaluation protocol across models. This setup allows us to rigorously compare their representational capacity, generalization behavior, and fine-grained manipulation performance within a high-fidelity simulation environment. For the remaining baselines (OpenVLA~\citep{kim2024openvla} and DiscreteDiffusionVLA~\citep{liang2025discrete}), we match the total number of training steps to that of our model to ensure a fair comparison.

\begin{figure*}[t]
  \centering
  \includegraphics[width=0.8\textwidth]{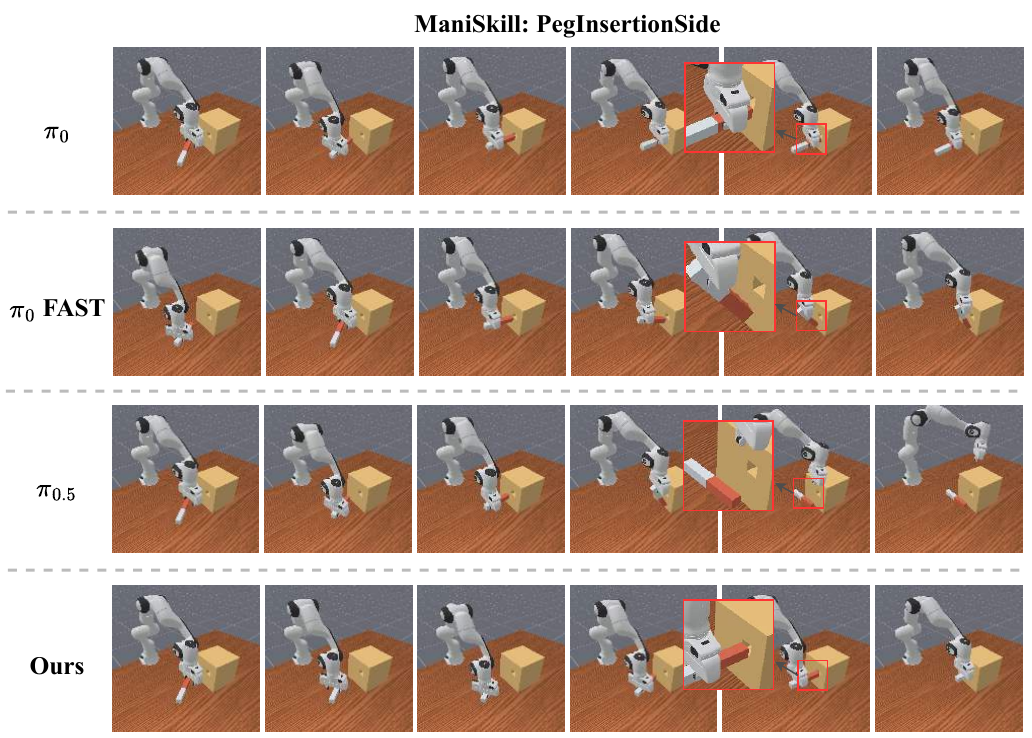}
  \caption{
      \textbf{Comparison on the ManiSkill benchmark.} 
      In the task \textit{PegInsertionSide}, our E0 achieved the best performance in this highly dexterous task. It was able to precisely align the peg with the hole and successfully insert it, while other models performed poorly.
  }
  \label{fig-model_compare_on_maniskill}
\end{figure*}

As shown in Figure~\ref{fig-model_compare_on_maniskill}, the \textit{PegInsertionSide} task highlights clear behavioral differences between our model and baseline models. Both $\pi_0$ and $\pi_0$-FAST struggle to produce stable insertion trajectories: their end-effector often approaches the peg with incorrect orientation, leading to repeated collisions with the box surface instead of aligning with the hole. Although $\pi_{0.5}$ demonstrates slightly improved spatial reasoning, its execution remains unstable, showing drifting motions and failed insertion attempts.

In contrast, our E0 exhibits highly reliable and precise insertion behavior. Across the entire rollout sequence, E0 maintains a well-controlled approach trajectory, accurately aligns the peg with the hole, and executes a smooth and decisive insertion. This qualitative advantage—clear from the consistent orientation control and absence of collision-induced jitter suggests that E0 learns more stable, contact-aware action patterns and benefits significantly from our discrete diffusion–based action representation.

These visual differences are mirrored by the quantitative results in Table~\ref{tab-maniskill_results}. Across five challenging manipulation tasks—\textit{PegInsertionSide}, \textit{PickCube}, \textit{StackCube}, \textit{PlugCharger}, and \textit{PushCube}—our model achieves the highest average success rate (55.2\%), highlighting its balanced capability in precision manipulation, geometry-aware reasoning, and stable interactions.

Overall, the combined qualitative and quantitative evidence demonstrates that E0 achieves state-of-the-art generalization and stability on ManiSkill, and remains robust even in tasks where many of the strongest existing policies collapse. This validates the effectiveness of our discrete diffusion design in capturing both fine motor control and long-horizon structural dependencies in complex manipulation environments.

\begin{table*}[htbp]
\centering
\caption{
\textbf{ManiSkill task performance (success rate, SR) results.}
Evaluation on five challenging manipulation tasks—\textit{PegInsertionSide}, \textit{PickCube}, \textit{StackCube}, \textit{PlugCharger}, and \textit{PushCube}—from the ManiSkill benchmark. Compared with both diffusion-based and autoregressive VLAs, our proposed E0 achieves the highest average SR. While prior models often fail on precise insertion or long-horizon coordination, E0 maintains stable performance, highlighting the effectiveness of discrete diffusion-based action representation. \textbf{Bold} numbers denote the best results per column.
}
\label{tab-maniskill_results}
\resizebox{0.8\textwidth}{!}{
\begin{tabular}{c|ccccc|c}
\toprule
\multirow{2}{*}{\textbf{Model}} & \textbf{PegInsertionSide} & \textbf{PickCube} & \textbf{StackCube} & \textbf{PlugCharger} & \textbf{PushCube} & \textbf{Average} \\
& SR (\%) $\uparrow$ & SR (\%) $\uparrow$ & SR (\%) $\uparrow$ & SR (\%) $\uparrow$ & SR (\%) $\uparrow$ & SR (\%) $\uparrow$ \\
\midrule
Diffusion Policy~\cite{chi2025diffusion}   & 0.0  & 40.0 & \textbf{80.0} & 0.0  & 88.0  & 41.6 \\
OpenVLA~\cite{kim2024openvla}              & 0.0  & 8.0  & 8.0  & 0.0  & 8.0   & 4.8  \\
Octo~\cite{mees2024octo}                   & 0.0  & 0.0  & 0.0  & 0.0  & 0.0   & 0.0  \\
DD-VLA~\cite{liang2025discrete}               & 0.0  & 4.0  & 70.0  & 0.0  & \textbf{100.0 }  & 34.8 \\
RDT~\cite{liu2024rdt}                      & 13.2 & 77.2 & 74.0 & 1.2  & \textbf{100.0} & 53.1 \\
$\pi_0$ FAST~\cite{pertsch2025fast}        & 0.0  & \textbf{80.0} & 52.0 & 0.0  & 92.0  & 44.8 \\
$\pi_0$~\cite{black2024pi_0}               & 4.0  & 60.0 & 48.0 & 0.0  & \textbf{100.0} & 42.4 \\
$\pi_{0.5}$~\cite{intelligence2025pi05}    & 8.0  & 68.0 & 72.0 & \textbf{4.0}  & \textbf{100.0}  & 50.4 \\
\midrule
\rowcolor{gray!20}
E0 (ours)                                  & \textbf{24.0} & 76.0 & 72.0 & \textbf{4.0}  & \textbf{100.0} & \textbf{55.2} \\
\bottomrule
\end{tabular}}

\end{table*}

\section{More Details on VLABench} \label{sec:vlabench_detailed}

To further assess the robustness, semantic understanding, and generalization capability of our model across diverse language-conditioned manipulation scenarios, we additionally evaluate it on \textbf{VLABench}~\cite{zhang2024vlabench}. VLABench offers a broad and heterogeneous collection of tasks that span various object categories, spatial arrangements, and visually complex environments. Each task is paired with natural language instructions that require accurate semantic grounding, attribute-level understanding, and precise alignment between linguistic cues and visual observations. The benchmark’s extensive variability in object layouts, appearance conditions, and linguistic expressions makes it a rigorous platform for evaluating multimodal reasoning and visuomotor robustness.

Following common practice in prior work and to ensure fair comparison, \textbf{we train a single unified model rather than training separate models for individual tasks}. \textbf{We adopt the official VLABench dataset and retrain all baseline methods under identical data and computational settings}. All experiments are conducted on a single RTX PRO 6000 GPU for 30,000 training steps, ensuring consistent training dynamics across models. For the remaining baselines (OpenVLA~\citep{kim2024openvla} and DiscreteDiffusionVLA~\citep{liang2025discrete}), we match the total number of training steps to that of our model to ensure a fair comparison.

As shown in Table~\ref{tab-vlabench_sr_results}, our model achieves the highest overall success rate on VLABench, outperforming a variety of baseline approaches across multiple language-guided manipulation tasks. These results indicate that our discrete-diffusion action representation provides stronger semantic grounding, more stable visuomotor behavior, and improved generalization across diverse linguistic and visual conditions.

In accordance with the official VLABench protocol, we conduct mixed training using the full dataset released by the benchmark. A noteworthy aspect of this dataset is that it contains a large number of trajectories unrelated to the evaluation tasks. This increases the difficulty of generalization: models lacking sufficient robustness may overfit to these irrelevant instruction–action pairs, ultimately reducing performance on the actual evaluation tasks. As a result, VLABench serves as a particularly challenging benchmark for evaluating semantic alignment, instruction following, and cross-task transfer.

We inspected all 50 evaluation rollouts of $\pi_{0.5}$ and grouped their outcomes by the ground-truth target location. Table~\ref{tab:painting_position_breakdown} reports the number of trials and successful task completions for each group. The correct target appeared on the left in 9 trials, in the center in 31 trials, and on the right in 10 trials. Trajectory inspection revealed center-directed behavior in all 50 rollouts.

All 18 successful trials belonged to the group with a center target, yielding a conditional task success rate of 58.1\%. No trial with a target on either side succeeded. However, 13 trials with a center target also failed, showing that center-directed behavior was insufficient for successful completion. This breakdown documents a positional bias in the evaluated rollouts without identifying the separate contributions of language conditioning, visual recognition, and motion execution.

\begin{figure*}[t]
  \centering
  \includegraphics[width=0.8\textwidth]{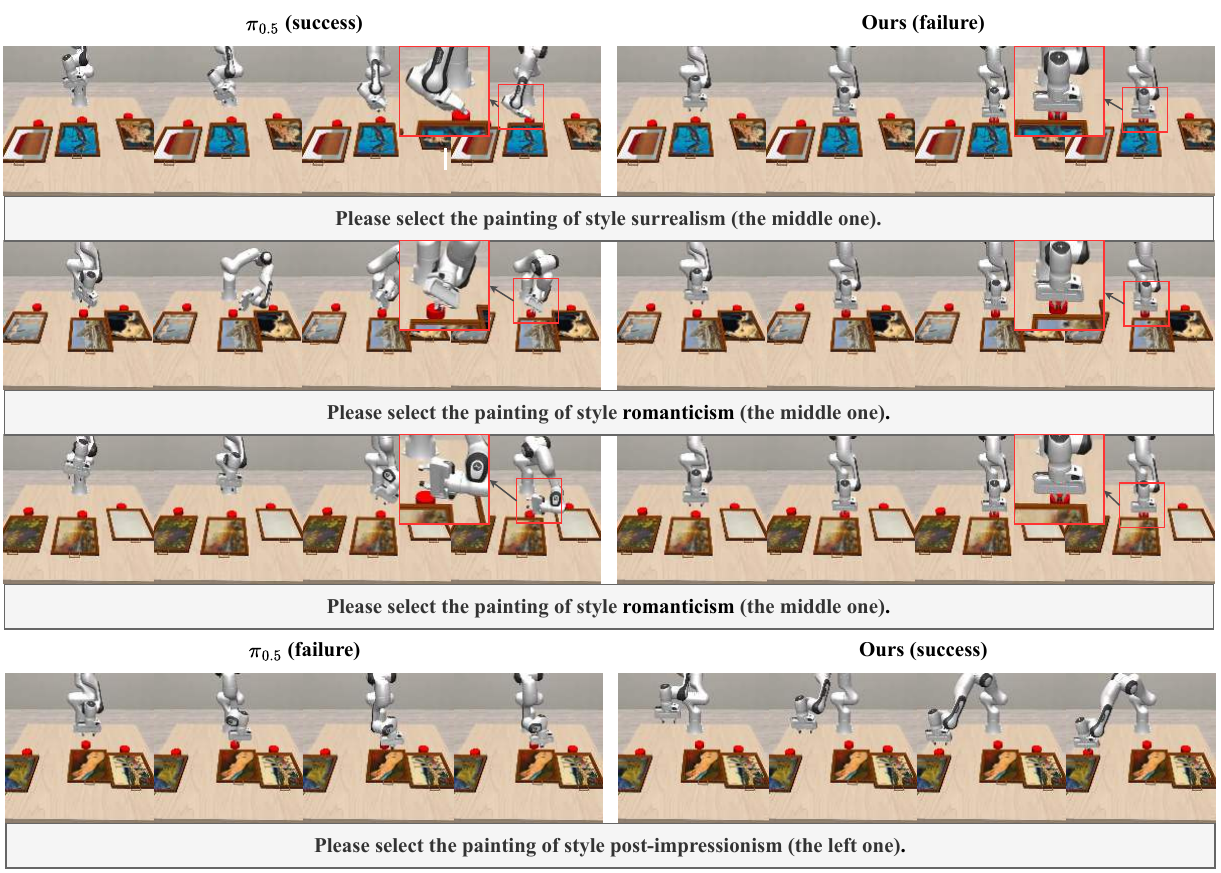}
  \caption{
      \textbf{Additional qualitative comparisons on the VLABench \textit{select painting} task.} This figure extends the results presented in Figure~\ref{fig-vlabench_painting_compare_pi05}, showcasing additional examples and outcomes.
  }
  \label{fig-more-vlabench-painting}
\end{figure*}

Figure~\ref{fig-more-vlabench-painting} provides additional qualitative examples for $\pi_{0.5}$ and E0. These examples complement the aggregate task results. E0's 12\% success rate on Select Painting remains a performance limitation, and the available examples do not establish an aggregate advantage in semantic selection.

In addition to reporting the success rate (Table~\ref{tab-vlabench_sr_results}), we also include the process score (Table~\ref{tab-vlabench_ps_results}), a complementary metric that measures whether the policy executes the correct intermediate reasoning steps even when the final prediction is annotated as failure. Our E0 achieves the highest average PS across all five evaluation tasks, indicating more accurate step-by-step action grounding and stronger adherence to instruction semantics throughout the execution sequence. Together, these results highlight the robustness, compositional understanding, and semantic reliability of our model, even in the presence of ambiguous labels and visually confounding selection tasks.

\begin{table}[h]
    \centering
    \small
    \caption{
    \textbf{Target-position breakdown on Select Painting.} We report $\pi_{0.5}$'s task outcomes over 50 evaluation rollouts, grouped by the ground-truth target location. Success rate is computed within each row. Center-directed behavior was observed in all 50 rollouts, but this observation does not imply that the center button was successfully pressed in every trial.
    }
    \label{tab:painting_position_breakdown}

    \small
    \setlength{\tabcolsep}{8pt}
    \renewcommand{\arraystretch}{1.15}

    \begin{tabular}{@{}lrrr@{}}
        \toprule
        Target location & Trials & Successes & SR (\%) \\
        \midrule
        Left   &  9 &  0 &  0.0 \\
        Center & 31 & 18 & 58.1 \\
        Right  & 10 &  0 &  0.0 \\
        \midrule
        Overall & 50 & 18 & 36.0 \\
        \bottomrule
    \end{tabular}
\end{table}

\begin{table*}[htbp]
\centering
\caption{
\textbf{VLABench task performance (success rate, SR) results.} Evaluation on five language-conditioned visual reasoning tasks from VLABench, including \textit{Select Toy}, \textit{Select Fruit}, \textit{Select Painting}, \textit{Select Poker}, and \textit{Select Mahjong}. These tasks require fine-grained visual grounding and action understanding under compositional instructions. Compared to baseline models, our proposed E0 achieves the highest average success rate, showing stronger instruction following and cross-domain transferability. \textbf{Bold} numbers denote the best results in each column.
}
\label{tab-vlabench_sr_results}
\resizebox{0.8\textwidth}{!}{
\begin{tabular}{c|ccccc|c}
\toprule
\multirow{2}{*}{\textbf{Model}} & \textbf{Select Toy} & \textbf{Select Fruit} & \textbf{Select Painting} & \textbf{Select Poker} & \textbf{Select Mahjong} & \textbf{Average} \\
& SR (\%) $\uparrow$ & SR (\%) $\uparrow$ & SR (\%) $\uparrow$ & SR (\%) $\uparrow$ & SR (\%) $\uparrow$ & SR (\%) $\uparrow$ \\
\midrule
OpenVLA~\cite{kim2024openvla}      & 0.0 & 0.0 & 20.0 & 0.0 & 4.0 & 4.80 \\
DD-VLA~\cite{liang2025discrete}      & 24.0 & 36.0 & 0.0 & 12.0 & 0.0 & 14.40 \\
$\pi_0$~\cite{black2024pi_0}        & \textbf{54.0} & \textbf{48.0} & 16.0 & 6.0  & 7.0  & 26.20 \\
$\pi_0$ FAST~\cite{pertsch2025fast} & 46.0 & 42.0 & 26.0 & 30.0 & \textbf{20.8} & 32.97 \\
$\pi_{0.5}$~\cite{intelligence2025pi05} & 24.0 & 18.0 & \textbf{36.0} & 20.0 & 6.5 & 20.90 \\
\midrule
\rowcolor{gray!20}
E0 (ours)                       & \textbf{54.0} & 34.0 & 12.0 & \textbf{72.0} & 18.75  & \textbf{38.15} \\
\bottomrule
\end{tabular}}

\end{table*}

\begin{table*}[htbp]
\centering
\caption{
\textbf{VLABench task performance (process score, PS) results.}
We report process-level performance across five language-conditioned selection tasks from VLABench~\cite{zhang2024vlabench}.
While baselines such as $\pi_0$~\cite{black2024pi_0}, $\pi_0$ FAST~\cite{pertsch2025fast}, and $\pi_{0.5}$~\cite{intelligence2025pi05} demonstrate task-specific strengths, our proposed E0 attains the highest average process score, reflecting more consistent and fine-grained reasoning throughout multi-step action modeling.
\textbf{Bold} numbers indicate the best performance in each column.
}
\label{tab-vlabench_ps_results}
\resizebox{0.8\textwidth}{!}{
\begin{tabular}{c|ccccc|c}
\toprule
\multirow{2}{*}{\textbf{Model}} & \textbf{Select Toy} & \textbf{Select Fruit} & \textbf{Select Painting} & \textbf{Select Poker} & \textbf{Select Mahjong} & \textbf{Average} \\
&  PS $\uparrow$  &  PS $\uparrow$  &  PS $\uparrow$  &  PS $\uparrow$  &  PS $\uparrow$  &  PS $\uparrow$ \\
\midrule
OpenVLA~\cite{kim2024openvla}       & 0.00 & 0.00 & 0.20 & 0.1066 & 0.1800 & 0.0973 \\
$\pi_0$~\cite{black2024pi_0}        & \textbf{0.76} & \textbf{0.72} & 0.16 & 0.1000 & 0.0814 & 0.3643 \\
$\pi_0$ FAST~\cite{pertsch2025fast} & 0.72 & 0.68 & 0.26 & 0.4200 & \textbf{0.3333} & 0.4827 \\
$\pi_{0.5}$~\cite{intelligence2025pi05} & 0.52 & 0.49 & \textbf{0.36} & 0.2867 & 0.1739 & 0.3661 \\
\midrule
\rowcolor{gray!20}
E0 (ours)                       & \textbf{0.76} & 0.65 & 0.12 & \textbf{0.7733} & 0.2500 & \textbf{0.5107} \\
\bottomrule

\end{tabular}}

\end{table*}

\section{More Details on RoboTwin}  \label{sec:robotwin_detailed}

\begin{figure*}[h]
  \centering
  \includegraphics[width=0.8\textwidth]{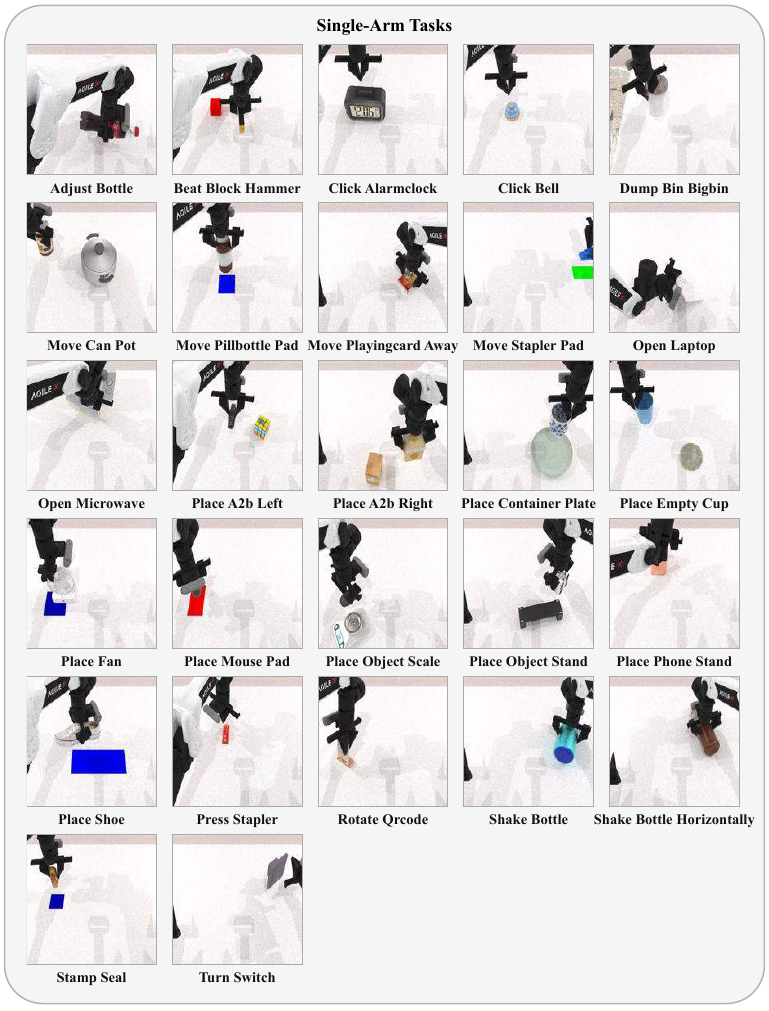}
  \caption{ 
  \textbf{RoboTwin single-arm tasks.} 
  A total of 27 diverse single-arm manipulation tasks are included, covering a wide range of objects and actions.
  }
  \label{fig-robotwin-tasks-single}
\end{figure*}

\begin{figure*}[h]
  \centering
  \includegraphics[width=0.8\textwidth]{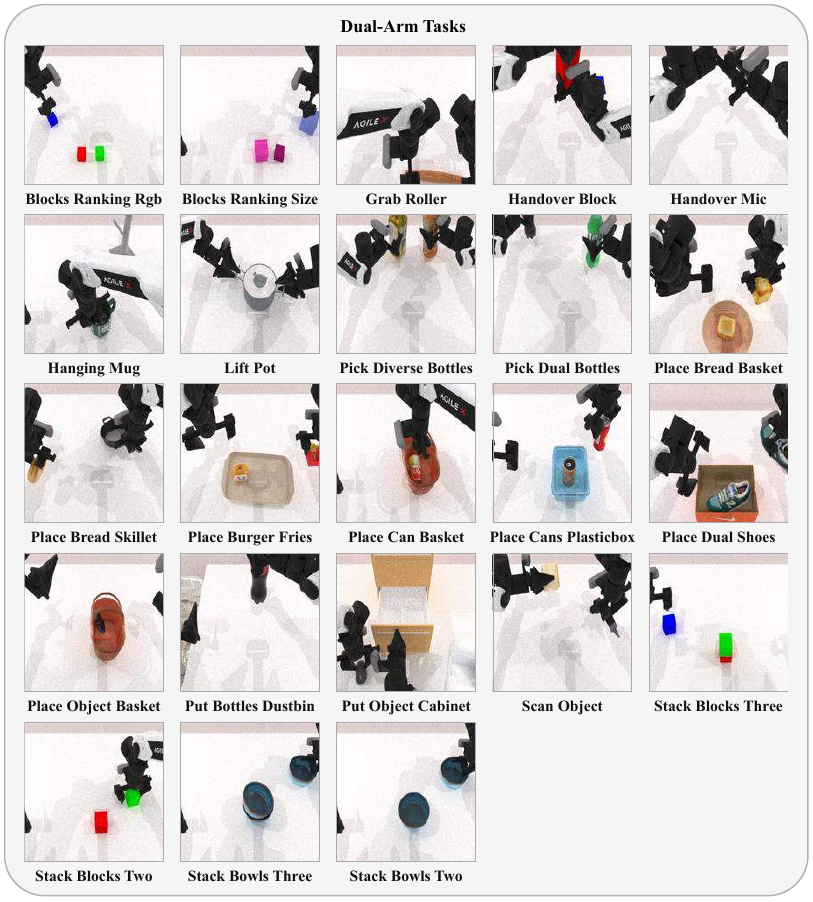}
  \caption{ 
  \textbf{RoboTwin dual-arm tasks.} 
  A total of 23 dual-arm manipulation tasks are included, covering a wide range of objects and actions. Some tasks require coordinated control between both arms, posing higher challenges for the model.
  }
  \label{fig-robotwin-tasks-dual}
\end{figure*}

To further systematically evaluate the robustness and generalization capability of our model, we conducted an additional experiment using \textbf{RoboTwin}~\cite{chen2025robotwin} as the primary benchmark for dual-arm manipulation. RoboTwin provides 50 diverse tasks (see Figure~\ref{fig-robotwin-tasks-single} and Figure~\ref{fig-robotwin-tasks-dual}), covering a wide range of object categories, both single-arm and dual-arm scenarios, and various interaction patterns. Each task includes both clean and heavily domain-randomized trajectories, enabling controlled evaluation under both standard and challenging conditions. Compared to prior simulation benchmarks, RoboTwin introduces significant domain randomization, including clutter, lighting variations, background textures, tabletop height, and linguistic diversity. These features make it particularly well-suited for assessing the robustness of VLA-based manipulation models in cluttered, visually diverse, and previously unseen environments.

For a fair comparison and to demonstrate generalization, we did not train a separate model for each individual task. Instead, \textbf{we mixed the initial data from all 50 official tasks and trained both our model and the baseline model $\pi_0$ under the same conditions}. Training was performed using a single RTX PRO 6000 GPU over 30,000 steps. Evaluation was conducted using a unified random seed. As shown in Table~\ref{tab-robotwin-all}, our model achieved the highest average performance across all tasks, surpassing the baseline by 8.0\%. It performed best on single-arm tasks, achieving a 13.7\% higher average success rate. Even on the more demanding dual-arm tasks that require precise coordination, our model still outperformed the baseline, with a 1.3\% higher average success rate.

\begin{figure*}[htbp]
  \centering
  \includegraphics[width=0.8\textwidth]{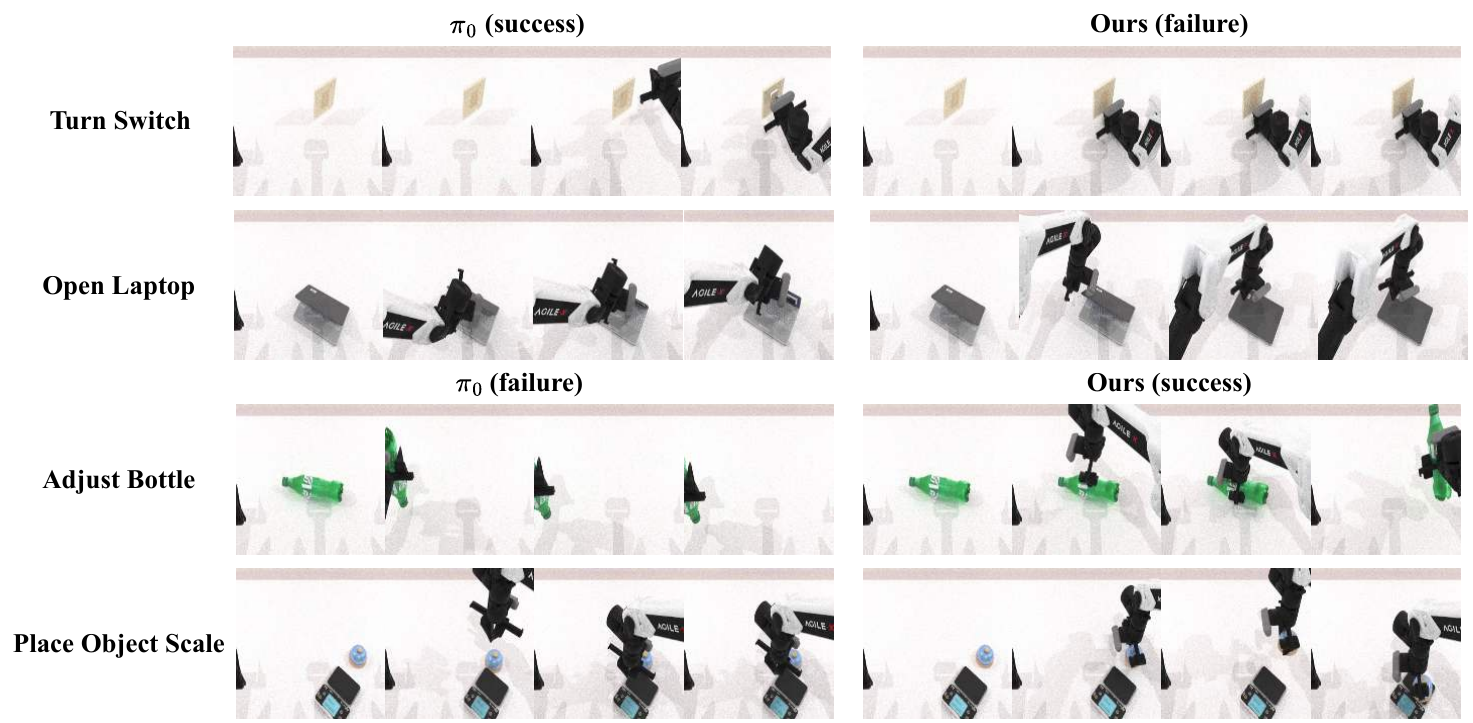}
  \caption{
      \textbf{Comparison on the RoboTwin benchmark.} In the \textbf{single-arm} tasks, our E0 shows the  performance deviations from $\pi_0$ across four tasks.
  }
  \label{fig-robotwin-tasks-compare-single}
\end{figure*}

\begin{figure*}[htbp]
  \centering
  \includegraphics[width=0.8\textwidth]{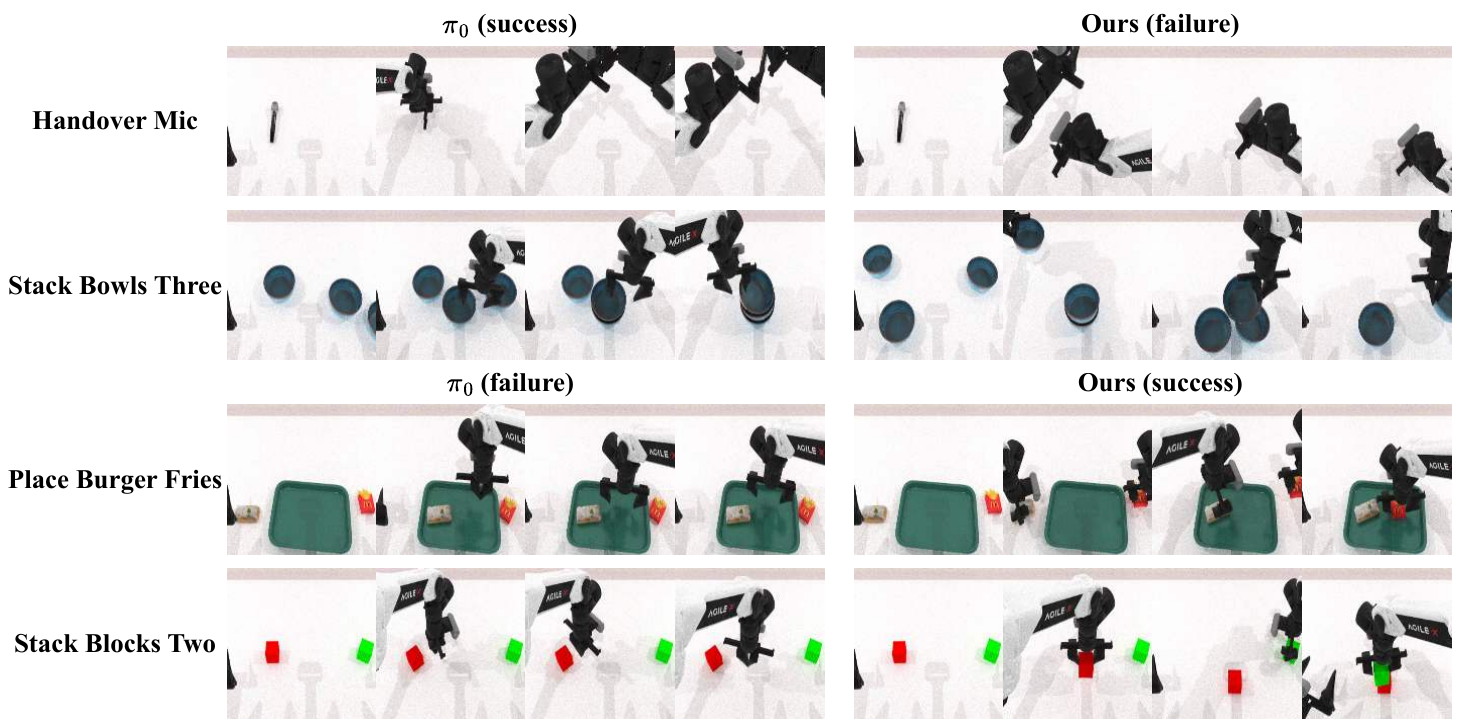}
  \caption{ 
      \textbf{Comparison on the RoboTwin benchmark.} In the \textbf{dual-arm} tasks, our E0 shows the  performance deviations from $\pi_0$ across four tasks.
  }
  \label{fig-robotwin-tasks-compare-dual}
\end{figure*}

Furthermore, Figure~\ref{fig-robotwin-tasks-compare-single} and Figure~\ref{fig-robotwin-tasks-compare-dual} present four tasks from the single-arm and dual-arm settings, respectively, where our model E0 shows the largest performance deviations from the baseline $\pi_0$. These include the two tasks with the highest positive improvement in success rate and the two tasks with the largest negative performance gap. This comparison further reveals the strengths and limitations of our model across diverse manipulation scenarios.

In the single-arm tasks (Figure~\ref{fig-robotwin-tasks-compare-single}), E0 achieves notable success over $\pi_0$ in \textit{Adjust Bottle} and \textit{Place Object Scale}. In \textit{Adjust Bottle}, our model succeeds where $\pi_0$ fails, indicating stronger capability in precise pose correction and object reorientation. In \textit{Place Object Scale}, E0 is able to perform accurate placement under spatial constraints, suggesting enhanced visual grounding and motor control. However, in \textit{Turn Switch} and \textit{Open Laptop}, E0 underperforms compared to $\pi_0$. Both tasks involve articulated or constrained motion (rotating a switch and opening a hinged object) where fine motor precision, torque control, or structural understanding are critical. These results suggest that while E0 generalizes well to visually diverse tasks, it may struggle with fine-grained control and interaction dynamics in tasks that require specific mechanical affordances.

In the dual-arm tasks (Figure~\ref{fig-robotwin-tasks-compare-dual}), E0 shows strong gains over the baseline in \textit{Place Burger Fries} and \textit{Stack Blocks Two}. Notably, the former involves cluttered environments, where our model maintains high success rates, suggesting better visual robustness and policy adaptability. However, in \textit{Handover Mic} and \textit{Stack Bowls Three}, E0 performs worse than $\pi_0$, revealing difficulties in tasks requiring complex bimanual coordination or long-horizon, sequential execution, especially when inter-step dependencies are critical.

Overall, these results highlight that while E0 demonstrates strong generalization across a wide range of tasks, particularly in visually complex environments and precision-driven single-arm manipulation, certain task categories still present performance bottlenecks. Tasks requiring fine-grained temporal coordination, long-horizon planning, or dual-arm synchronization continue to challenge the model's current capabilities. One possible contributing factor is the nature of mixed-task training: when training on a large and diverse set of tasks simultaneously, specific skills, such as delicate manipulation or tightly coupled bimanual control, may be underrepresented or diluted. This can lead to suboptimal specialization on tasks that demand precise timing or coordination. Addressing these limitations by exploring more balanced task sampling, modular policy design, or adaptive task curricula will be a central focus of our future work to further enhance robustness and skill transfer in complex manipulation settings.

\begin{table*}[h]
\centering
\caption{
    \textbf{RoboTwin results across all 50 tasks.} 
    \textcolor{red}{Red numbers} indicate improvement, while \textcolor{green}{green numbers} indicate decline. Compared comprehensively with $\pi_0$, our model E0 achieves the highest average success rate across single-arm, dual-arm, and overall task categories.
}
\label{tab-robotwin-all}
\renewcommand{\arraystretch}{1.05}
\setlength{\tabcolsep}{3pt}

\resizebox{0.8\textwidth}{!}{\begin{tabular}{c|c|c|c||c|c|c|c}
\toprule
\textbf{Task (Left)} & $\pi_0$ SR (\%) & E0 SR (\%) & Diff & 
\textbf{Task (Right)} & $\pi_0$ SR (\%) & E0 SR (\%) & Diff \\
\midrule
\multicolumn{4}{c||}{\textbf{Single-Arm Tasks}}                     &   stamp seal                   & 27.0 & 40.0 & \textcolor{red}{+13.0 } \\
adjust bottle                & 41.0 & 97.0 & \textcolor{red}{+56.0} & turn switch                  & 39.0 & 20.0 & \textcolor{green}{-19.0 } \\
beat block hammer            & 66.0 & 73.0 & \textcolor{red}{+7.0 } &  \multicolumn{4}{c}{\textbf{Dual-Arm Tasks}}  \\
click alarmclock             & 44.0 & 37.0 & \textcolor{green}{-7.0  } &  blocks ranking rgb           & 17.0 & 34.0 & \textcolor{red}{+17.0 } \\
click bell                   & 28.0 & 7.0  & \textcolor{green}{-21.0 } & blocks ranking size          & 7.0  & 11.0 & \textcolor{red}{+4.0  } \\
dump bin bigbin              & 58.0 & 54.0 & \textcolor{green}{-4.0  } & grab roller                  & 86.0 & 94.0 & \textcolor{red}{+8.0  } \\
move can pot                 & 41.0 & 44.0 & \textcolor{red}{+3.0  } & handover block               & 9.0  & 0.0  & \textcolor{green}{-9.0  } \\
move pillbottle pad          & 24.0 & 49.0 & \textcolor{red}{+25.0 } &  handover mic                 & 92.0 & 25.0 & \textcolor{green}{-67.0 } \\
move playingcard away        & 54.0 & 92.0 & \textcolor{red}{+38.0 } & hanging mug                  & 12.0 & 3.0  & \textcolor{green}{-9.0  } \\
move stapler pad             & 2.0  & 16.0 & \textcolor{red}{+14.0 } & lift pot                     & 32.0 & 40.0 & \textcolor{red}{+8.0  } \\
open laptop                  & 54.0 & 36.0 & \textcolor{green}{-18.0 } & pick diverse bottles         & 36.0 & 42.0 & \textcolor{red}{+6.0  } \\
open microwave               & 50.0 & 69.0 & \textcolor{red}{+19.0 } & pick dual bottles            & 52.0 & 45.0 & \textcolor{green}{-7.0  } \\
place a2b left               & 22.0 & 57.0 & \textcolor{red}{+35.0 } & place bread basket           & 35.0 & 60.0 & \textcolor{red}{+25.0 } \\
place a2b right              & 14.0 & 48.0 & \textcolor{red}{+34.0 } & place bread skillet          & 34.0 & 47.0 & \textcolor{red}{+13.0 } \\
place container plate        & 85.0 & 96.0 & \textcolor{red}{+11.0 } &  place burger fries           & 52.0 & 84.0 & \textcolor{red}{+32.0 } \\
place empty cup              & 51.0 & 88.0 & \textcolor{red}{+37.0 } &  place can basket             & 40.0 & 52.0 & \textcolor{red}{+12.0 } \\
place fan                    & 27.0 & 16.0 & \textcolor{green}{-11.0 } &  place cans plasticbox        & 18.0 & 47.0 & \textcolor{red}{+29.0 } \\
place mouse pad              & 6.0  & 39.0 & \textcolor{red}{+33.0 } & place dual shoes             & 15.0 & 13.0 & \textcolor{green}{-2.0  } \\
place object scale           & 15.0 & 58.0 & \textcolor{red}{+43.0 } & place object basket          & 55.0 & 55.0 & 0.0  \\
place object stand           & 58.0 & 83.0 & \textcolor{red}{+25.0 } & put bottles dustbin          & 31.0 & 4.0  & \textcolor{green}{-27.0 } \\
place phone stand            & 25.0 & 39.0 & \textcolor{red}{+14.0 } & put object cabinet           & 18.0 & 30.0 & \textcolor{red}{+12.0 } \\
place shoe                   & 47.0 & 53.0 & \textcolor{red}{+6.0  } & scan object                  & 32.0 & 33.0 & \textcolor{red}{+1.0  } \\
press stapler                & 52.0 & 91.0 & \textcolor{red}{+39.0 } & stack blocks three           & 17.0 & 10.0 & \textcolor{green}{-7.0  } \\
rotate qrcode                & 35.0 & 29.0 & \textcolor{green}{-6.0  } & stack blocks two             & 48.0 & 83.0 & \textcolor{red}{+35.0 } \\
shake bottle                 & 92.0 & 95.0 & \textcolor{red}{+3.0  }  & stack bowls three            & 58.0 & 28.0 & \textcolor{green}{-30.0 } \\
shake bottle horizontally    & 95.0 & 97.0 & \textcolor{red}{+2.0  }  & stack bowls two              & 92.0 & 78.0 & \textcolor{green}{-14.0 } \\
\midrule
\multicolumn{4}{c||}{\textbf{Average on Single-Arm Tasks (27 tasks)}} &
\multicolumn{4}{c}{\textbf{42.7 ($\pi_0$) $\rightarrow$ 56.4 (E0)} \textcolor{red}{(+13.7)}} \\
\multicolumn{4}{c||}{\textbf{Average on Dual-Arm Tasks (23 tasks)}} &
\multicolumn{4}{c}{\textbf{38.6 ($\pi_0$) $\rightarrow$ 39.9 (E0)} \textcolor{red}{(+1.3)}} \\
\multicolumn{4}{c||}{\textbf{Average (50 tasks)}} &
\multicolumn{4}{c}{\textbf{40.8  ($\pi_0$) $\rightarrow$ 48.8 (E0)} \textcolor{red}{(+8.0)}} \\
\bottomrule
\end{tabular}}
\end{table*}

\section{Hardware Details} \label{sec:hardware_detailed}

We conducted real-world experiments using a \textbf{Franka Emika Research 3 robotic arm}, equipped with two \textbf{Intel RealSense D435i cameras}. One camera was mounted on the robot’s wrist, providing a dynamic view that changes with the motion of the robotic arm. The second camera was placed laterally in the workspace to capture a stable third-person perspective. We utilized Gello~\cite{wu2024gello} to control the Franka arm and collect demonstration data.

\begin{figure}[t]
  \centering
  \includegraphics[width=0.5\textwidth]{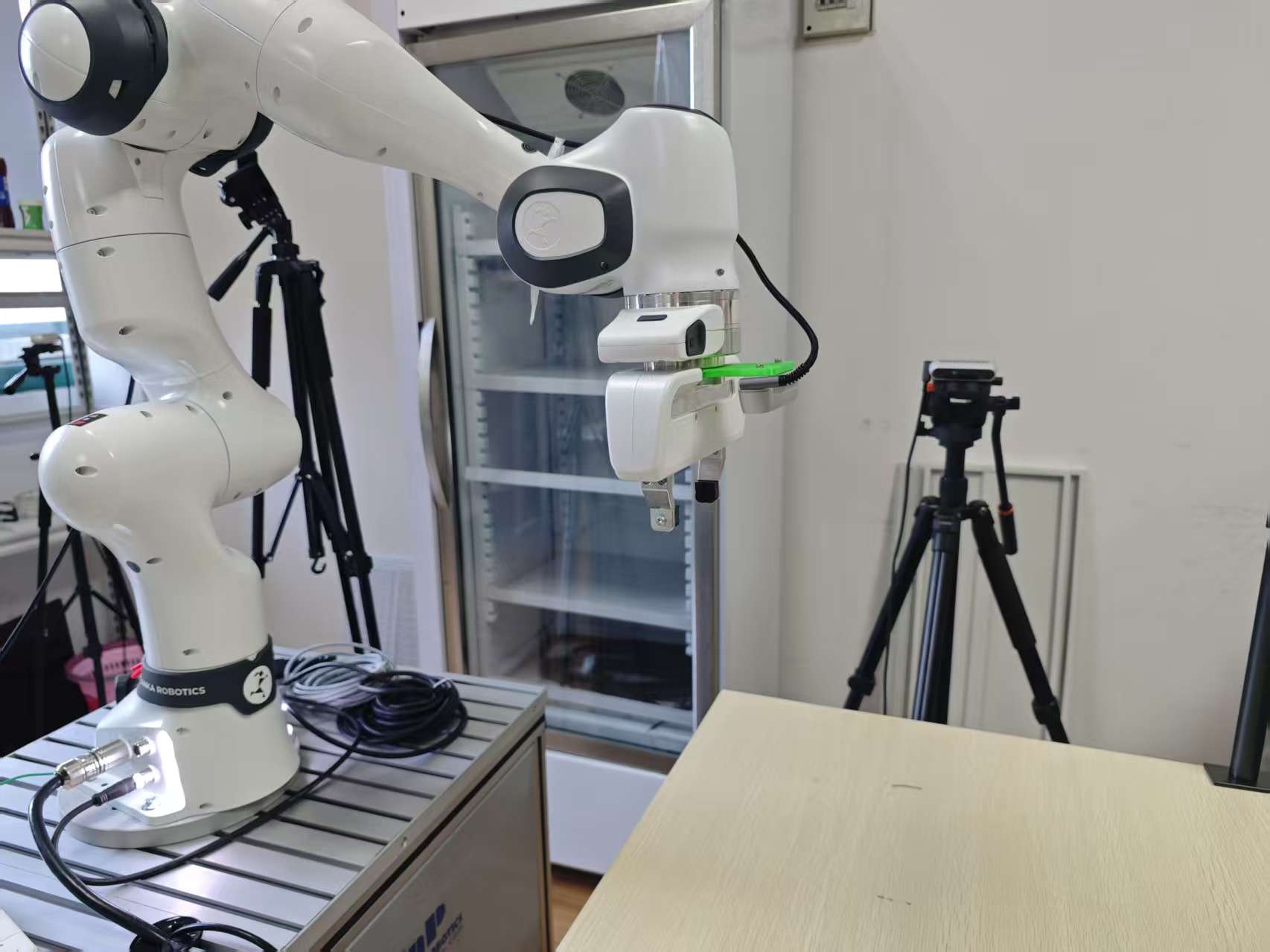}
  \caption{\textbf{Real-world evaluation setup}. Observations are captured using a side-view camera and a wrist-mounted camera, while an additional tripod-mounted camera records the scene from a rear-side perspective.}
  \label{fig:real_env}
\end{figure}

In the real-world evaluation setup (see Figure~\ref{fig:real_env}), we capture the visual observations required by the model using both a side-view camera and a wrist-mounted camera on the robot. In addition, a tripod-mounted camera is positioned behind and slightly to the side of the workspace to record the entire experiment for analysis and visualization.

\section{More Details on Real-World Experiments} \label{sec:real_detailed}

To demonstrate the real-world transferability, robust generalization, and fine-grained control capabilities of our model, we conducted a series of eight diverse real-world tasks using the Franka robotic arm. These tasks included five short-horizon tasks (pick a cube, press a button, stack cubes, pull a drawer, and close a door), as well as three long-horizon tasks (pick up two cubes consecutively, pull open a drawer and place a cube inside, and place a dish and close the door).

\textbf{For the short-horizon tasks, we collected 50 trajectories per task and trained a unified model across all tasks}. \textbf{For the long-horizon tasks, we collected 80 trajectories per task and trained a separate unified model}. All models were trained under identical settings on a single NVIDIA RTX PRO 6000 GPU for 30,000 steps. Detailed comparisons with baseline models are provided and analyzed in the main paper (see Table~\ref{tab:realworld_results}), and thus are not repeated here.

\begin{figure*}[t]
  \centering
  \includegraphics[width=0.8\textwidth]{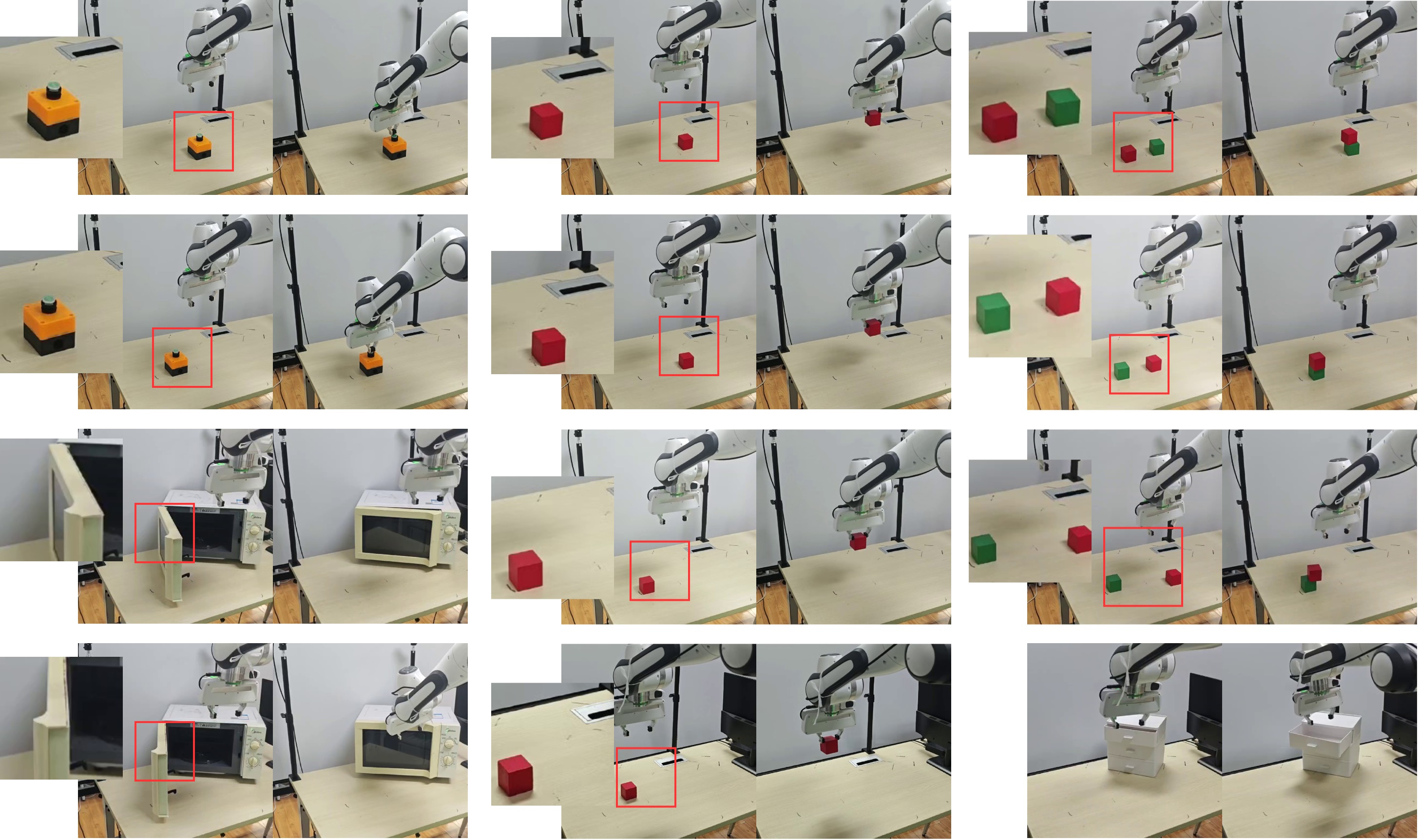}
  \caption{\textbf{Comparison of keyframes in real-world short-horizon tasks.}}
  \label{fig:real_robot_short_tasks}
\end{figure*}

To further evaluate the model's generalization in real-world scenarios, we designed several qualitative tests. As illustrated in Figure~\ref{fig:real_robot_short_tasks}, the model successfully recognizes and localizes red cubes placed at varying positions on the table in the \textit{pick block} task. In the \textit{close door} task, it adapts to different initial opening angles of the microwave door. In the \textit{press button} task, it accurately identifies small circular buttons at varying positions.

A particularly notable generalization test was performed in the stacking cubes task. During data collection, the green cube was consistently placed to the left of the red cube. However, during testing, we reversed their relative positions while keeping the instruction fixed: “Pick up the red cube and stack it on the green cube.” Remarkably, the model correctly interpreted the instruction, accurately identified the new target positions, and demonstrated strong generalization to scenarios not seen during training.

\begin{figure*}[t]   
  \centering
  \includegraphics[width=0.8\textwidth]{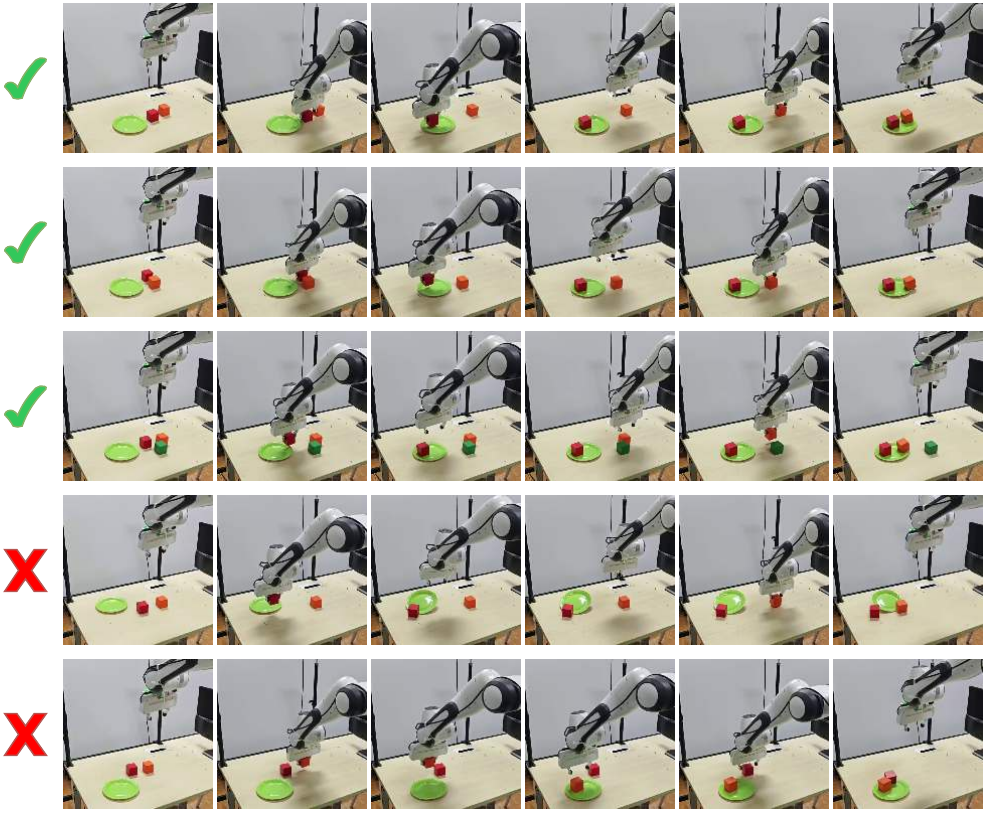}
  \caption{
  \textbf{Comparison of keyframes in the real-world \textit{pick twice} task under unseen scenarios.}
    During evaluation, we tested the model across various unseen scenarios. In most cases, the model successfully completed the task. In a few cases, task quality is slightly compromised but still acceptable—for example, placement deviations caused by an oversized green dish, or changes in the order of object picking due to color variations.
  }
  \label{fig:real_robot_cube_manipulation}
\end{figure*}

For the long-horizon tasks (e.g., \textit{pick twice}), the model similarly demonstrated robust generalization. As shown in Figure~\ref{fig:real_robot_cube_manipulation}, despite variations in the placement distance, positions of the colored cubes and dishes, and the addition of distractor cubes with new colors, the model was able to follow task instructions with high accuracy. Notable exceptions include Figure~\ref{fig:real_robot_cube_manipulation} (Row 4), where the green dish was placed near the robot's physical limit, resulting in marginal placement, and Row 5, where the picking order of cubes was incorrect. Nevertheless, the model successfully located the green dish, intentionally placed in a corner, and placed both cubes into it.

\begin{figure*}[t]
  \centering
  \includegraphics[width=0.8\textwidth]{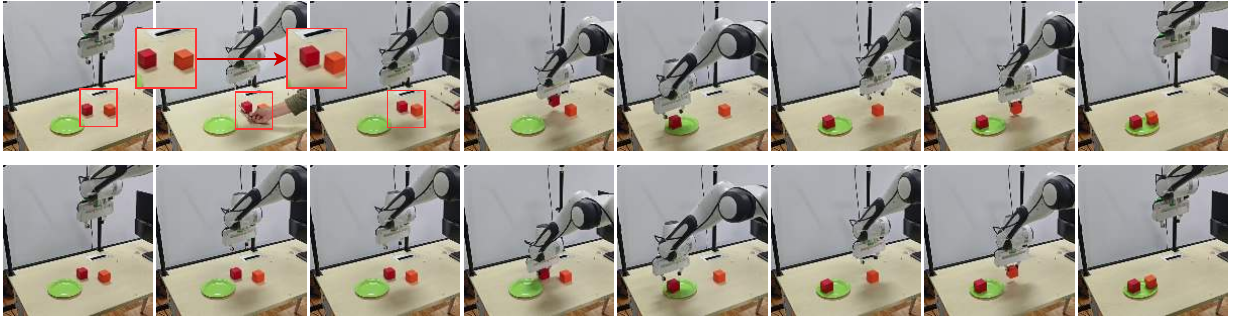}
  \caption{
    \textbf{Comparison of keyframes in the real-world \textit{pick twice} task with and without human intervention.} We introduce human perturbation by manually shifting the target cube to disrupt the model’s original plan. After the interruption, the model is able to promptly adapt and replan its actions.
  }
  \label{fig:real_robot_intervention_comparison}
\end{figure*}

Furthermore, Figure~\ref{fig:real_robot_intervention_comparison} presents a controlled experiment validating the model’s ability to make real-time decisions based on observations. In two identical trials, we introduced manual interference in one run by shifting the cube after the robot had initially targeted it. The robot responded appropriately by re-localizing the cube and completing the task successfully. This showcases the model’s ability to reason beyond simple trajectory mimicry and adapt dynamically to environmental changes.

\section{More Real-World Generalization Experiments} \label{sec:real_more_generalization_detailed}

\begin{figure*}[htbp]
  \centering
  \includegraphics[width=0.8\textwidth]{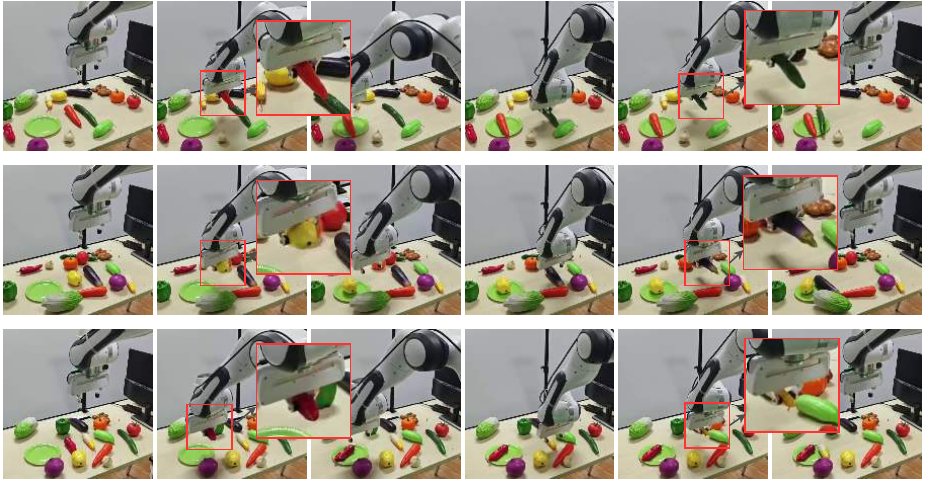}
  \caption{
  \textbf{Qualitative results on the real-world \textit{Pick Vegetables Twice} task under randomly arranged tabletop scenes}. Each row shows one complete execution sequence consisting of two consecutive pick-and-place operations. Top row: carrot and cucumber. Middle row: potato and eggplant. Bottom row: pepper and corn. Across all settings, the surrounding distractor objects are placed in completely random configurations, demonstrating the robustness of our policy under visually diverse and cluttered real-world environments.
  }
  \label{fig:real_robot_3_tasks}
\end{figure*}

To further assess the model’s capacity for real-world generalization, we conducted an additional long-horizon pick-twice task experiment in highly cluttered environments, as illustrated in Figure~\ref{fig:real_robot_3_tasks}. The experimental setup involved a table scattered with various toy vegetable models. During data collection, object placement was randomized across scenes without any memory restoration between episodes. In total, we collected 120 diverse and highly unstructured trajectories for training under these chaotic conditions.

Training was again performed using a single NVIDIA RTX PRO 6000 GPU for 30,000 steps. During evaluation, we maintained the same principle of random object placement and scene variability. The results show that the model was capable of adapting to highly disordered environments, precisely interpreting and following language instructions to successfully complete the tasks. This experiment highlights the model’s strong environmental generalization and its robustness in real-world, unpredictable scenarios.

\section{Spherical Perspective Generalization}\label{sec:sphercical_detailed}

\emph{The Spherical Perspective is purely an isolated, extra experiment in this section. All other evaluations in the paper—which establish the core generalization results we report—strictly evaluate TDD without applying Spherical Perspective Augmentation.}

We additionally modify the model and evaluate it against the baseline in environments with dynamic camera perturbations. To improve robustness against dynamic camera perturbations, we introduce a spherical warping augmentation and a corresponding relative spherical embedding. This design is motivated by a key limitation in current robot learning setups: Most policies are trained and validated under fixed camera viewpoints, which leads to a strong dependency on camera placement. As a result, when the camera position or orientation is altered, the inference accuracy of the policy degrades significantly. By explicitly augmenting training with spherical viewpoint perturbations and embedding the relative camera offsets into the representation space, our approach mitigates this reliance on fixed viewpoints and enables more robust generalization under dynamic or shifted camera configurations.

\noindent\textbf{Spherical Warping.} Given an RGB image $I \in \mathbb{R}^{H \times W \times 3}$ and camera intrinsics $(f_x, f_y, c_x, c_y)$, each pixel $(u,v)$ is first unprojected to a 3D point at fixed depth $d_\mathrm{center}$:
\begin{equation}
    X = \frac{(u - c_x) \, d_\mathrm{center}}{f_x}, \quad 
    Y = \frac{(v - c_y) \, d_\mathrm{center}}{f_y}, \quad 
    Z = d_\mathrm{center}.
\end{equation}

We then apply spherical perturbations defined by yaw ($\Delta \theta$) and pitch ($\Delta \phi$) rotations:
\begin{equation}
    R = R_x(\Delta \phi)\, R_y(\Delta \theta), 
    \quad 
    \begin{bmatrix} X' \\ Y' \\ Z' \end{bmatrix} = R \begin{bmatrix} X \\ Y \\ Z \end{bmatrix}.
\end{equation}

Finally, the rotated points are projected back to the image plane:
\begin{equation}
    u' = \frac{X' f_x}{Z'} + c_x, 
    \quad v' = \frac{Y' f_y}{Z'} + c_y,
\end{equation}
yielding a warped image $I'$. This augmentation simulates dynamic camera views around the robot.

\noindent\textbf{Relative Spherical Embedding.} To explicitly encode the relative perturbation of each camera, we define a 3D offset vector $\delta = [d, \; \theta, \; \phi]$, where $(d, \theta, \phi)$ denote radial, horizontal, and vertical displacements. For the base camera, we use $\delta_\mathrm{orig} = [-d, -\theta, -\phi]$, and for the warped camera, $\delta_\mathrm{warp} = [d, \theta, \phi]$. These offsets are mapped into the token embedding space through a learnable projection function $f_\mathrm{proj}$:
\begin{equation}
    e_\mathrm{rel} = f_\mathrm{proj}(\delta).
\end{equation}

Let $z_\mathrm{img} \in \mathbb{R}^{N\times d}$ denote the image tokens extracted by the vision encoder. We integrate the relative embedding additively:
\begin{equation}
    z'_\mathrm{img} = z_\mathrm{img} + e_\mathrm{rel}.
\end{equation}

This design allows the model to not only utilize visual features but also explicitly account for the relative camera perturbations, thereby improving cross-view consistency and robustness in action generation.

\begin{table}[!htbp]
\centering
\caption{
\textbf{Evaluation under camera viewpoint perturbations in the LIBERO environments.} 
We evaluate the robustness of different models when camera positions and orientations are dynamically perturbed. 
}
\label{tab-ablation-view-change}
\resizebox{0.85\textwidth}{!}{
\begin{tabular}{c|cccc|c}
\toprule
\multirow{2}{*}{\textbf{Model}} & \textbf{Libero-Spatial} & \textbf{Libero-Object} & \textbf{Libero-Goal} & \textbf{Libero-Long} & \textbf{Average} \\
& SR (\%) $\uparrow$ & SR (\%) $\uparrow$ & SR (\%) $\uparrow$ & SR (\%) $\uparrow$ & SR (\%) $\uparrow$ \\
\midrule
$\pi_0$~\cite{black2024pi_0}                    & 28.0 & 16.6 & 31.6 & 2.6 & 19.7 \\
$\pi_0$~\cite{black2024pi_0} + view               & 53.8 & 75.2 & 57.8 & 16.2 & 50.8 (+31.1)  \\
\rowcolor{gray!20}
E0 (Ours)                                       & 74.4 & 85.2 & 77.4 & 28.8 & 66.5 \\
\rowcolor{gray!20}
E0 (Ours) + view                                & 92.0 & 96.8 & 83.0 & 63.8 & 83.9 (+17.4) \\
\bottomrule
\end{tabular}}
\end{table}

As shown in Table~\ref{tab-ablation-view-change}, both $\pi_0$ and our E0 exhibit substantial performance degradation without view augmentation (average SR of 19.7\% and 66.5\%, respectively), indicating strong sensitivity to camera changes. Incorporating spherical view augmentation (\textit{+view}) markedly improves robustness, boosting average SR to 50.8\% and 83.9\%. These results demonstrate that view-aware design significantly enhances generalization to unseen or dynamic camera configurations, a key requirement for reliable real-world deployment.

To enhance policy robustness under dynamic camera perturbations, we introduce a \emph{spherical warping augmentation} coupled with a \emph{relative spherical embedding}. This design mitigates the common issue of viewpoint overfitting in visuomotor policies trained under fixed camera setups. By encoding 3D camera displacements into the latent representation, the model achieves consistent feature alignment across varying viewpoints.

\section{Real-World Tasks Instructions}  \label{sec:real_instruction_detailed}

\noindent\textbf{\textit{Pick block.}} Pick up the red block on the table.

\noindent\textbf{\textit{Close door.}} Close the microwave oven door.

\noindent\textbf{\textit{Press button.}} Press the green button.

\noindent\textbf{\textit{Pull drawer.}} Pull out the top drawer.

\noindent\textbf{\textit{Stack block.}} Pick up the red block and stack it on the green block.

\noindent\textbf{\textit{Pick twice.}} Pick up the red and orange cubes in turn and place them in the green dish.

\noindent\textbf{\textit{Open \& put.}} Open the top drawer and pick up the red cube and put it in.

\noindent\textbf{\textit{Put \& close.}} Pick up the red plate, put it in the microwave, and close the door.

\noindent\textbf{\textit{Pick vegetables twice 1.}} Pick up the red pepper and yellow corn in turn and place them in the green dish.

\noindent\textbf{\textit{Pick vegetables twice 2.}} Pick up the yellow potato and purple eggplant in turn and place them in the green dish.

\noindent\textbf{\textit{Pick vegetables twice 3.}} Pick up the red carrot and green cucumber in turn and place them in the green dish.

\section{Limitations}

Our evaluation focuses on robotic manipulation, with physical experiments conducted on a single Franka platform. Although the simulation experiments include bimanual manipulation, performance on additional physical embodiments and dexterous hands with more than 20 control dimensions remains unverified. The limited number of real-world trials and evaluation rounds also constrains the precision of task-level comparisons. The action representation depends on normalization statistics estimated from the training data. Distribution shifts in action ranges or control requirements may therefore require recalibrating the normalization and quantization scheme.

TDD requires iterative denoising and categorical input and output representations. Increasing the discretization resolution, action horizon, or number of control channels can increase computational and memory requirements. The reported inference measurements apply to the tested hardware and configurations, and deployment under tighter latency budgets requires further evaluation.

\section{Benefits of Discrete Action Modeling}
\label{sec:Benefits of Discrete Action Tokens}
Recent works advocate for discrete action models in robotics \citep{florence2021decisiveness,shafiullah2022behavior,liang2025discrete,li2025discrete,luo2026being}, attributing their benefits to several key aspects. \textbf{First}, they mitigate the multi-peaked prediction failure inherent in continuous prediction, which mathematically forces a model to predict the conditional expectation $\hat{y} = \mathbb{E}[y|x]$, often resulting in invalid averaged actions; in contrast, discrete prediction minimizes Cross-Entropy to learn a probability mass function that captures distinct modes (see Section \ref{sect:appendix-Multimodal}). \textbf{Second}, discrete modeling preserves the backbone’s pre-trained vision and language capabilities—analogous to extending a vocabulary to new languages—while offering a potential path to inherit the scaling behavior of unified transformers for large-scale VLA research (see Section \ref{sect:appendix-vlm}). \textbf{Third}, it facilitates architectural unification by keeping action generation inside the transformer with the same training objective (cross-entropy loss) used by VLMs, enabling unified modeling for images, language, and actions (see Section \ref{sect:appendix-unify}). \textbf{Fourth}, discrete modeling matches the true quantized nature of real-world robot control, where hardware constraints like encoder resolution inherently discretize continuous signals (see Section \ref{sect:appendix-nature}).

\subsection{Addressing Multi-Peaked Prediction Failure}\label{sect:appendix-Multimodal}

A significant body of literature has explicitly articulated why discrete action prediction often outperforms continuous action prediction \citep{florence2021decisiveness,shafiullah2022behavior}.

A primary argument is that standard continuous prediction methods fail to model multi-peaked data effectively because they average conflicting modes, leading to invalid actions. As noted by Google Research (see https://research.google/blog/decisiveness-in-imitation-learning-for-robots/) \citep{florence2021decisiveness}, this is critical in robotics. For example, in a pick-and-place task, a robot might successfully grasp an object by approaching from the left or the right. However, a continuous model trained on both strategies tends to output the average—driving the robot straight into the object. While the robot must be capable of precise continuous adjustments, it must first commit to a distinct high-level mode (left vs. right) rather than blending them.

This challenge is further addressed in works such as Behavior Transformer (BeT) \cite{shafiullah2022behavior}. The authors explicitly state that standard continuous regression fails due to this mode-averaging effect. To resolve this, BeT proposes discretizing the continuous action space to leverage the classification capabilities of sequence models. By using a Transformer to output a categorical distribution (via Softmax), the model can represent multiple distinct modes simultaneously, preserving the multi-peaked nature of the data.

The theoretical foundation for the ``Multi-Peaked Prediction Failure'' was rigorously analyzed in the Implicit Behavioral Cloning (IBC) paper \citep{florence2022implicit}. Although IBC utilizes Energy-Based Models rather than discrete tokens, its mathematical evidence—demonstrating that MSE regression collapses on non-unimodal data—serves as the primary justification for the field's shift toward alternative architectures. Discrete tokens can be viewed as a ``grid-based'' implementation of the principles championed in the IBC paper.

Given that the ``Theory of Multi-Peaked Distribution''—in particular, the necessity of avoiding the “mean” trap—is now widely accepted, providing a full proof is not our primary objective. We therefore present only a proof sketch below.

\subsubsection{Definitions and Setup}

Let $S$ denote the state/observation space (e.g., visual input) and $A \in \mathbb{R}^d$ the continuous action space.

\begin{definition}[Expert Policy]
The true data-generating distribution is denoted by $p(A|S=s)$. We assume access to a dataset of $n$ i.i.d. samples $\mathcal{D} = \{(s_i, a_i)\}_{i=1}^n \sim p(S, A)$.
\end{definition}

\begin{definition}[Explicit Prediction]
A policy $\hat{\pi}: S \to A$ is parameterized as $f_\theta(s)$:
\begin{equation}
    \mathcal{L}(\theta) = \frac{1}{n} \sum_{i=1}^n \| f_\theta(s_i) - a_i \|_2^2.
\end{equation}
\end{definition}

\begin{definition}[Multi-Peaked]
For a fixed state $s$, the distribution $p(A|S=s)$ is \textbf{multi-peaked} if it consists of $k \geq 2$ distinct modes (local maxima) separated by regions of low probability density. Formally, there exist disjoint compact sets $M_1, \dots, M_k \subset \mathbb{R}^d$ such that:
\begin{equation}
    p(A|S=s) \approx \sum_{j=1}^k w_j \cdot \delta_{M_j}(A), \quad \text{where } \sum_{j=1}^k w_j = 1, \quad w_j > 0.
\end{equation}
We assume separation between modes, i.e., $\|\mu_i - \mu_j\|_2 > \epsilon$ for centers $\mu_i, \mu_j$ of $M_i, M_j$.
\end{definition}

\subsubsection{Theorem: The Failure Continuous Modeling on Multi-Peaked Distributions}

\begin{theorem}
The population-optimal predictor $\hat{\pi}^*(s) = \arg\min_f \mathbb{E}_{A \sim p(A|S=s)} [ \|f(s) - A\|_2^2 ]$ is the conditional expectation $\mathbb{E}[A|S=s]$. For a multi-peaked distribution $p(A|S=s)$ with separated modes, this mean lies in a low-density region between modes, yielding actions with high expert log-probability cost and leading to invalid control.
\end{theorem}

\begin{proof}
We proceed in three steps: deriving the optimality, analyzing the geometric placement of the mean, and demonstrating the consequent performance degradation.

\paragraph{Step 1: Optimality implies Conditional Expectation}

The population loss is defined as:
\begin{equation}
    L(f) = \mathbb{E}_{A \sim p(A|S=s)} [ \|f(s) - A\|_2^2 ] = \int \|f(s) - a\|_2^2 \, p(a|s) \, da.
\end{equation}
To find the minimizer, we take the functional derivative with respect to $f(s)$ and set it to zero:
\begin{equation}
    \frac{\partial L}{\partial f(s)} = 2 \int (f(s) - a) \, p(a|s) \, da = 0 \implies f(s) \int p(a|s) da = \int a \, p(a|s) \, da.
\end{equation}
Thus, the optimal predictor is:
\begin{equation}
    \hat{\pi}^*(s) = \mathbb{E}[A|S=s].
\end{equation}
The minimal loss achieved is the trace of the conditional covariance, $\text{Tr}(\text{Cov}(A|S=s))$, confirming that the loss strictly targets the first moment of the distribution.

\paragraph{Step 2: The Mean Lies in Low-Density Regions}
Under the multi-peaked assumption, let the modes have centers $\mu_1, \dots, \mu_k$ with weights $w_j$. The conditional mean is a convex combination:
\begin{equation}
    \mu^* = \mathbb{E}[A|S=s] = \sum_{j=1}^k w_j \mu_j.
\end{equation}
Since the modes are separated and $k \geq 2$, $\mu^*$ lies strictly inside the convex hull of the $\mu_j$'s. If we assume an idealized multi-peaked distribution where density is zero outside the modes ($p(a|s) = 0$ for $a \notin \bigcup_j M_j$), then by convexity, the mean of disjoint sets lies in the gap between them.
\begin{equation}
    \mu^* \notin \bigcup_{j=1}^k M_j \implies p(\mu^*|s) \approx 0.
\end{equation}
Even with non-zero density $\eta$ between modes, $p(\mu^*|s) \ll \max_a p(a|s)$.

\paragraph{Step 3: Invalidity and Performance Degradation}
We evaluate the policy using the Expert Cost (negative log-likelihood), which corresponds to the probability of the generated action under the expert's distribution.
For the MSE-optimal policy $\hat{\pi}^*(s) = \mu^*$:
\begin{equation}
    \text{Expert Cost}(\mu^*|s) = -\log p(\mu^*|s).
\end{equation}
As the density between modes $p(\mu^*|s) \to 0$ (e.g., in ``set-valued'' functions described in \citep{florence2022implicit}), the cost approaches infinity:
\begin{equation}
    \lim_{p(\mu^*|s) \to 0} -\log p(\mu^*|s) = +\infty.
\end{equation}
In contrast, a policy that samples from the modes $\tilde{a} \sim p(A|S=s)$ has a finite expected cost equal to the entropy $H(p(A|S=s))$.

\textbf{Physical Interpretation:} In a robotics obstacle avoidance scenario, if $\mu_1$ corresponds to ``Go Left'' and $\mu_2$ corresponds to ``Go Right'', the mean $\mu^*$ corresponds to ``Go Straight''. Since ``Go Straight'' hits the obstacle, the optimal action is physically invalid despite being statistically optimal.
\end{proof}

\subsubsection{Solution: Discrete Tokenization}
Discrete approaches model $p(A|S=s)$ as a categorical distribution over discrete bins. The policy outputs a probability vector $q$ via a Softmax function:
\begin{equation}
    \hat{\pi}(s) = \arg\max_j q_j, \quad q \sim \text{Softmax}(W h(s)).
\end{equation}
Optimizing the Cross-Entropy (CE) loss:
\begin{equation}
    \mathcal{L}_{\text{CE}} = -\sum_j p_j \log q_j,
\end{equation}
directly maximizes the log-likelihood of the modes. Unlike the continuous loss, CE does not force averaging; it allows the distribution to be multi-peaked (e.g., $P(\text{Left})=0.5, P(\text{Right})=0.5, P(\text{Straight})=0.0$), successfully avoiding the mean trap.

\subsubsection{Empirical Analysis of Action Multimodality} \label{app:multimodality}

We empirically examine the relationship between action multimodality and mean-action deviation using 1,693 LIBERO demonstration episodes, comprising 273,465 frames and 52,334 ten-step action chunks. For each of 10,000 query states, we retrieve 30 task-matched nearest neighbors from different episodes and cluster their future action chunks into two empirical modes. We measure the normalized Euclidean distance between the conditional mean action and its nearest empirical mode.

As shown in Table~\ref{tab:multimodal_geometry}, the mean-to-nearest-mode error increases monotonically with multimodality, from 0.3731 in the low-multimodality group to 0.5458 in the high-multimodality group, corresponding to a 46.29\% relative increase. The absolute difference is 0.1727, with a task-clustered bootstrap 95\% confidence interval of $[0.0955, 0.2510]$. Multimodality is positively associated with mean-action deviation (Spearman $\rho=0.2914$, $p=5.93\times10^{-195}$).

These results provide empirical evidence that greater action-mode separation is associated with a larger geometric deviation between the conditional mean and locally observed action modes.

\begin{table}[t]
\centering
\small
\caption{\textbf{Effect of action multimodality on mean-action
deviation.} We report the normalized distance between the conditional
mean action and its nearest empirical action mode on LIBERO
demonstrations. Lower values indicate smaller geometric deviation.}
\label{tab:multimodal_geometry}
\small
\begin{tabular}{lcc}
\toprule
Multimodality & Number of Queries & Mean-to-Nearest-Mode Error
$\downarrow$ \\
\midrule
Low    & 3,333 & \textbf{0.3731} \\
Medium & 3,334 & 0.4307 \\
High   & 3,333 & 0.5458 \\
\bottomrule
\end{tabular}
\end{table}

\subsubsection{Mode Consistency of Generated Actions} \label{app:multimodality2}

We further assess whether action multimodality is associated with prediction errors in learned generative policies. We compare E0 with a training-matched $\pi_{0.5}$ baseline using identical LIBERO demonstrations, pretrained $\pi_{0.5}$-base initialization, a global batch size of 32, 30,000 optimizer steps, AdamW, gradient clipping, EMA, action horizon, and normalization statistics.

Both policies are evaluated on the same 3,000 recorded LIBERO observations, including third-person images, wrist images, robot states, and language instructions. We use nearest-mode error as the primary metric, since it measures proximity to any locally supported action mode without penalizing a valid strategy solely for differing from the recorded demonstration.

As shown in Table~\ref{tab:multimodal_group}, both models exhibit increasing nearest-mode error with higher multimodality. For $\pi_{0.5}$, the error increases from 0.7446 to 0.8214, a relative increase of 10.32\%. E0 shows the same overall trend but achieves consistently lower errors across all three groups. This indicates a systematic improvement in mode proximity.

\begin{table}[t]
\centering
\caption{\textbf{Nearest-mode prediction error across multimodality
levels.} E0 and $\pi_{0.5}$ are evaluated on the same LIBERO observations.
Lower values indicate greater proximity to empirical action modes.}
\label{tab:multimodal_group}
\small
\begin{tabular}{lccc}
\toprule
Model & Low $\downarrow$ & Medium $\downarrow$ & High $\downarrow$ \\
\midrule
$\pi_{0.5}$ & 0.7446 & 0.7834 & 0.8214 \\
E0 (ours) & \textbf{0.6980} & \textbf{0.7299} & \textbf{0.7705} \\
\bottomrule
\end{tabular}
\end{table}

Table~\ref{tab:multimodal_comparison} presents the overall comparison across all 3,000 paired observations. E0 reduces nearest-mode error from 0.7831 to 0.7328, corresponding to a 6.43\% relative improvement. The paired difference is $-0.0503$, with a task-clustered bootstrap 95\% confidence interval of $[-0.0764,-0.0301]$ and paired Wilcoxon $p=9.55\times10^{-13}$.

Beyond nearest-mode proximity, E0 reduces the off-mode rate from 11.07\% to 8.03\% ($p=2.10\times10^{-7}$), decreases target-action ADE from 0.8286 to 0.5680, and improves demonstration-consistent mode selection from 78.50\% to 87.63\% ($p=1.21\times10^{-34}$). The task-clustered 95\% confidence interval for the ADE difference is $[-0.2947,-0.2295]$.

Together, these results show that E0 produces action predictions that are more closely aligned with locally observed action modes and recorded demonstrations under the evaluated conditions.

\begin{table}[t]
\centering
\caption{\textbf{Mode consistency and action prediction accuracy on LIBERO.} Results are computed over 3,000 paired observations. Relative improvements are reported for error metrics, while mode-selection improvement is reported in percentage points.}
\label{tab:multimodal_comparison}
\small
\setlength{\tabcolsep}{5pt}
\begin{tabular}{lccc}
\toprule
Metric & $\pi_{0.5}$ & E0 (ours) & Improvement \\
\midrule
Nearest-mode error $\downarrow$
& 0.7831 & \textbf{0.7328} & 6.43\% lower \\
Off-mode rate $\downarrow$
& 11.07\% & \textbf{8.03\%} & 27.41\% lower \\
Target-action ADE $\downarrow$
& 0.8286 & \textbf{0.5680} & 31.44\% lower \\
Mode selection $\uparrow$
& 78.50\% & \textbf{87.63\%} & +9.13 pp \\
\bottomrule
\end{tabular}
\end{table}

\subsection{Preserving Capabilities and Knowledge of Pre-trained Vision-Language Models}\label{sect:appendix-vlm}

Recent literature suggests that discrete modeling effectively preserves a backbone’s pre-trained vision and language capabilities—analogous to extending a vocabulary to new languages—thereby allowing VLA models to inherit the scaling properties of unified transformers \citep{liang2025discrete,li2025discrete}. Fully leveraging this pre-trained knowledge is critical; yet, while utilizing the same cross-entropy objective is hypothesized to minimize interference with the backbone's original representations, a formal theoretical validation is currently missing. Given the widespread empirical adoption of this approach, we provide a proof sketch to substantiate this intuition rather than an exhaustive derivation. Specifically, we analyze the gradient dynamics with respect to the backbone's hidden states $h$, demonstrating the superiority of discrete modeling by comparing the Gradient Boundedness and feature-space compatibility of the two loss functions.

\subsubsection{Definitions}

Let $f_\theta(x)$ denote the VLM backbone that produces a high-dimensional hidden state
$h \in \mathbb{R}^d$ for a given context, where $\theta$ represents the pre-trained parameters.

\paragraph{1. Continuous Case.}
We project $h$ to a continuous action $y \in \mathbb{R}^k$ using a linear head $W_{\mathrm{mse}}$:
\begin{equation}
\mathcal{L}_{\mathrm{MSE}} = \frac{1}{2} \left\| W_{\mathrm{mse}} h - y_{\text{target}} \right\|_2^2 .
\end{equation}

\paragraph{2. Discrete Case.}
We project $h$ to logits over a vocabulary $V$ using a linear head $W_{\mathrm{ce}}$
(the token embedding matrix), followed by a Softmax:
\begin{equation}
\mathcal{L}_{\mathrm{CE}}
= - \log \left(
\frac{\exp\left(w_{\text{target}}^\top h\right)}
{\sum_{j \in V} \exp\left(w_j^\top h\right)}
\right) .
\end{equation}

\subsubsection{Proof 1: Gradient Stability (The Lipschitz Constraint)}

\paragraph{Proposition.}
\emph{The gradient of the discrete loss is strictly bounded, whereas the gradient of the
continuous loss is unbounded. Unbounded gradients cause large weight updates that move
$\theta$ out of the pre-trained trusted region, leading to catastrophic forgetting.}

\paragraph{Derivation.}

\paragraph{1. Gradient of the Discrete Loss.}
Let $p$ be the probability vector produced by the Softmax.
The gradient with respect to the hidden state $h$ is
\begin{equation}\label{eqn:bounded}
\nabla_h \mathcal{L}_{\mathrm{CE}}
= \sum_{j \in V} \left( p_j - \mathbb{I}[j = \text{target}] \right) w_j .
\end{equation}

\begin{proof}
Let the logit for class $j$ be defined as
\begin{equation}
z_j = w_j^\top h .
\end{equation}
The loss function is
\begin{equation}
\mathcal{L}
= -\log(p_{\text{target}})
= -z_{\text{target}} + \log\left( \sum_{j \in V} \exp(z_j) \right) .
\end{equation}
We seek the gradient with respect to $h$. By the chain rule,
\begin{equation}
\nabla_h \mathcal{L}
= \sum_{j \in V}
\frac{\partial \mathcal{L}}{\partial z_j}
\frac{\partial z_j}{\partial h} .
\end{equation}
\paragraph{Step A: Derivative with Respect to Logits.}
This is the standard Softmax–Cross-Entropy derivative.
\begin{itemize}
\item If $j = \text{target}$:
\begin{equation}
\frac{\partial \mathcal{L}}{\partial z_{\text{target}}}
= -1 + \frac{\exp(z_{\text{target}})}{\sum_k \exp(z_k)}
= -1 + p_{\text{target}} .
\end{equation}

\item If $j \neq \text{target}$:
\begin{equation}
\frac{\partial \mathcal{L}}{\partial z_j}
= \frac{\exp(z_j)}{\sum_k \exp(z_k)}
= p_j .
\end{equation}
\end{itemize}
These two cases can be unified using the indicator function:
\begin{equation}
\frac{\partial \mathcal{L}}{\partial z_j}
= p_j - \mathbb{I}[j = \text{target}] .
\end{equation}

\paragraph{Step B: Derivative of Logits with Respect to the Hidden State.}
Since $z_j = w_j^\top h$, we have
\begin{equation}
\frac{\partial z_j}{\partial h} = w_j .
\end{equation}

\paragraph{Step C: Applying the Chain Rule.}
Substituting the results from Steps A and B yields
\begin{equation}
\nabla_h \mathcal{L}
= \sum_{j \in V}
\left( p_j - \mathbb{I}[j = \text{target}] \right) w_j .
\end{equation}
\end{proof}

Since $p_j \in [0,1]$, the scalar error term is strictly bounded:
\begin{equation}
\left| p_j - \mathbb{I}[j = \text{target}] \right| \le 1 .
\end{equation}
Therefore, the backward signal passed to the backbone is bounded by the norm of the
embedding vectors $w_j$, and the gradient cannot explode regardless of model error.

\paragraph{2. Gradient of the Continuous Loss.}
For the regression objective,
\begin{equation}
\nabla_h \mathcal{L}_{\mathrm{continuous}}
= W_{\mathrm{continuous}}^\top \left( W_{\mathrm{continuous}} h - y_{\text{target}} \right) .
\end{equation}
The residual vector
\begin{equation}
r = W_{\mathrm{continuous}} h - y_{\text{target}}
\end{equation}
is unbounded. Under domain shift or mismatched scaling between the latent space and
target actions, $\|r\|$ may grow arbitrarily large.

\paragraph{3. Conclusion.}
\begin{equation}
\|\nabla_h \mathcal{L}_{\mathrm{continuous}}\| \propto \|\text{Error}\|
\quad \text{vs.} \quad
\|\nabla_h \mathcal{L}_{\mathrm{CE}}\| \le C .
\end{equation}
Large gradients in the continuous case propagate into $\theta$, causing large parameter updates.
In non-convex optimization, such updates move the model outside the basin of attraction
formed during pre-training. In contrast, the CE loss preserves gradient scales consistent
with pre-training dynamics, maintaining VLM capabilities.

\subsubsection{Proof 2: Feature Space Distortion (Scale Invariance Conflict)}

\paragraph{Proposition.}
\emph{Continuous prediction imposes constraints on the magnitude of $h$ that conflict
with Transformer normalization layers, whereas discrete classification is scale-invariant.}

\paragraph{Derivation.}

\paragraph{1. Transformer Dynamics.}
Transformers rely heavily on Layer Normalization:
\begin{equation}
\mathrm{LN}(h) = \frac{h - \mu}{\sigma} .
\end{equation}
In classification, predictions depend primarily on the angle (cosine similarity)
between $h$ and token embeddings $w$. Scaling $h$ by a constant $c$ affects confidence
but not the decision:
\begin{equation}
\arg\max (W (c h)) = \arg\max (W h) .
\end{equation}
Thus, semantic information is encoded mainly in the \emph{direction} of $h$.

\paragraph{2. Conflict in the Continuous Objective.}
For continuous modeling,
\begin{equation}
y = W_{\mathrm{continuous}} h .
\end{equation}
To produce large-magnitude actions, the backbone must increase $\|h\|$:
\begin{equation}
\|y\| \le \|W_{\mathrm{continuous}}\| \cdot \|h\| .
\end{equation}
Minimizing $\mathcal{L}_{\mathrm{continuous}}$ therefore induces gradients that explicitly
manipulate the norm of $h$ to match physical units, such as velocity or force.

\paragraph{3. Orthogonality of Objectives.}
Let $g_{\text{text}}$ denote the gradient induced by pre-training objectives, which
encourage angular separability of semantic features.
The continuous-version gradient $g_{\text{robot}}$ contains a magnitude-altering component.

If this magnitude requirement conflicts with LayerNorm or the natural feature scale,
the optimizer must substantially modify the internal weights $\theta$:
\begin{equation}
\langle g_{\text{text}}, g_{\text{robot}}^{\text{magnitude}} \rangle \approx 0 .
\end{equation}
This orthogonal conflict distorts semantic feature clusters.
The discrete objective $\mathcal{L}_{\mathrm{CE}}$, being scale-invariant, updates only
feature direction and thus avoids this conflict.

\subsection{Architectural Unification and Scaling via Discrete Action Spaces}\label{sect:appendix-unify}

Extensive literature advocates for unified architectures across diverse modalities, converging on the consensus that such alignment is crucial for fully exploiting scaling laws \citep{liang2025discrete,li2025discrete,luo2026being}. Given that pre-trained Vision-Language Models (VLMs)—and particularly Large Language Models (LLMs)—fundamentally operate on discrete tokens, adopting discrete representations for actions facilitates seamless architectural unification. These approaches ensure that action generation occurs within the transformer using the same Cross-Entropy training objective as the backbone, effectively treating images, language, and actions as a unified sequence. As this unification represents a foundational design paradigm widely validated in current research, we proceed based on this architectural premise without deriving a separate formal proof for its utility.

\subsection{The Intrinsic Quantization of Physical Hardware}\label{sect:appendix-nature}

Discrete diffusion aligns with the intrinsic nature of robotic control, where hardware constraints—such as encoder resolution, control frequency, and actuation latency—inevitably discretize continuous signals. Unlike continuous approaches that assume infinite precision, our discrete modeling framework is adaptable, capable of matching the varying resolution requirements of diverse physical systems. Empirically, we observe that while control accuracy initially improves as the number of discrete bins increases, it eventually reaches a saturation point. Beyond this threshold, increasing bin density yields no further gains and may even degrade performance (see Figure~\ref{fig-ablation}(a)). This phenomenon indicates that for any given robot, there exists an optimal quantization granularity dictated by practical physical dynamics rather than theoretical capacity. \textbf{Crucially, this effective optimal resolution often diverges from the nominal resolution stated in manufacturer specifications, suggesting that usable precision is limited by real-world factors (e.g., noise or mechanical backlash) rather than just sensor bit-depth.}

\section{Benefits of Tweedie Discrete Diffusion}\label{sec:fisher_equivalence}

Finally, we establish the theoretical superiority of Tweedie Discrete Diffusion Models over traditional discrete frameworks. We focus our comparison on mask-based discrete diffusion, which currently represents the state-of-the-art in this domain \citep{google_deepmind_gemini_diffusion_2025,nie2025large,ye2025dream,wu2025fast}.

Our analysis demonstrates that Tweedie Discrete Diffusion exhibits robust convergence properties. Specifically, the minimization of the training objective (Cross-Entropy, CE) guarantees the optimization of the underlying generative dynamics (Fisher divergence). In contrast, mask-based models—owing to the absence of a Gaussian forward process—lack these bidirectional bounds. Consequently, a reduction in CE loss does not strictly imply a reduction in Fisher divergence, leading to a potential misalignment between the training objective and the generative dynamics.

\subsection{Fisher Equivalence Theorem for Tweedie Discrete Diffusion}

\subsubsection{Theorem Statement and Physical Interpretation}

\begin{theorem}[Equivalence between Excess Risk and Fisher Divergence]
\label{thm:fisher_equivalence}
Benefiting from the adoption of a Gaussian forward process, \emph{Tweedie Discrete Diffusion} satisfies the following fundamental equivalence:
\begin{equation}
\mathrm{ExcessCE}(\theta) \asymp D_F\!\left(p_t \,\|\, p_{t,\theta}\right),
\end{equation}
where $\mathrm{ExcessCE}(\theta)$ denotes the excess cross-entropy risk and
$D_F(p_t \| p_{t,\theta})$ denotes the Fisher divergence (score matching error).
The equivalence constants are explicit and depend only on the signal-to-noise
ratio $\sigma_t^2/\alpha_t$ and a probability margin constant
$\delta(\sigma_t)$, and are independent of the number of classes $K$.
\end{theorem}

\paragraph{Physical Meaning and Convergence Analysis.}
The above equivalence implies that during training, whenever the cross-entropy
objective decreases by $\varepsilon$, the Fisher divergence necessarily
decreases by $\Theta(\varepsilon)$, and vice versa. Consequently:
\begin{itemize}
    \item \textbf{Tweedie Discrete Diffusion} enjoys strong convergence
    guarantees: optimization of the training objective (CE) directly ensures
    optimization of the underlying generative dynamics (score/Fisher).
    \item \textbf{Mask-Based Discrete Diffusion}, which does not employ a
    Gaussian forward process, lacks such bidirectional bounds. As a result, a
    reduction in CE does not necessarily correspond to a genuine decrease in
    Fisher divergence, potentially leading to a mismatch between the training
    objective and the generative dynamics.
\end{itemize}

\subsubsection{Proof and Derivation}

\paragraph{A. Forward Process and Training Objective.}
In Tweedie Discrete Diffusion, the forward noising process is defined as a
continuous Gaussian channel:
\begin{equation}
x_t = \sqrt{\alpha_t}\,x_0 + \sigma_t \epsilon,
\qquad \epsilon \sim \mathcal{N}(0, I).
\end{equation}
This formulation differs fundamentally from classical discrete diffusion
models based on Markov transition matrices. The model predicts the original
category $x_0 \in \{e_1,\dots,e_K\}$ via a softmax distribution
$q_\theta(k \mid x_t)$, and is trained using the cross-entropy risk
\begin{equation}
R(\theta) = \mathbb{E}\big[-\log q_\theta(x_0 \mid x_t)\big].
\end{equation}

\paragraph{B. Bayes Optimality and Excess Risk.}
Let $\boldsymbol{\pi}(x_t)$ denote the true posterior distribution, where
\begin{equation}
\pi_k(x_t) = \mathbb{P}(x_0 = e_k \mid x_t).
\end{equation}
By standard results in statistical learning theory, the Bayes-optimal predictor
\begin{equation}
q^*(\cdot \mid x_t) = \boldsymbol{\pi}(x_t)
\end{equation}
minimizes the expected risk, yielding
\begin{equation}
R^* = \min_\theta R(\theta)
= \mathbb{E}\big[H(\boldsymbol{\pi}(\cdot \mid x_t))\big].
\end{equation}
The excess cross-entropy risk is therefore given by
\begin{equation}
\mathrm{ExcessCE}(\theta)
:= R(\theta) - R^*
= \mathbb{E}_{x_t}\!\left[
\mathrm{KL}\big(\boldsymbol{\pi}(\cdot \mid x_t)
\,\|\, q_\theta(\cdot \mid x_t)\big)
\right].
\end{equation}

\paragraph{C. Core Bridge: Tweedie’s Identity.}
Since the forward process is Gaussian, Tweedie’s identity relates the posterior
expectation to the score of the marginal distribution:
\begin{equation}
\nabla_{x_t} \log p_t(x_t)
= \frac{\sqrt{\alpha_t}}{\sigma_t^2}
\big(\mathbb{E}[x_0 \mid x_t] - x_t\big)
= \frac{\sqrt{\alpha_t}}{\sigma_t^2}
\big(\boldsymbol{\pi}(x_t) - x_t\big).
\label{eq:tweedie_identity}
\end{equation}

This \textbf{Tweedie’s formula} is proved as follows.
\begin{proof}

We first consider the standard additive Gaussian case \( x_t = x_0 + \sigma \epsilon \).

\begin{equation}
\begin{aligned}
    \mathbb E[x_0|x_t] &= \int_{-\infty}^{\infty} x_0 p (x_0|x_t) d x_0 \\
     &= \int_{-\infty}^{\infty} x_0 \frac{p (x_t|x_0) p(x_0)}{p(x_t)} d x_0\\
     &= \frac{\int_{-\infty}^{\infty} x_0 p (x_t|x_0) p(x_0) d x_0}{p(x_t)}\\
     &= \frac{\int_{-\infty}^{\infty} x_0 \frac{1}{\sqrt{2\pi\sigma_t^2}} e^{-\frac{(x_t - x_0)^2}{2\sigma_t^2}} p(x_0) d x_0}{p(x_t)}\\
     &= \frac{\int_{-\infty}^{\infty} \Big[(x_0 - x_t) \frac{1}{\sqrt{2\pi\sigma_t^2}} e^{-\frac{(x_t - x_0)^2}{2\sigma_t^2}} p(x_0)  + x_t \frac{1}{\sqrt{2\pi\sigma_t^2}} e^{-\frac{(x_t - x_0)^2}{2\sigma_t^2}} p(x_0) \Big] d x_0}{p(x_t)}\\
     &= \frac{\int_{-\infty}^{\infty} \Big[\sigma_t^2\frac{(x_0 - x_t)}{\sigma_t^2} \frac{1}{\sqrt{2\pi\sigma_t^2}} e^{-\frac{(x_t - x_0)^2}{2\sigma_t^2}} p(x_0)  + x_t \frac{1}{\sqrt{2\pi\sigma_t^2}} e^{-\frac{(x_t - x_0)^2}{2\sigma_t^2}} p(x_0) \Big] d x_0}{p(x_t)}\\
     &= \frac{\int_{-\infty}^{\infty} \sigma_t^2\frac{(x_0 - x_t)}{\sigma_t^2} \frac{1}{\sqrt{2\pi\sigma_t^2}} e^{-\frac{(x_t - x_0)^2}{2\sigma_t^2}} p(x_0)d x_0  + \int_{-\infty}^{\infty}x_t \frac{1}{\sqrt{2\pi\sigma_t^2}} e^{-\frac{(x_t - x_0)^2}{2\sigma_t^2}} p(x_0)d x_0 }{p(x_t)}\\
     &= \frac{\sigma_t^2\int_{-\infty}^{\infty} \frac{d\Big[\frac{1}{\sqrt{2\pi\sigma_t^2}}e^{-\frac{(x_t - x_0)^2}{2\sigma_t^2}}\Big]}{dx_t} p(x_0)d x_0  + \int_{-\infty}^{\infty}x_t \frac{1}{\sqrt{2\pi\sigma_t^2}} e^{-\frac{(x_t - x_0)^2}{2\sigma_t^2}} p(x_0)d x_0 }{p(x_t)}\\
     &= \frac{\sigma_t^2\int_{-\infty}^{\infty} \frac{d\Big[p(x_t|x_0)\Big]}{dx_t} p(x_0)d x_0  + \int_{-\infty}^{\infty}x_t p(x_t|x_0) p(x_0)d x_0 }{p(x_t)}\\
     &= \frac{\sigma_t^2 \frac{d}{dx_t}\int_{-\infty}^{\infty} p(x_t|x_0)p(x_0)d x_0  + x_t\int_{-\infty}^{\infty} p(x_t|x_0) p(x_0)d x_0 }{p(x_t)}\\
     &= \frac{\sigma_t^2 \frac{d p(x_t)}{dx_t}  + x_t p(x_t) }{p(x_t)}\\
     & = \sigma_t^2 \frac{d}{dx_t} \log p(x_t) + x_t\\
\end{aligned}
\end{equation}
i.e.,
\begin{equation}\label{eqn:tweedie}
    \boxed{\mathbb E[x_0|x_t]
=x_t+\sigma_t^2\nabla_{x_t}\log p_t(x_t)}.
\end{equation}

When the diffusion involves a scaling factor \( \sqrt{\alpha_t} x_0 \), we apply a variable transformation
\[
x_t = \sqrt{\alpha_t}\, x_0 + \sigma_t\, \epsilon,
\]
equivalently
\[
x_t' = \frac{x_t}{\sqrt{\alpha_t}} = x_0 + \frac{\sigma_t}{\sqrt{\alpha_t}}\epsilon.
\]

Based on \(
x_t' = \frac{x_t}{\sqrt{\alpha_t}} = x_0 + \frac{\sigma_t}{\sqrt{\alpha_t}}\epsilon
\) and Eqn. \eqref{eqn:tweedie}, we have:
\begin{equation}
\begin{aligned}
     \mathbb E[x_0|x_t'] &=x_t' + \frac{\sigma_t^2}{\sqrt{\alpha_t}} \nabla_{x_t'}\log p_t(x_t')\\
  &= \frac{x_t}{\sqrt{\alpha_t}} + \frac{\sigma_t^2}{\alpha_t} \sqrt{\alpha_t}\nabla_{x_t}\log p_t(x_t'),
\end{aligned}
\end{equation}
which yields:
\[
\boxed{\mathbb E[x_0|x_t] = \frac{1}{\sqrt{\alpha_t}} (x_t + \sigma_t^2\nabla_{x_t}\log p_t(x_t)}.
\]

\end{proof}

An analogous identity holds for the model distribution $p_{t,\theta}$. Comparing
the two yields the pointwise relation
\begin{equation}
\boldsymbol{\pi}(x_t) - q_\theta(\cdot \mid x_t)
= \frac{\sigma_t^2}{\sqrt{\alpha_t}}
\big(
\nabla \log p_t(x_t) - \nabla \log p_{t,\theta}(x_t)
\big).
\label{eq:posterior_score_relation}
\end{equation}
Taking the squared expectation under $p_t$ gives an exact proportionality
between posterior error and Fisher divergence:
\begin{equation}
\mathbb{E}_{p_t}\big\|\boldsymbol{\pi} - q_\theta\big\|_2^2
= \frac{\sigma_t^4}{\alpha_t}\cdot
2\,D_F\!\left(p_t \,\|\, p_{t,\theta}\right).
\label{eq:fisher_exact}
\end{equation}

\paragraph{D. Establishing Bidirectional Bounds.}
Using \eqref{eq:fisher_exact}, we relate the excess cross-entropy risk to the
Fisher divergence.

\emph{Lower bound (Pinsker’s inequality).}
By Pinsker’s inequality,
$\mathrm{KL}(P\|Q) \ge \tfrac{1}{2}\|P-Q\|_2^2$, we obtain
\begin{equation}
\mathrm{ExcessCE}(\theta)
\ge \frac{1}{2}
\mathbb{E}\big\|\boldsymbol{\pi} - q_\theta\big\|_2^2
= \frac{\sigma_t^4}{\alpha_t}
D_F\!\left(p_t \,\|\, p_{t,\theta}\right).
\end{equation}

\emph{Upper bound (strong convexity of KL).}
Assume that Gaussian convolution ensures posterior probabilities are bounded
away from $0$ and $1$, i.e., there exists
$\delta(\sigma_t) \in (0, \tfrac12]$ such that
$\pi_k(x_t) \in [\delta, 1-\delta]$ almost everywhere. Then the KL divergence
satisfies
\begin{equation}
\mathrm{KL}(P\|Q)
\le \frac{1}{2\delta(1-\delta)} \|P-Q\|_2^2.
\end{equation}
Combining this with \eqref{eq:fisher_exact} yields
\begin{equation}
\mathrm{ExcessCE}(\theta)
\le
\frac{\sigma_t^4}{\alpha_t\,\delta(1-\delta)}
D_F\!\left(p_t \,\|\, p_{t,\theta}\right).
\end{equation}

\paragraph{Conclusion.}
The excess cross-entropy risk is thus sandwiched between constant multiples of
the Fisher divergence, completing the proof.

\subsection{Empirical Training Diagnostics}
\label{app:fisher_empirical}

We examine the empirical relationship between the categorical training objective and score-related diagnostics by monitoring cross-entropy (CE), a categorical Fisher--Rao (F-R) proxy, and a time-scaled score-error proxy throughout training. The two proxies are recorded for diagnostic purposes only and are not included as additional training losses. This analysis complements the theoretical connection between categorical posterior estimation and continuous score estimation by examining their associated training behavior.

\paragraph{Training dynamics.}
Table~\ref{tab:fisher_diagnostics} reports selected measurements at seven logged steps spanning six approximately equal-width training intervals. For each quantity, we also report its value relative to the corresponding measurement at step 10, allowing changes on different numerical scales to be compared.

Between steps 10 and 29,990, CE decreases from 1.013256 to 0.030871, while the categorical F-R proxy decreases from 9.606386 to 0.621511. The time-scaled score-error proxy decreases from $5.43317\times10^{-3}$ to $1.02788\times10^{-4}$. These changes correspond to relative reductions of 96.95\%, 93.53\%, and 98.11\%, respectively. The score-error proxy exhibits local fluctuations despite its overall reduction. For example, it increases from $2.995\times10^{-4}$ at step 5,000 to $3.767\times10^{-4}$ at step 10,000 before decreasing at the subsequent reported checkpoints. Thus, the observed improvement concerns the overall training trend and does not imply monotonic decreases at every step.

\begin{table}[t]
    \centering
    \caption{
    \textbf{Training diagnostics at selected optimization steps.}
    We report cross-entropy, a categorical Fisher--Rao proxy, and a time-scaled score-error proxy. Relative values are normalized by the corresponding measurement at step 10. Scientific-notation entries are rounded for display.
    }
    \label{tab:fisher_diagnostics}
    \small
    \setlength{\tabcolsep}{5pt}
    \renewcommand{\arraystretch}{1.15}
    \begin{tabular}{rcccccc}
        \toprule
        & \multicolumn{2}{c}{CE}
        & \multicolumn{2}{c}{Categorical F-R proxy}
        & \multicolumn{2}{c}{Scaled score proxy} \\
        \cmidrule(lr){2-3}
        \cmidrule(lr){4-5}
        \cmidrule(lr){6-7}
        Step
        & Value & Relative
        & Value & Relative
        & Value & Relative \\
        \midrule
        10
        & 1.013256 & 1.0000
        & 9.606386 & 1.0000
        & $5.433\times10^{-3}$ & 1.0000 \\
        5,000
        & 0.095172 & 0.0939
        & 1.299624 & 0.1353
        & $2.995\times10^{-4}$ & 0.0551 \\
        10,000
        & 0.080838 & 0.0798
        & 1.184651 & 0.1233
        & $3.767\times10^{-4}$ & 0.0693 \\
        15,000
        & 0.062705 & 0.0619
        & 1.007325 & 0.1049
        & $2.552\times10^{-4}$ & 0.0470 \\
        20,000
        & 0.047389 & 0.0468
        & 0.839231 & 0.0874
        & $2.089\times10^{-4}$ & 0.0384 \\
        25,000
        & 0.039821 & 0.0393
        & 0.749144 & 0.0780
        & $1.247\times10^{-4}$ & 0.0230 \\
        29,990
        & 0.030871 & 0.0305
        & 0.621511 & 0.0647
        & $1.028\times10^{-4}$ & 0.0189 \\
        \bottomrule
    \end{tabular}
\end{table}

\paragraph{Window-averaged behavior.}
To characterize the improvement beyond individual endpoints, we additionally compare early- and late-window means. The mean CE decreases from 0.215779 to 0.031886, and the mean categorical F-R proxy decreases from 2.375973 to 0.636343. These window-level measurements show that the reduction in both quantities is also present when averaging across multiple logged steps.

\paragraph{Correlation and local co-movement.}
Across logged training steps, CE and the categorical F-R proxy exhibit a strong positive association, with Pearson $r=0.991$ and Spearman $\rho=1.000$. We additionally examine their first differences to assess local co-movement beyond the shared long-run training trend. The first-difference correlation is $r=0.835$, indicating a positive association between their changes at successive logging points. This analysis complements the correlation between their recorded levels by examining shorter-term variation.

\stopcontents[appendix]

\newpage
\section*{NeurIPS Paper Checklist}

\begin{enumerate}

\item {\bf Claims}
    \item[] Question: Do the main claims made in the abstract and introduction accurately reflect the paper's contributions and scope?
    \item[] Answer: \answerYes{} 
    \item[] Justification: The claims in the abstract and introduction clearly reflect the main contributions, including the introduction of the Tweedie Discrete Diffusion framework and associated experiments.

\item {\bf Limitations}
    \item[] Question: Does the paper discuss the limitations of the work performed by the authors?
    \item[] Answer: \answerYes{} 
    \item[] Justification:  The paper provides a limitations section in the appendix.

\item {\bf Theory assumptions and proofs}
    \item[] Question: For each theoretical result, does the paper provide the full set of assumptions and a complete (and correct) proof?
    \item[] Answer: \answerYes{} 
    \item[] Justification: All theoretical results, including theorems, formulas, and proofs, are presented in the main text and appendix. The assumptions are clearly stated in the respective theorem statements, and all proofs are provided either in the main paper or in the supplemental material.

    \item {\bf Experimental result reproducibility}
    \item[] Question: Does the paper fully disclose all the information needed to reproduce the main experimental results of the paper to the extent that it affects the main claims and/or conclusions of the paper (regardless of whether the code and data are provided or not)?
    \item[] Answer: \answerYes{} 
    \item[] Justification: The paper provides a detailed description of the experimental setup, hyperparameters, and evaluation procedures in Section 4. We have also included instructions for reproducing the results, including model architectures and datasets used. The described procedures and experimental details should allow others to replicate the results.

\item {\bf Open access to data and code}
    \item[] Question: Does the paper provide open access to the data and code, with sufficient instructions to faithfully reproduce the main experimental results, as described in supplemental material?
    \item[] Answer: \answerYes{} 
    \item[] Justification: The paper provides the project webpage in the abstract where could find the code.

\item {\bf Experimental setting/details}
    \item[] Question: Does the paper specify all the training and test details (e.g., data splits, hyperparameters, how they were chosen, type of optimizer) necessary to understand the results?
    \item[] Answer: \answerYes{} 
    \item[] Justification: Yes, all experimental details including data splits, hyperparameters, and optimizer types are described in Section 4 of the paper. Additionally, the choices for hyperparameters are further explained in the appendix.
    \item[] Guidelines:
    \begin{itemize}
        \item The answer \answerNA{} means that the paper does not include experiments.
        \item The experimental setting should be presented in the core of the paper to a level of detail that is necessary to appreciate the results and make sense of them.
        \item The full details can be provided either with the code, in the appendix, or as supplemental material.
    \end{itemize}

\item {\bf Experiment statistical significance}
    \item[] Question: Does the paper report error bars suitably and correctly defined or other appropriate information about the statistical significance of the experiments?
    \item[] Answer: \answerNo{} 
    \item[] Justification: The paper does not report error bars or statistical significance tests for all the main experiments.

\item {\bf Experiments compute resources}
    \item[] Question: For each experiment, does the paper provide sufficient information on the computer resources (type of compute workers, memory, time of execution) needed to reproduce the experiments?
    \item[] Answer: \answerYes{} 
    \item[] Justification: Yes, the paper provides detailed information about the computational resources used for each experiment. 
    
\item {\bf Code of ethics}
    \item[] Question: Does the research conducted in the paper conform, in every respect, with the NeurIPS Code of Ethics \url{https://neurips.cc/public/EthicsGuidelines}?
    \item[] Answer: \answerYes{} 
    \item[] Justification: Yes, the research conducted in this paper fully conforms with the NeurIPS Code of Ethics.

\item {\bf Broader impacts}
    \item[] Question: Does the paper discuss both potential positive societal impacts and negative societal impacts of the work performed?
    \item[] Answer: \answerNo{} 
    \item[] Justification: This work is primarily foundational and focuses on improving the generalization and fine-grained control of VLA models for robotic action generation. The work itself does not target specific applications and therefore does not have a direct societal impact. However, if the technology were to be applied in industrial or service robots in the future, it could have positive effects, such as increasing productivity or assisting humans with tasks, as well as potential risks, such as safety issues in the case of misuse or incorrect operation. This paper does not evaluate such applications, but future deployment should consider safety, privacy, and responsible usage.
    
\item {\bf Safeguards}
    \item[] Question: Does the paper describe safeguards that have been put in place for responsible release of data or models that have a high risk for misuse (e.g., pre-trained language models, image generators, or scraped datasets)?
    \item[] Answer: \answerNA{} 
    \item[] Justification: The paper poses no such risks.

\item {\bf Licenses for existing assets}
    \item[] Question: Are the creators or original owners of assets (e.g., code, data, models), used in the paper, properly credited and are the license and terms of use explicitly mentioned and properly respected?
    \item[] Answer: \answerYes{} 
    \item[] Justification: Yes, all external assets used in this paper, including datasets and pre-trained models, are properly credited.

\item {\bf New assets}
    \item[] Question: Are new assets introduced in the paper well documented and is the documentation provided alongside the assets?
    \item[] Answer: \answerNA{} 
    \item[] Justification: The paper does not release new assets.

\item {\bf Crowdsourcing and research with human subjects}
    \item[] Question: For crowdsourcing experiments and research with human subjects, does the paper include the full text of instructions given to participants and screenshots, if applicable, as well as details about compensation (if any)? 
    \item[] Answer: \answerNA{} 
    \item[] Justification: The paper does not involve crowdsourcing nor research with human subjects.

\item {\bf Institutional review board (IRB) approvals or equivalent for research with human subjects}
    \item[] Question: Does the paper describe potential risks incurred by study participants, whether such risks were disclosed to the subjects, and whether Institutional Review Board (IRB) approvals (or an equivalent approval/review based on the requirements of your country or institution) were obtained?
    \item[] Answer: \answerNA{} 
    \item[] Justification: The paper does not involve crowdsourcing nor research with human subjects.

\item {\bf Declaration of LLM usage}
    \item[] Question: Does the paper describe the usage of LLMs if it is an important, original, or non-standard component of the core methods in this research? Note that if the LLM is used only for writing, editing, or formatting purposes and does \emph{not} impact the core methodology, scientific rigor, or originality of the research, declaration is not required.
    \item[] Answer: \answerNA{} 
    \item[] Justification: The core method development in this research does not involve LLMs as any important, original, or non-standard components.

\end{enumerate}

\end{document}